\documentclass[11pt]{article}

\usepackage[final]{acl}
\usepackage[utf8]{inputenc}
\usepackage{float}
\usepackage{tikz}
\usepackage{pgfplots}

\usepackage{multirow}
\usepackage{pgf-pie}
\usepackage{graphicx}
\usepackage{times}
\usepackage{latexsym}
\usepackage{xcolor}
\usepackage{totcount}

\newtotcounter{citnum} 
\usepackage{totcount}
\usepackage{etoolbox} 
\usepackage{float}
\usepackage{placeins}
\newtotcounter{unique_references}
\usepackage{caption}
\pretocmd{\bibitem}{\stepcounter{unique_references}}{}{}
\pgfplotsset{compat=1.17}

\definecolor{pastelblue}{HTML}{A1C9F4}
\definecolor{pastelred}{HTML}{FFB482}
\definecolor{pastelgreen}{HTML}{8DE5A1}
\definecolor{pastelyellow}{HTML}{FFD97D}
\definecolor{pastelpurple}{HTML}{D0BBFF}
\definecolor{pastelpink}{HTML}{FF9E9E}
\definecolor{pastelteal}{HTML}{9CF4E3}
\definecolor{pastelgray}{HTML}{CFCFCF}
\definecolor{pastelaqua}{HTML}{9EE5FF}
\usepackage[T1]{fontenc}

\usepackage[utf8]{inputenc}
\usepackage{booktabs}

\usepackage{graphicx}
\usepackage{amsmath} 
\usepackage{graphicx}
\usepackage{tikz}
\usepackage{forest}
\usepackage{latexsym}
\usepackage{forest}
\usetikzlibrary{positioning} 
\usepackage{natbib}
\usepackage[svgnames]{xcolor}

\DeclareRobustCommand{\HindiWords}[1]{%
  \raisebox{-\dp\strutbox}{%
    \includegraphics[page=#1]{HindiWords}%
  }%
}

\title{Beyond Monolingual Assumptions: A Survey on Code-Switched NLP in the Era of Large Language Models across Modalities}

\author{
\textbf{Rajvee Sheth\textsuperscript{$\S$$\scriptscriptstyle\Diamond$}}, 
{
\textbf{Samridhi Raj Sinha\textsuperscript{$\star$$\scriptscriptstyle\Diamond$}\thanks{Work done while interning at IIT Gandhinagar.}, 
\textbf{Mahavir Patil\textsuperscript{$\dagger$$\scriptscriptstyle\Diamond$}}\footnotemark[1]},
}
 \\
\textbf{Himanshu Beniwal\textsuperscript{$\S$$\scriptscriptstyle\Diamond$}}, 
\textbf{Mayank Singh\textsuperscript{$\S$$\scriptscriptstyle\Diamond$}}\\[4pt]
\textsuperscript{$\S$}IIT Gandhinagar,
\textsuperscript{$\star$}NMIMS Mumbai, 
\textsuperscript{$\dagger$}SVNIT Surat, \textsuperscript{$\scriptscriptstyle\Diamond$}LINGO Research Group\\[4pt]
\small{\textbf{Correspondence:} \href{mailto:singh.mayank@iitgn.ac.in}{singh.mayank@iitgn.ac.in}}
}

\begin{document}
\maketitle
\begin{abstract} Amidst the rapid advances of large language models (LLMs), most LLMs still struggle with mixed-language inputs, limited Code-switching (CSW) datasets, and evaluation biases, which hinder their deployment in multilingual societies. This survey provides the first comprehensive analysis of CSW-aware LLM research, reviewing \total{unique_references} studies spanning five research areas, 15+ NLP tasks, 30+ datasets, and 80+ languages. We categorize recent advances by architecture, training strategy, and evaluation methodology, outlining how LLMs have reshaped CSW modeling and identifying the challenges that persist. The paper concludes with a roadmap that emphasizes the need for inclusive datasets, fair evaluation, and linguistically grounded models to achieve truly multilingual capabilities.\footnote{A curated collection of all resources is maintained at 
\url{https://github.com/lingo-iitgn/awesome-code-mixing/}.}

\end{abstract}

\section{Introduction}

\begin{figure*}
    \centering
    \includegraphics[width=1\linewidth, keepaspectratio]{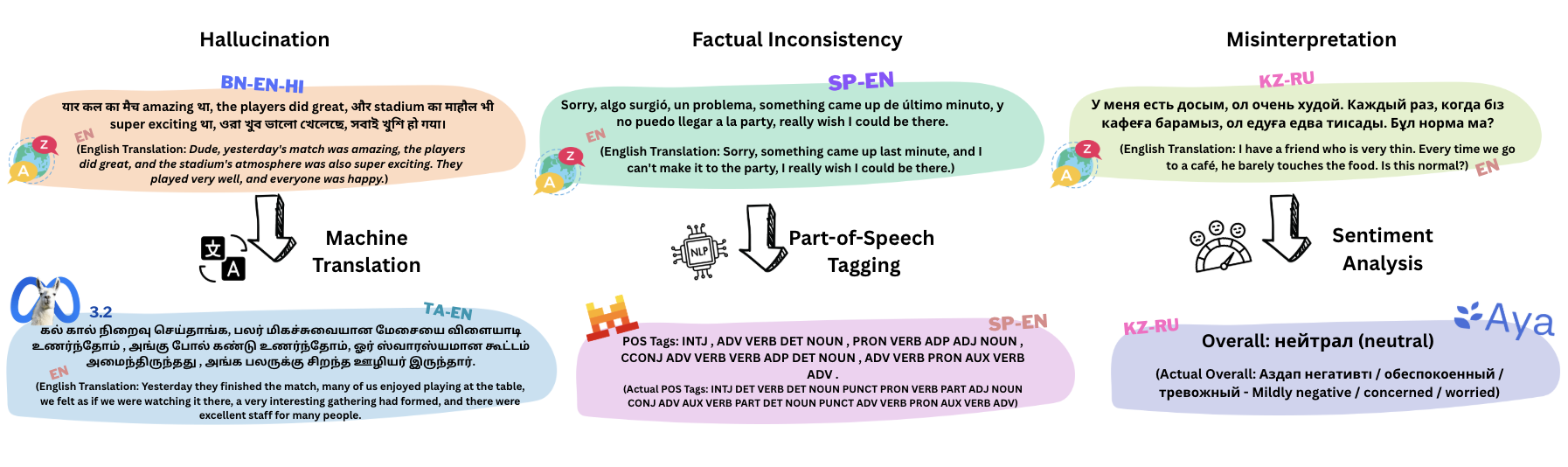}
    \caption{Common model failures on code-mixed text across tasks: \textit{\textbf{Takeaway} (a) hallucination in MT for a constructed trilingual (Bn-Hi-En) example, highlighting underexplored settings; (b) factual inconsistency in POS tagging for Spanish-English data \citet{zeng-2024-leveraging}; and (c) misinterpretation in sentiment analysis for Kazakh-Russian text \citet{goloburda2025qorgau}.}}
    \label{fig:architecture1}
    \vspace{-\baselineskip}
\end{figure*}

Code-switching (CSW), the alternation between two or more languages within a single utterance or discourse, is a pervasive feature of multilingual communication worldwide \citep{poplack1988contrasting}. With the rise of digital platforms, code-switched text has become ubiquitous across social media and online communication \citep{molina-etal-2016-overview, singh2017towards}, challenging NLP systems built on monolingual assumptions. Globally, approximately 43\% of the population is bilingual and an additional 13\% is trilingual \citep{preply2022,rosettastone2025}, representing over 4.5 billion multilingual speakers. Despite this prevalence, Monolingual ASR systems struggle with code-switched input: word error rates increase by 30–50\% \citep{singh2025hiacc}. Even multilingual models show semantic accuracy drops of 15\% \citep{winata-etal-2021-multilingual}, revealing 
a fundamental architectural gap. Similar challenges are observed in multilingual regions including India, Nigeria, and South Africa, where frequent CSW undermines monolingual ASR performance \citep{babatunde-etal-2025-beyond}. Figure~\ref{fig:architecture1} depicts intra- and inter-sentential code-mixing across multiple language pairs, emphasizing the linguistic variability that NLP systems must navigate.


The evolution of CSW research mirrors key milestones in NLP. The \textit{Early Statistical Era} (pre-2010) relied on rule-based and probabilistic models like n-grams, HMMs, and CRFs, laying the groundwork for bilingual text processing \citep{solorio-liu-2008-learning}. The \textit{Representation Learning Era} (2010–2017) introduced distributed embeddings (Word2Vec) along with neural architectures, advancing CSW tasks like LID, POS, and NER \citep{solorio-etal-2014-overview, sequiera-etal-2015-pos, molina-etal-2016-overview}. The \textit{Contextual Understanding Era} (2017–2020) brought GPT, BERT, XLM, and T5, enabling fine-tuning for code-switched data, though multilingual pretraining alone proved insufficient for robust CSW modeling \citep{winata-etal-2021-multilingual}. The  \textit{Foundation Model Era} (2020–present) leverages massive, instruction-tuned LLMs like GPT-3 and LLaMA capable of general-purpose reasoning through multilingual pretraining and prompt-based adaptation \citep{wang-etal-2025-investigating-scaling}.


LLMs have transformed CSW investigation across typologically diverse language pairs, including Arabic-English \citep{Issa2025DelvingIB}, Cantonese-Mandarin \citep{dai-etal-2025-next}, Chinese-English \citep{kong2025decoding}, Hinglish \citep{sheth-etal-2025-comi}, Korean-English \citep{yoo-etal-2025-code-switching}, Spanish–Guaraní \citep{kellert2025parsing}, and Ukrainian-Russian \citep{shynkarov-etal-2025-improving}, deepening our linguistic and sociocultural understanding of switching patterns \citep{Yoo2024CodeSwitchingRL, jehan2025evolution}. These advances are enabled by methodological innovations in LLMs, including in-context mixing \citep{shankar-etal-2024-context}, instruction tuning \citep{lee-etal-2024-commit}, speech processing \citep{kang-2024-covoswitch}, advanced metrics evaluating structural and socio-pragmatic aspects \citep{ugan2025pier, sterner-teufel-2025-minimal}, and curriculum strategies for transfer \citep{yoo-etal-2025-code-switching}.


\paragraph{Research Gap} Despite advances, LLMs struggle with zero-shot transfer to real-world CSW scenarios \citep{winata-etal-2023-decades}. 
Multilingual LLMs often underperform compared to fine-tuned smaller models, showing that ``\textit{multilingualism}'' alone does not ensure CSW proficiency \citep{zhang-etal-2023-multilingual}. LLMs also exhibit asymmetric performance: non-English tokens in English contexts degrade performance, while English tokens in other languages often enhance it \citep{mohamed2025lost}, followed by limited pretraining data for low-resource languages \citep{yoo-etal-2025-code-switching}. Prior works, including \citep{winata-etal-2023-decades} and \citep{sitaram2019survey}, have laid essential groundwork for understanding CSW in NLP but focus primarily on pre-LLM approaches and text-centric settings. Building on these foundations, we position this work as the first comprehensive survey of CSW NLP in the LLM era, covering recent advances in instruction tuning, PEFT, and multimodal (speech and vision-language) settings. We further survey the evolution of CSW research in the LLM era and present this unified taxonomy (Figure~\ref{fig:lit_survey}), with a detailed analysis in Appendix \S\ref{sec:Taxonomy} that categorizes prior work into five key research directions.

\paragraph{Position} We posit that Code-switching should be treated as a core modelling challenge rather than a downstream artefact of multilinguality. In low-resource CSW settings, current monolingual and multilingual models struggle to cope as language switching interacts with script alternation, orthographic variation, and spontaneous conversational behavior, which remain insufficiently represented in existing models and benchmarks across speech, dialogue, and multimodal contexts. Bridging this gap requires structured data creation, scalable modeling, and human-in-the-loop annotation, reinforced by language- and script-aware CS-specific architectures rather than broad, undifferentiated, multilingual post-training.


\paragraph{Contributions} The key contributions include: \textbf{(i)} We provide the first comprehensive survey of CSW research in the LLM era, analyzing \total{unique_references} studies across 15 NLP tasks, 30+ datasets spanning 80+ languages, across diverse language pairs, real-world applications, and key architectural innovations. \textbf{(ii)} We present a taxonomy (Appendix \S\ref{sec:Taxonomy} Figure~\ref{fig:lit_survey}) organizing LLM-based CSW research by architecture, training paradigm, and evaluation, while revealing key gaps in low-resource coverage, script diversity, cross-lingual transfer, the absence of unified evaluation frameworks. \textbf{(iii)} We present a roadmap for future CSW research, highlighting the need for inclusive datasets, equitable models, and fair metrics to support linguistically grounded advances in dialogue, speech, and multimodal contexts.

\section{Pre-LLM-Era works} 

Early computational approaches to code-switched word processing relied on rule-based and statistical models for foundational tasks such as language identification (LID) \citep{molina-etal-2016-overview, gundapu-mamidi-2018-word, Shekhar2020LanguageIF, solorio-liu-2008-learning, chittaranjan-etal-2014-word, king-etal-2014-iucl}, part-of-speech (POS) tagging \citep{vyas-etal-2014-pos, raha-etal-2019-development, pratapa-etal-2018-word, sequiera-etal-2015-pos}, named entity recognition (NER) \citep{ansari2019cross, singh-etal-2018-named, singh-etal-2018-language}, and sentiment analysis (SA) \citep{patwa-etal-2020-semeval, joshi-etal-2016-towards}. Methods included CRFs and CNNs for LID \citep{solorio-liu-2008-learning, chittaranjan-etal-2014-word}, CNNs with n-grams for POS tagging \citep{vyas-etal-2014-pos}, character-level RNNs and SVMs for NER \citep{singh-etal-2018-named}, and SVM-based sentiment classification \citep{joshi-etal-2016-towards}. BiLSTM-CRF models with embeddings later improved LID and NER, reducing perplexity \citep{chopra-etal-2021-switch-point, zhang-etal-2023-multilingual}, while switch-point sampling enhanced LID performance \citep{chatterjere-etal-2020-minority}. However, these approaches were limited by task and language-specific designs, shallow features, scarce labeled data, and poor cross-linguistic transfer \citep{molina-etal-2016-overview, Shekhar2020LanguageIF}. The fragmented nature of research led to isolated solutions, preventing the use of shared representations or unified frameworks across diverse CSW contexts \citep{winata-etal-2023-decades, liu-etal-2022-mulzdg, chi2025understanding}. The rise of LLMs has shifted CSW research toward unified, multilingual frameworks across speech processing, conversational and generation tasks, motivating surveys to examine their adaptation, emerging trends, and future directions.

\section{Code-Switching Task Landscape: Capabilities and Gaps}

\subsection{Traditional Tasks}

The integration of LLMs into traditional NLP tasks has revealed both transformative capabilities and inherent limitations in CSW contexts. In \textbf{language identification}, innovative fine-grained techniques such as TongueSwitcher for boundary detection in morphologically mixed German-English words \citep{sterner-teufel-2023-tongueswitcher}, MaskLID for training-free iterative identification of subdominant languages \citep{kargaran-etal-2024-masklid}, and equivalence constraint-guided methods for grammatical switch points \citep{Kuwanto2024LinguisticsTM} have established new benchmarks, with applications extending to hope/offensive speech detection and efficient zero/few-shot adaptation via models like COOLI and SetFit \citep{Ahmad2025HopeSD, balouchzahi-etal-2021-mucs-lt, pannerselvam-etal-2024-setfit}. \textbf{Part-of-Speech tagging} has benefited from contextual embeddings (e.g., mBERT on Arabic-English and Hinglish \citep{sabty-etal-2020-contextual, aguilar-solorio-2020-english}), bilingual pretraining on datasets like GLUECoS \citep{winata-etal-2021-multilingual, prasad-etal-2021-effectiveness}, parallel synthetic data generation (PACMAN \citep{chatterjee-etal-2022-pacman}), prompt-based CSW synthesis (PRO-CS and CoMix \citep{bansal-etal-2022-pro, arora-etal-2023-comix, kumar-etal-2022-utilizing}), and S-index-augmented XLM-R fine-tuning \citep{absar-2025-fine}. \textbf{Named Entity Recognition} has evolved from early embedding-attention approaches for Spanglish tweets \citep{wang-etal-2018-code} to synthetic CSW pretraining with MELM \citep{zhou-etal-2022-melm} and two-stage CMB models \citep{pu-etal-2022-cmb}, supported by benchmarks like MultiCoNER and toolkits such as CodemixedNLP \citep{malmasi-etal-2022-multiconer, jayanthi-etal-2021-codemixednlp}. Advances, including contextualized embeddings \citep{sabty-etal-2020-contextual}, pseudo-labeling \citep{el-mekki-etal-2022-um6p}, switch-point–biased self-training \citep{chopra-etal-2021-switch-point}, prompt-based methods like PRO-CS \citep{bansal-etal-2022-pro}, and data augmentation in CoSDA-ML \citep{ijcai2020p0533}, have further enhanced zero-shot transfer capabilities. 

Despite task-specific advances, multilingual models underperform on code-mixed inputs compared to monolingual settings \citep{wang-etal-2025-investigating-scaling}, primarily due to the limited representativeness of CSW in pretraining data \citep{dogruoz-etal-2023-representativeness}. These issues are further aggravated in low-resource and typologically distant languages \citep{sravani-mamidi-2023-enhancing}. Key challenges include modeling ambiguous switch points \citep{chopra-etal-2021-switch-point} and mitigating hallucinations in generative CSW outputs \citep{wang-etal-2025-gpt}. While prompting techniques, zero-shot models (e.g., GLiNER \citep{zaratiana-etal-2024-gliner}), generative frameworks (e.g., GPT-NER \citep{wang-etal-2025-gpt}), and LLM-based post-processing \citep{dai-etal-2025-next, khatri-etal-2023-translate} offer promising few-shot adaptability, these methods alone cannot fully overcome the deeper structural and resource-related challenges inherent in real-world code-mixing.


\subsection{Emerging Contemporary Tasks} 
LLMs have advanced performance in complex CSW tasks, yet continue to expose limitations in cross-lingual reasoning and cultural adaptation. In \textbf{Natural Language Inference}, early conversational datasets revealed persistent annotation disagreements and cultural ambiguities \citep{khanuja-etal-2020-new, huang-yang-2023-culturally}. Synthetic CoSDA-ML data enabled zero-shot transfer \citep{ijcai2020p0533}, while in-context mixing (ICM) prompting improved contextual reasoning \citep{shankar-etal-2024-context, prasad-etal-2021-effectiveness, kumar-etal-2022-utilizing}, though pragmatic variability continues to cause marked drops relative to monolingual performance. Similarly, \textbf{Question Answering} benefited from LLM-based architectures such as COMMIT \citep{lee-etal-2024-commit}, multimodal knowledge-distillation approaches \citep{raj-khan-etal-2021-towards-developing}, non-English prompting for grammaticality improvements \citep{behzad-etal-2024-ask}, curriculum-based CSW pretraining \citep{yoo-etal-2025-code-switching}, domain-specific embedding migration (MIGRATE) for low-resource reasoning \citep{hong-etal-2025-migrate}, and large-scale African benchmarks such as MEGAVERSE \citep{ahuja-etal-2024-megaverse}, building on earlier multilingual reading comprehension systems \citep{gupta-etal-2018-uncovering}. Parallel advances emerged in \textbf{Intent Classification} and slot filling, where contrastive pretraining across languages \citep{lin-etal-2024-contrastivemix}, prompt-based methods such as PRO-CS \citep{bansal-etal-2022-pro}, multilingual semantic parsing \citep{duong-etal-2017-multilingual-semantic, whitehouse-etal-2022-entitycs}, and zero-shot transfer with XLM-R \citep{Arora2020CrosslingualTL, krishnan-etal-2021-multilingual, wang-etal-2022-zero} improved cross-lingual generalization.

While LLMs have advanced emerging CSW tasks, persistent limitations remain in contextual reasoning and discourse grounding. Studies on code-mixed QA, NLI, intent detection, and dialogue show that semantic evidence is often fragmented across languages, leading models to rely on shallow lexical cues rather than compositional reasoning \citep{gupta-etal-2018-uncovering, chakravarthy-etal-2020-detecting, krishnan-etal-2021-multilingual}. Inference tasks further exhibit label instability due to culturally contingent interpretations \citep{khanuja-etal-2020-new}, while generative models struggle to maintain discourse coherence and stable switching patterns \citep{mehnaz-etal-2021-gupshup}. Text Generation suffers from data scarcity, especially for typologically distant and non-Latin-script languages \citep{sravani-mamidi-2023-enhancing}. These challenges, compounded by sociolinguistic variation in pragmatic norms \citep{park-etal-2024-multiprageval}, motivate linguistically grounded and lightweight modeling approaches for realistic CSW deployment \citep{raj-khan-etal-2021-towards-developing}.

\noindent See Appendix \S\ref{sec:appendix-tasks1} and \S\ref{sec:appendix-tasks2} for a detailed discussion of remaining tasks, with associated datasets and approaches in Appendix \S\ref{sec:appendix-supp} (Table~\ref{tab:cs-task-datasets}).


\subsection{Underexplored Frontiers Tasks}
Although core CSW tasks have advanced, conversational, speech, and multimodal CSW remain underexplored, posing both opportunities and challenges for adapting LLMs to naturalistic multilingual mixing.
\textbf{Reasoning tasks}, including mathematical problem-solving and cross-language entailment, struggle with logical complexity and semantic drift in CSW contexts \citep{raihan-etal-2023-sentmix, mohamed2025lost}, abstract level phenomena such as metaphor comprehension, analogical reasoning, and verb-level code-mixing preferences expose cultural biases and shallow understanding \citep{kodali2025human, mehnaz-etal-2021-gupshup, choudhary-etal-2026-llms}. Beyond reasoning, \textbf{code generation} from mixed prompts achieves only moderate functional correctness \citep{yang-chai-2025-codemixbench, khatri-etal-2023-translate}, despite progress in controllable CSW generation using encoder–decoder models \citep{mondal-etal-2022-cocoa}. \textbf{Conversational systems and dialogue} show emerging gains, with RAG-based architectures improving CSW customer support \citep{kruk2025banglassist}, multilingual dialogue benchmarks enabling few-shot agents for low-resource pairs such as Choctaw–English \citep{brixey-traum-2025-code}, and personality-aware response generation supporting coherent Hinglish multi-party dialogue \citep{kumar-chakraborty-2024-harmonizing}. In parallel, \textbf{Safety-oriented studies} emphasize region-specific prompting for Kazakh-Russian evaluation \citep{goloburda-etal-2025-qorgau}, while while recent work shows that code-mixing itself can be exploited as a trigger for \textbf{backdoor attacks}, raising concerns about robustness and security in CSW-aware NLP systems. \textbf{Document processing} has also been explored through multilingual OCR and contrastive representation learning for Vietnamese-English text \citep{dereza-etal-2024-million, do-etal-2024-contrastivemix}. 


\paragraph{\textit{Takeaway}} \textit{Although notable progress has been
made in core CSW NLU tasks, many frontier areas such as safety and visual processing remain underexplored, highlighting opportunities to extend research beyond existing linguistic and computational paradigms.}

\section{Datasets and Resources}

\subsection{Datasets}

The development of CSW datasets has evolved from manual annotation to LLM-driven scalable creation, highlighting trade-offs between expanded multilingual coverage and the authenticity of natural code-switching. However, as pre-training datasets continue to scale, manual curation becomes a challenge. For \textbf{Multilingual coverage}, large-scale corpora pre-trained on mixed-language text enhance NLU transfer through synthetic augmentation \citep{zhang-etal-2024-enhancing-multilingual}, while manually annotated datasets like the Multilingual Identification of English CSW benchmark switch across unseen languages \citep{sterner-2024-multilingual}, SwitchLingua spans 420k samples and over 80 hours of audio across 12 languages and 63 ethnic groups with LLM-assisted bias reduction \citep{xie2025switchlingua}, and MEGAVERSE provides LLM-driven benchmarks covering 22 datasets in 83 languages for multimodal evaluation \citep{ahuja-etal-2024-megaverse}. MultiCoNER uses LLM synthetic augmentation across 3 domains and 12 languages with 33 entity classes for code-mixed NER \citep{malmasi-etal-2022-multiconer}, NusaX offers human annotated parallel sentiment corpus for 10 Indonesian languages \citep{winata-etal-2023-nusax}, and GLOSS synthesizes texts for absent language pairs without manual curation \citep{hsu-etal-2023-code}. For \textbf{Low-resource languages}, targeted datasets address critical underrepresentation through BnSentMix for Bengali–English \citep{alam-etal-2025-bnsentmix}, DravidianCodeMix spanning Tamil-, Kannada-, and Malayalam–English \citep{chakravarthi2022dravidiancodemix}, Marathi–English corpora \citep{joshi-etal-2023-my}, SentMix for trilingual NLI \citep{raihan-etal-2023-sentmix}, GPT-3.5 synthetic Afrikaans– and Yoruba–English data \citep{terblanche-etal-2024-prompting}, and X-RiSAWOZ with over 18k utterances \citep{moradshahi-etal-2023-x}, collectively diversifying CSW NLP and reducing dependency on high-resource pairs.In \textbf{Synthetic data generation}, diverse approaches have addressed annotation scarcity: Bengali–English dependency parsing with large synthetic treebanks (270K+ sentences) \citep{winata-etal-2019-code}, PhraseOut for Hinglish NMT \citep{jasim-etal-2020-phraseout}, semi-supervised generation using pre-trained encoders \citep{gupta-etal-2020-semi}, CoSDA-ML for zero-shot NLI across 19 languages \citep{ijcai2020p0533}, ternary sequence labeling with mBERT for Hinglish MT \citep{gupta-etal-2021-training}, VACS for perplexity reduction \citep{ijcai2019p0719}, COMMIT for low-resource QA \citep{lee-etal-2024-commit}, LLM-generated puns and sentiment data (including 49K+ synthetic samples) for Spanglish and Malayalam–English \citep{zeng-2024-leveraging, sarrof2025homophonic}, In-Context Mixing for intent classification \citep{shankar-etal-2024-context}, SynCS for zero-shot gains via parallel alignment \citep{wang-etal-2025-investigating-scaling}, and naturalistic parallel CSW datasets for PLM evaluation \citep{Leon2024CodeMixedPS}. Despite these advances, the move toward LLM-driven, large-scale datasets raises concerns about capturing sociolinguistic nuance and authentic representation, with human evaluations showing only 60–65\% acceptability, and highlighting that high-quality resources for underrepresented languages still depend heavily on expert curation and community involvement \citep{kodali2025human}. \\

\noindent Refer to Tables~\ref{tab:cs-datasets-expanded} and~\ref{tab:speech-datasets-expanded} in Appendix \S\ref{sec:appendix-supp} for a summary of CSW text and speech datasets.

\subsection{Frameworks and Toolkits} 
To address the growing complexity of CSW research, frameworks and toolkits have emerged to standardize methodologies and streamline data creation across annotation and generation. \textbf{Annotation frameworks}, include CoSSAT, which supports fine-grained word-level and syllable-level speech annotation \citep{shah-etal-2019-cossat}; COMMENTATOR, which integrates LLMs for robust text annotation and prediction \citep{sheth-etal-2024-commentator}; CHAI, which leverages RLAIF to iteratively refine code-mixed translation annotations \citep{zhang2025chaillmsimprovingcodemixed}; and multimodal tools such as ToxVidLM, extending annotation to video by jointly modeling visual and textual CSW signals \citep{maity-etal-2024-toxvidlm}. \textbf{Synthetic data generation toolkits} include GCM, which produces linguistically grounded code-mixed text using established switching theories \citep{rizvi-etal-2021-gcm} (as utilized in \citep{huzaifah-etal-2024-evaluating}); and CodemixedNLP, an open-source toolkit offering models, datasets, and synthetic augmentation for seven Hinglish tasks \citep{jayanthi-etal-2021-codemixednlp}. Together, these tools enable scalable corpus creation and reproducible CSW research for downstream tasks such as machine translation and sentiment analysis \citep{sravani-mamidi-2023-enhancing, zeng-2024-leveraging}.

\paragraph{\textit{Takeaway}} \textit{LLM-augmented datasets such as SwitchLingua, BnSentMix, and COMMIT expand CSW resources for low-resource languages and improve model performance. However, synthetic data may lack naturalness and cultural nuance, introducing biases. Semi-automated, human-in-the-loop annotation toolkits can help create more authentic and equitable CSW benchmarks.}

\section{Model Training \& Adaptation}
\subsection{Mainstream Pre-training Approaches} 
Pre-training encodes mixed-language structure at scale, yielding transferable representations for diverse CSW tasks. \textbf{Specialized code-mixed models} trained on real code-mixed corpora consistently outperform multilingual baselines by directly capturing CSW dynamics. HingBERT and related models pre-trained on large-scale real-world data outperform mBERT and XLM-R on downstream NLP tasks \citep{nayak-joshi-2022-l3cube}. Probing studies further show that fine-tuning mBERT on curated naturalistic CSW data yields stronger attention patterns than synthetic mixing across Spanish-English and Hinglish pairs \citep{santy-etal-2021-bertologicomix}. Linguistically constrained synthetic embeddings improve over bilingual baselines for sentiment analysis(SA) and POS tagging \citep{pratapa-etal-2018-word}, while switch-aware architectures such as CONFLATOR emphasize language junctions to achieve state-of-the-art results on Hinglish SA and translation \citep{mohammed-etal-2023-conflator}. For \textbf{Task-adaptive pre-training}, targeted strategies explicitly encode CSW structure. Boundary-aware masked language modeling that integrates synthetic CSW data improves downstream QA and SA performance on CSW benchmarks \citep{das-etal-2023-improving}. Model-merging approaches combining continued pre-training with checkpoint fusion outperform standard fine-tuning \citep{kodali2025adapting}. Alignment-based methods leveraging parallel text enhance SA analysis and QA \citep{fazili-jyothi-2022-aligning}, while joint LID–POS multi-task models better capture social media CSW patterns \citep{dowlagar-mamidi-2021-pre}. Multilingual augmentation through synthetic CSW generation improves zero-shot intent detection and slot filling \citep{krishnan-etal-2021-multilingual}, and large-scale CSW pre-training with diverse synthetic mixtures yields stronger benchmarks and improved language alignment \citep{wang-etal-2025-investigating-scaling}.



\subsection{Mainstream Fine-tuning Approaches}

Fine-tuning adapts models to task-specific CSW distributions, improving in-domain performance but limiting generalization to unseen language pairs. \textbf{Task-specific fine-tuning} yields competitive in-domain results but depends heavily on curated CSW data: transformer-based fine-tuning achieves better word-level LID on low-resource Kannada-English pair \citep{lambebo-tonja-etal-2022-transformer}, fine-tuned XLM-RoBERTa introduces the S-index for measuring switching intensity and demonstrating effective generalization \citep{absar-2025-fine}, fine-tuned mBERT provides baselines for sentiment analysis on noisy social media data \citep{palomino-ochoa-luna-2020-palomino}, fine-tuned multilingual models like mBART and mT5, often combined with back-translation and ensembling, deliver fluency and accuracy for translation \citep{arindam-etal-2023-lost,khan-etal-2022-sit}, LLM fine-tuning with syntactic post-processing enhances Cantonese-to-Mandarin translation quality across domains \citep{dai-etal-2025-next}, and efficient monolingual ASR fine-tuning substantially lowers WER on Yoruba-English code-switched speech compared to larger zero-shot multilingual models, though it degrades performance on the non-target (English) language \citep{babatunde-etal-2025-beyond}. \textbf{Multi-task fine-tuning} leverages synergies for added robustness but can introduce negative transfer or require careful task balancing: syntax-aware joint training of language modeling and parsing lowers perplexity on Mandarin-English data \citep{winata-etal-2018-code}, intermediate-task fine-tuning on bilingual auxiliaries yields consistent gains in NLI, QA, and sentiment across Hinglish and Spanish-English \citep{prasad-etal-2021-effectiveness}, shared representations enhance offensive speech detection on Hinglish tweets and joint NER modeling in low-resource Arabic dialects \citep{amazouz2017addressing}, multi-directional fine-tuning and adapter-based methods improve translation and modular transfer \citep{kartik-etal-2024-synthetic, RATHNAYAKE2024107239}, and contrastive multi-task pretraining boosts zero-shot information retrieval and transfer \citep{do-etal-2024-contrastivemix}.

\noindent A detailed discussion of remaining pre-training and fine-tuning approaches is provided in Appendix \S\ref{sec:appendix-pre} and \S\ref{sec:appendix-fine}.

\subsection{Post-training Approaches}

While post-training approaches enable rapid CSW adaptation with minimal or no labeled data, their effectiveness varies widely across language pairs and task types. \textbf{Zero-shot} CSW methods rely on prompting, heuristic switching, or synthetic augmentation, including prompt-based CSW generation with GPT-3.5 \citep{yong-etal-2023-prompting}, entity-driven switching for slot filling and dialogue \citep{whitehouse-etal-2022-entitycs, liu-etal-2022-mulzdg}, and data-centric augmentation for MT and classification \citep{gupta-2022-malm, lai-etal-2021-saliency-based, krishnan-etal-2021-multilingual, ijcai2020p0533}. However, even strong LLMs such as GPT-4 exhibit significant performance drops in zero-shot CSW, with outcomes highly sensitive to pretraining language composition \citep{zhang-etal-2023-multilingual, tatariya-etal-2023-transfer}. \textbf{One- and few-shot} methods leverage limited to few examples through adapted prompting, including similarity-based prompting with ChatGPT \citep{11016028}, RAG-based in-context learning for hate speech detection \citep{srivastava2025dweshvaani}, multi-task LLM fine-tuning for harmful content in memes \citep{kumar2025multi}, generative transformers for emotion detection in Bangla-English-Hindi \citep{goswami-etal-2023-offmix}, and translation with LLM classification for affective tasks \citep{10938193}. \textbf{Instance-based prompting} further enhances performance, with PRO-CS using mBERT with Hinglish prompts improving NER and POS tagging \citep{bansal-etal-2022-pro}, GLOSS synthesizing CSW text for unseen pairs through self-training \citep{hsu-etal-2023-code}, DweshVaani's RAG retrieving Hinglish examples boosting hate speech detection \citep{srivastava2025dweshvaani}, and In-Context Mixing improving intent classification on MultiATIS++ \citep{shankar-etal-2024-context}, instruction tuning for low-resource CSW scenarios \citep{lee-etal-2024-commit}, and synthetic data augmentation for sentiment analysis \citep{zeng-2024-leveraging}. 

\paragraph{\textit{Takeaway}} While prompting and retrieval enable rapid CSW adaptation, they struggle with informal mixing and low-resource pairs, highlighting that post-training flexibility cannot substitute for pre-training on diverse, naturalistic code-switched data \citep{bansal-etal-2022-pro, shankar-etal-2024-context}.

\section{Evaluation \& Benchmarking}
CSW benchmarks have evolved from narrow task evaluations to broader frameworks measuring switching patterns, cross-language performance, and contextual understanding. We review CSW benchmarks across text, speech, and multimodal tasks, with comprehensive details in Appendix (\S\ref{sec:appendix-benchmarks}) and Table~\ref{tab:cs-benchmarks-expanded} in Appendix \S\ref{sec:appendix-supp}. Evaluating CSW systems requires diverse metrics, encompassing standard performance measures, code-switching–specific metrics, and human-centric evaluation. Full descriptions of evaluation methods, and metrics are given in Appendix (\S\ref{sec:appendix-metrics}).

\section{Multi- \& Cross-Modal Applications}
\subsection{Speech Processing} 
Advances in recognition and multimodal integration have improved CSW speech processing, yet limited data availability continues to constrain performance across languages. \textbf{Speech translation} has advanced through end-to-end modeling for English–Spanish \citep{weller-etal-2022-end}, streaming Mandarin–English via self-training \citep{alastruey-etal-2023-towards}, and Whisper-based fine-tuning approaches such as CoVoSwitch \citep{kang-2024-covoswitch} and CoSTA \citep{p-s-v-n-etal-2025-costa}. \textbf{End-to-end ASR} research increasingly emphasizes adaptation over scale, using linguistic augmentation \citep{chi-bell-2022-improving}, monolingual fine-tuning that outperforms multilingual baselines \citep{babatunde-etal-2025-beyond}, retrieval-augmented refinement \citep{article}, and architectural innovations including attention-guided Whisper adaptation \citep{aditya2024attention}, mixture-of-experts models \citep{10890030}, and hybrid CTC/attention systems with language biasing \citep{liu2024enhancing}. Complementary signals from text-derived LID \citep{wang-li-2023-text} and semi-supervised learning \citep{biswas-etal-2020-semi} further mitigate data scarcity. Beyond audio-only pipelines, \textbf{audio-visual recognition} leverages visual cues to improve CSW ASR across African and Indian language pairs \citep{babatunde-etal-2025-beyond,HEMANT2025110408}, while data-centric strategies such as phrase-level mixing \citep{hussein2024speech} and zero-resource benchmarks \citep{huang2024zero} support robust evaluation.

Despite these advances, persistent challenges include high error rates at language switch points \citep{chi-bell-2022-improving}, limited generalization from synthetic data \citep{kugathasan-sumathipala-2021-neural}, and fine-tuning trade-offs in monolingual performance and computational cost \citep{babatunde-etal-2025-beyond}.

\subsection{Vision-Language Processing} 
Applied CSW research in real-world deployments remains limited, particularly for multimodal vision-language tasks. \textbf{Visual question answering} has advanced through knowledge distillation for Hinglish queries \citep{raj-khan-etal-2021-towards-developing}. \textbf{Multimodal systems} tackle challenges in harmful meme detection through visual-text fusion \citep{kumar2025multi, maity-etal-2024-toxvidlm}, while CLIP variants enable image-text retrieval in CSW settings \citep{Kumari2024CM_CLIPUC}. Collectively, these efforts highlight growing real-world CSW applications while underscoring the need for domain-, language-, and region-aware adaptation.



\subsection{Cross-Modal Integration} 
Beyond text-only modeling, cross-modal integration enables CSW systems to leverage phonetic, acoustic, and visual cues for robust multilingual understanding. \textbf{Phonetic modeling} supports discriminative language modeling \citep{winata-etal-2019-code}, transliteration and back-transliteration \citep{tasawong-etal-2023-typo, fernando-ranathunga-2021-data}, and tasks such as abusive content detection and translation alignment \citep{gautam-etal-2021-translate, chou-etal-2023-advancing}, with recent gains from transformer-based phonetic guidance and Wav2Vec2–GPT-2 fusion \citep{yang-tu-2022-combining, perera-sumanathilaka-2025-machine}. \textbf{Multimodal fusion} further improves code-mixed ASR and video-based toxicity detection by integrating audio, visual, and textual signals \citep{maity-etal-2024-toxvidlm, perera-sumanathilaka-2025-machine, 10890030}.

\paragraph{\textit{Takeaway}} \textit{Closing CSW performance gaps will require scalable, phonetic-aware multimodal pre-training, as approaches like Wav2Vec2 fusion already achieve 8–10\% ASR error reductions in high-resource switching scenarios.}

\section{Open Problems and Future Directions}


\paragraph{Data scarcity and quality issues} A key challenge in building CSW-friendly NLP systems is the lack of appropriate training data. The field remains heavily English-centric, with over 72\% of speech and 92\% of social media datasets involving English \citep{dogruoz-etal-2023-representativeness}, leaving non-English pairs underrepresented. Low-resource languages face \textit{higher computational costs} due to inefficient tokenization \citep{nag-etal-2024-cost}, while regional biases reduce generalization, models trained on one region often fail on the same language pair from another \citep{dogruoz-etal-2023-representativeness}.

\paragraph{Model Architecture and performance gaps} LLMs frequently exhibit \textit{language confusion}, generating responses in unintended languages, amplified by standard fine-tuning \citep{yoo-etal-2025-code-switching, marchisio-etal-2024-understanding}. Trained predominantly on \textit{monolingual text}, models remain ill-equipped for \textit{naturalistic CSW}. While LLMs excel at \textit{synthetic code-mixed data generation} \citep{pratapa-etal-2018-language, winata-etal-2019-code}, they show fragility in \textit{zero-shot transfer} with sharp accuracy drops \citep{zhang-etal-2023-multilingual, tatariya-etal-2023-transfer, 11016028}. (Refer to Table~\ref{tab:multimodal_failures} in Appendix \S\ref{sec:appendix-supp} for representative failures.)


\paragraph{Benchmarking and evaluation limitations} LLM-based evaluators often overestimate performance relative to human judgments, especially for low-resource and non–Latin-script languages \citep{hada-etal-2024-large}. Widely used metrics like BLEU and WER fail to capture the linguistic diversity which leads to poorly estimating the quality of code-mixed data \citep{srivastava-singh-2021-challenges}. Similarly, perplexity correlates poorly with both ASR performance and human judgments \citep{cheong-etal-2021-intrinsic, arora-etal-2023-comix, garg-etal-2021-mipe}. Standard semantic similarity metrics further struggle to model cross-lingual equivalence in mixed contexts \citep{maimaiti2025improving}. (Refer to Table~\ref{tab:eval_failures} in Appendix \S\ref{sec:appendix-supp} failure modes.) \\

\paragraph{Real-World Impact and Applications} Advances in CSW research unlock \textit{transformative applications} with significant societal impact: multilingual conversational assistants for accessible public services, cross-lingual educational platforms adaptive to learners' natural language practices, healthcare interfaces for multilingual populations, and digital preservation tools for endangered dialects. Such applications carry substantial social and economic value by reducing language barriers, democratizing access to information, and empowering multilingual communities within the digital economy. BanglAssist for customer service \citep{kruk2025banglassist} and code-switched dialogue agents for language learning \citep{brixey-traum-2025-code} illustrates this potential.

\paragraph{Language Bias and Trilingual Neglect} Current CSW research exhibits \textit{pronounced bias toward high-resource language pairs}, with the majority of studies focusing on English-Spanish, English-Hindi, and English-Mandarin combinations \citep{sitaram2019survey, winata-etal-2023-decades}. Performance metrics demonstrate significant disparities: English-Spanish and English-Hindi systems achieve mid-90s F1 scores, while less common pairs like Arabic-Egyptian Arabic or low-resource African language combinations show substantially lower accuracy  \citep{nguyen2021automatic}. African and Southeast Asian code-switched language pairs remain critically underexplored, with very few publicly available datasets \citep{terblanche-etal-2024-prompting}. Despite evidence that 7\% of India's population is trilingual, with over 250 million speakers engaging in multilingual discourse. Only isolated examples exist, such as SentMix (Bangla-English-Hindi trilingual dataset for NLI) and English-Hindi-Bengali Language Identification \citep{raihan-etal-2023-sentmix}, leaving a critical gap in understanding how models process switches across three or more languages. This bias perpetuates a cycle where \textit{resources and research investments concentrate on already well-studied pairs} \citep{terblanche-etal-2024-prompting, dogruoz-etal-2021-survey}, further marginalizing underrepresented multilingual communities and limiting the development of truly inclusive CSW technologies. \\

\noindent Additional challenges are discussed in Appendix (\S\ref{sec:appendix-future}).

\subsubsection*{Future Directions}

\paragraph{Toward Inclusive CSW Datasets} Progress in CSW NLP relies on expansive, inclusive datasets, yet \textit{large-scale conversational resources capturing naturalistic CSW interactions remain critically lacking}. Multimodal efforts like MEGAVERSE \citep{ahuja-etal-2024-megaverse} show promise but fall short in linguistic and domain diversity. SwitchLingua \citep{xie2025switchlingua}, while large and multilingual, relies on structured and synthesized text rather than fully natural conversational speech. CS‑FLEURS \citep{yan2025csfleursmassivelymultilingualcodeswitched} uses mostly synthetic or TTS-generated audio, limiting its ability to capture spontaneous CSW patterns. Multi-domain multilingual dialogue corpora \citep{moradshahi-etal-2023-x}, though broader in scope, highlight the need for future efforts to expand coverage, diversity, and naturalistic interactions.

\paragraph{Next-Generation Architectures} must \textit{jointly model text, speech, and vision} to enable switch-point detection, contextual understanding, and natural multilingual interactions, while ASR and TTS systems should leverage self-supervised encoders, cross-lingual, and emotion-aware conditioning. Promising directions include Speech-Conditioned LLMs combined  with MoE for ASR \citep{10890030} and curriculum learning strategies for multilingual transfer \citep{yoo-etal-2025-code-switching}. These approaches address phonemic confusion, data scarcity and the need for adaptive language mixing \citep{hamed-etal-2025-impact}.

\paragraph{Holistic Evaluation Paradigms} As CSW models become more multimodal and adaptive, evaluation must move \textit{beyond isolated task-level metrics toward human-aligned assessment} of multilingual competence. Future frameworks should jointly capture switch-point accuracy, semantic consistency, fluency, and \textit{sociolinguistic appropriateness}. While benchmarks such as CS-Sum \citep{suresh2025cs} and CodeMixBench \citep{yang-chai-2025-codemixbench} mark important progress \citep{hamed-etal-2025-impact}, evaluation must also account for regional and dialectal variation.

\paragraph{Ethics and Safety in Multilingual Contexts} Beyond performance and evaluation, future CSW systems must address \textit{critical ethical vulnerabilities} stemming from multilingual safety alignment gaps \citep{song-etal-2025-multilingual}, which disproportionately affect low-resource and marginalized language communities \citep{hamed-etal-2025-impact}. Safety evaluations reveal persistent failures under unseen language mixture patterns, as demonstrated by the \textit{Qorgau} framework in Kazakh–Russian settings \citep{goloburda-etal-2025-qorgau}. Addressing these challenges requires CSW-aware ethical AI that emphasizes inclusivity, transparency, and accountability through bias-aware training, fairness-sensitive evaluation, and \textit{participatory data curation with speaker communities}.

\section{Conclusion}

CSW research has undergone a major transformation with the rise of LLMs, evolving from task-specific statistical methods to unified multilingual and instruction-based frameworks. However, this survey shows that gains remain largely confined to high-resource language pairs, while LLMs struggle with spontaneous mixing, reasoning, and sociolinguistic variation in low-resource settings. These challenges are further amplified by limited dataset coverage and the lack of robust, CSW-aware evaluation frameworks. Meaningful progress in CSW NLP therefore requires moving beyond generic multilinguality toward targeted data curation, linguistically informed architectures, and evaluation protocols grounded in real-world language-mixing.

\newpage
\section*{Limitations}
Despite providing a broad survey, this paper has several limitations:
\begin{enumerate}
    \item \textbf{Coverage Bias} The survey highlights widely studied language pairs and might have missed indigenous or minority code-mixed languages.
    \item \textbf{Evolving Landscape} Given the rapid pace of LLM research, some approaches and benchmarks described may soon be outdated or replaced by newer paradigms.
    \item \textbf{Evaluation Constraints} While we include recent advances in speech and multimodal processing, the volume of research in these areas significantly lags behind text-based NLU, resulting in our taxonomy covering more text based NLU. 
    \item \textbf{Practical deployment} The survey mainly covers academic progress, leaving ethical, computational, and accessibility concerns in real-world deployment less examined.
    \item \textbf{Quantitative Reporting} All quantitative claims are representative values reported in the cited primary works rather than aggregated or meta-analyzed statistics.
\end{enumerate}

\section*{Ethics Statement}

This study involves a review and synthesis of previously published research and publicly available datasets. No human or user data were collected or analyzed. All works included in this survey were cited appropriately to acknowledge original authorship. The review process was conducted with transparency and fairness, avoiding selective reporting or biased interpretations.
Our study promotes fairness and inclusivity in multilingual NLP by focusing on underrepresented code-mixed language scenarios, encouraging equitable research attention toward linguistically diverse communities. 
The study adheres to established ethical standards for research in computational linguistics.

\section*{Acknowledgments}

This work is supported by the Anusandhan National Research Foundation (ANRF), India, through the project titled ``Curating and Constructing Benchmarks and Development of ML Models for Low-Level NLP Tasks in Hindi-English Code-Mixing''\footnote{More information about the Project is present here: \url{https://lingo.iitgn.ac.in/codemixing}}. The authors express their gratitude to Prathamesh Shanbhag, Naren Kumar S, Pranjal Yadav, Indrayudh Mandal, Sailesh Panda and Shruti Singh for their assistance in reveiwing the manuscript and providing feedback providing valuable feedback, and contributing to experimental evaluation and error analysis. Himanshu Beniwal is supported by the Prime Minister Research Fellowship (PMRF ID-1702154), India.

\bibliography{custom}

@article{poplack1988contrasting,
  title={Contrasting patterns of code-switching in two communities},
  author={Poplack, Shana},
  journal={Codeswitching: Anthropological and sociolinguistic perspectives},
  volume={48},
  pages={215--244},
  year={1988},
  publisher={Mouton de Gruyter Berlin \& New York},
url={https://yorkspace.library.yorku.ca/server/api/core/bitstreams/295cce20-53a9-4b98-85af-d0d5273e1e25/content}
}

@article{yan2025csfleursmassivelymultilingualcodeswitched,
      title={CS-FLEURS: A Massively Multilingual and Code-Switched Speech Dataset}, 
      author={Brian Yan and Injy Hamed and Shuichiro Shimizu and Vasista Lodagala and William Chen and Olga Iakovenko and Bashar Talafha and Amir Hussein and Alexander Polok and Kalvin Chang and Dominik Klement and Sara Althubaiti and Puyuan Peng and Matthew Wiesner and Thamar Solorio and Ahmed Ali and Sanjeev Khudanpur and Shinji Watanabe and Chih-Chen Chen and Zhen Wu and Karim Benharrak and Anuj Diwan and Samuele Cornell and Eunjung Yeo and Kwanghee Choi and Carlos Carvalho and Karen Rosero},
      year={2025},
      eprint={2509.14161},
      archivePrefix={arXiv},
      journal={arXiv preprint arXiv:2509.14161},
      primaryClass={cs.CL},
      url={https://arxiv.org/abs/250}}

@inproceedings{al-ali-aldarmaki-2024-mixat,
    title = "Mixat: A Data Set of Bilingual Emirati-{E}nglish Speech",
    author = "Al Ali, Maryam Khalifa  and
      Aldarmaki, Hanan",
    editor = "Melero, Maite  and
      Sakti, Sakriani  and
      Soria, Claudia",
    booktitle = "Proceedings of the 3rd Annual Meeting of the Special Interest Group on Under-resourced Languages @ LREC-COLING 2024",
    month = may,
    year = "2024",
    address = "Torino, Italia",
    publisher = "ELRA and ICCL",
    url = "https://aclanthology.org/2024.sigul-1.26/",
    pages = "222--226"}

@inproceedings{lyu10_interspeech,
  title     = "{SEAME}: A Mandarin-English Code-switching Speech Corpus in South-East Asia",
  author    = "Lyu, Dau-Cheng and Tan, Tien-Ping and Chng, Eng-Siong and Li, Haizhou",
  booktitle = "Interspeech 2010",
  year      = "2010",
  pages     = "1986--1989",
  address   = "Makuhari, Chiba, Japan",
  publisher = "ISCA",
  url       = "https://www.isca-archive.org/interspeech_2010/lyu10_interspeech.pdf"
}

@inproceedings{li2022talcs,
  title     = "{TALCS}: An Open-Source Mandarin-English Code-Switching Corpus and a Speech Recognition Baseline",
  author    = "Li, Chengfei and Deng, Shuhao and Wang, Yaoping and Wang, Guangjing and Gong, Yaguang and others",
  booktitle = "Interspeech 2022",
  year      = "2022",
  pages     = "1--5",
  publisher = "ISCA",
  doi       = "10.21437/Interspeech.2022-1339",
  url       = "https://arxiv.org/abs/2206.13135"
}

@inproceedings{hamed-etal-2020-arzen,
  title     = "{A}rz{E}n: A Speech Corpus for Code-switched Egyptian Arabic-English",
  author    = "Hamed, Injy and Vu, Ngoc Thang and Abdennadher, Slim",
  booktitle = "Proceedings of the 12th Language Resources and Evaluation Conference (LREC)",
  month     = may,
  year      = "2020",
  address   = "Marseille, France",
  publisher = "European Language Resources Association",
  pages     = "4276--4284",
  url       = "https://aclanthology.org/2020.lrec-1.526"
}

@inproceedings{diwan2021mucs,
  title     = "{MUCS} 2021: Multilingual and Code-Switching {ASR} Challenges for Low Resource Indian Languages",
  author    = "Diwan, Anuj and Vaideeswaran, Rakesh and Shah, Sanket and Singh, Ankita and Raghavan, Srinivasa and others",
  booktitle = "Interspeech 2021",
  year      = "2021",
  pages     = "2446--2450",
  publisher = "ISCA",
  doi       = "10.21437/Interspeech.2021-1339",
  url       = "https://www.isca-archive.org/interspeech_2021/diwan21_interspeech.html"
}

@inproceedings{chouli2021escwa,
  title     = "Arabic Code-Switching Speech Recognition Using Monolingual Data",
  author    = "Ali, Ahmed and Chowdhury, Shammur Absar and Hussein, Amir and Hifny, Yasser",
  booktitle = "Interspeech 2021",
  year      = "2021",
  pages     = "3475--3479",
  publisher = "ISCA",
  note      = "Releases the ESCWA corpus",
  doi       = "10.21437/Interspeech.2021-1365",
  url       = "https://www.isca-archive.org/interspeech_2021/ali21b_interspeech.pdf"
}

@inproceedings{alharbi2024leveraging,
  title     = "Leveraging {LLM} for Augmenting Textual Data in Code-Switching {ASR}",
  author    = "Alharbi, Haneen and Khurana, Sameer and Glass, James and Ali, Ahmed",
  booktitle = "Proceedings of the 1st Workshop on Synthetic Data for Generative AI (SynData4GenAI)",
  year      = "2024",
  publisher = "ISCA",
  note      = "Introduces the Saudilang Code-switch Corpus (SCC)",
  url       = "https://www.isca-archive.org/syndata4genai_2024/alharbi24_syndata4genai.pdf"
}

@article{singh2025hiacc,
  title={HiACC: Hinglish adult \& children code-switched corpus},
  author={Singh, Shruti and Singh, Muskaan and Kadyan, Virender},
  journal={Data in Brief},
  pages={111886},
  year={2025},
  publisher={Elsevier},
url={https://doi.org/10.1016/j.dib.2025.111886}
}

@article{ostapenko2022speaker,
  title={Speaker information can guide models to better inductive biases: A case study on predicting code-switching},
  author={Ostapenko, Alissa and Wintner, Shuly and Fricke, Melinda and Tsvetkov, Yulia},
  journal={arXiv preprint arXiv:2203.08979},
  year={2022}
}

@inproceedings{solorio-etal-2014-overview,
    title = "Overview for the First Shared Task on Language Identification in Code-Switched Data",
    author = "Solorio, Thamar  and
      Blair, Elizabeth  and
      Maharjan, Suraj  and
      Bethard, Steven  and
      Diab, Mona  and
      Ghoneim, Mahmoud  and
      Hawwari, Abdelati  and
      AlGhamdi, Fahad  and
      Hirschberg, Julia  and
      Chang, Alison  and
      Fung, Pascale",
    editor = "Diab, Mona  and
      Hirschberg, Julia  and
      Fung, Pascale  and
      Solorio, Thamar",
    booktitle = "Proceedings of the First Workshop on Computational Approaches to Code Switching",
    month = oct,
    year = "2014",
    address = "Doha, Qatar",
    publisher = "Association for Computational Linguistics",
    url = "https://aclanthology.org/W14-3907/",
    doi = "10.3115/v1/W14-3907",
    pages = "62--72"
}

@inproceedings{molina-etal-2016-overview,
    title = "Overview for the Second Shared Task on Language Identification in Code-Switched Data",
    author = "Molina, Giovanni  and
      AlGhamdi, Fahad  and
      Ghoneim, Mahmoud  and
      Hawwari, Abdelati  and
      Rey-Villamizar, Nicolas  and
      Diab, Mona  and
      Solorio, Thamar",
    booktitle = "Proceedings of the Second Workshop on Computational Approaches to Code Switching",
    month = nov,
    year = "2016",
    address = "Austin, Texas",
    publisher = "Association for Computational Linguistics",
    url = "https://aclanthology.org/W16-5805/",
    doi = "10.18653/v1/W16-5805",
    pages = "40--49"
}

@inproceedings{singh2017towards,
  title={Towards translating mixed-code comments from social media},
  author={Singh, Thoudam Doren and Solorio, Thamar},
  booktitle={International Conference on Computational Linguistics and Intelligent Text Processing},
  pages={457--468},
  year={2017},
  organization={Springer},
url={https://doi.org/10.1007/978-3-319-77116-8_34}
}

@inproceedings{santy-etal-2021-bertologicomix,
    title = "{BERT}ologi{C}o{M}ix: How does Code-Mixing interact with Multilingual {BERT}?",
    author = "Santy, Sebastin  and
      Srinivasan, Anirudh  and
      Choudhury, Monojit",
    booktitle = "Proceedings of the Second Workshop on Domain Adaptation for NLP",
    month = apr,
    year = "2021",
    address = "Kyiv, Ukraine",
    publisher = "Association for Computational Linguistics",
    url = "https://aclanthology.org/2021.adaptnlp-1.12/",
    pages = "111--121",
}

@inproceedings{olaleye-etal-2025-afrocs,
    title = "{A}fro{CS}-xs: Creating a Compact, High-Quality, Human-Validated Code-Switched Dataset for {A}frican Languages",
    author = "Olaleye, Kayode  and
      Oncevay, Arturo  and
      Sibue, Mathieu  and
      Zondi, Nombuyiselo  and
      Terblanche, Michelle  and
      Mapikitla, Sibongile  and
      Lastrucci, Richard  and
      Smiley, Charese  and
      Marivate, Vukosi",
    booktitle = "Proceedings of the 63rd Annual Meeting of the Association for Computational Linguistics (Volume 1: Long Papers)",
    month = jul,
    year = "2025",
    address = "Vienna, Austria",
    publisher = "Association for Computational Linguistics",
    url = "https://aclanthology.org/2025.acl-long.1601/",
    doi = "10.18653/v1/2025.acl-long.1601",
    pages = "33391--33410",
    ISBN = "979-8-89176-251-0",

}

@inproceedings{goswami-etal-2023-offmix,
    title = "{O}ff{M}ix-3{L}: A Novel Code-Mixed Test Dataset in {B}angla-{E}nglish-{H}indi for Offensive Language Identification",
    author = "Goswami, Dhiman  and
      Raihan, Md Nishat  and
      Mahmud, Antara  and
      Anastasopoulos, Antonios  and
      Zampieri, Marcos",
    booktitle = "Proceedings of the 11th International Workshop on Natural Language Processing for Social Media",
    month = nov,
    year = "2023",
    address = "Bali, Indonesia",
    publisher = "Association for Computational Linguistics",
    url = "https://aclanthology.org/2023.socialnlp-1.3/",
    doi = "10.18653/v1/2023.socialnlp-1.3",
    pages = "21--27"
}

@inproceedings{shankar-etal-2024-context,
    title = "In-context Mixing ({ICM}): Code-mixed Prompts for Multilingual {LLM}s",
    author = "Shankar, Bhavani  and
      Jyothi, Preethi  and
      Bhattacharyya, Pushpak",
    booktitle = "Proceedings of the 62nd Annual Meeting of the Association for Computational Linguistics (Volume 1: Long Papers)",
    month = aug,
    year = "2024",
    address = "Bangkok, Thailand",
    publisher = "Association for Computational Linguistics",
    url = "https://aclanthology.org/2024.acl-long.228/",
    doi = "10.18653/v1/2024.acl-long.228",
    pages = "4162--4176",
}

@inproceedings{huzaifah-etal-2024-evaluating,
    title = "Evaluating Code-Switching Translation with Large Language Models",
    author = "Huzaifah, Muhammad  and
      Zheng, Weihua  and
      Chanpaisit, Nattapol  and
      Wu, Kui",
    booktitle = "Proceedings of the 2024 Joint International Conference on Computational Linguistics, Language Resources and Evaluation (LREC-COLING 2024)",
    month = may,
    year = "2024",
    address = "Torino, Italia",
    publisher = "ELRA and ICCL",
    url = "https://aclanthology.org/2024.lrec-main.565/",
    pages = "6381--6394",
}

@inproceedings{lee-etal-2024-commit,
    title = "{COMMIT}: Code-Mixing {E}nglish-Centric Large Language Model for Multilingual Instruction Tuning",
    author = "Lee, Jaeseong  and
      Jung, YeonJoon  and
      Hwang, Seung-won",
    editor = "Duh, Kevin  and
      Gomez, Helena  and
      Bethard, Steven",
    booktitle = "Findings of the Association for Computational Linguistics: NAACL 2024",
    month = jun,
    year = "2024",
    address = "Mexico City, Mexico",
    publisher = "Association for Computational Linguistics",
    url = "https://aclanthology.org/2024.findings-naacl.198/",
    doi = "10.18653/v1/2024.findings-naacl.198",
    pages = "3130--3137"
}

@inproceedings{zeng-2024-leveraging,
    title = "Leveraging Large Language Models for Code-Mixed Data Augmentation in Sentiment Analysis",
    author = "Zeng, Linda",
    editor = "Hale, James  and
      Chawla, Kushal  and
      Garg, Muskan",
    booktitle = "Proceedings of the Second Workshop on Social Influence in Conversations (SICon 2024)",
    month = nov,
    year = "2024",
    address = "Miami, Florida, USA",
    publisher = "Association for Computational Linguistics",
    url = "https://aclanthology.org/2024.sicon-1.6/",
    doi = "10.18653/v1/2024.sicon-1.6",
    pages = "85--101",
}

@inproceedings{jasim-etal-2020-phraseout,
    title = "{P}hrase{O}ut: A Code Mixed Data Augmentation Method for {M}ultilingual{N}eural Machine Tranlsation",
    author = "Jasim, Binu  and
      Namboodiri, Vinay  and
      Jawahar, C V",
    editor = "Bhattacharyya, Pushpak  and
      Sharma, Dipti Misra  and
      Sangal, Rajeev",
    booktitle = "Proceedings of the 17th International Conference on Natural Language Processing (ICON)",
    month = dec,
    year = "2020",
    address = "Indian Institute of Technology Patna, Patna, India",
    publisher = "NLP Association of India (NLPAI)",
    url = "https://aclanthology.org/2020.icon-main.63/",
    pages = "470--474",
}

@inproceedings{yadav-2026-competence,
    title = "Competence Collapse in Code-Mixed Generation: Spectral Evidence and Mechanistic Recovery via Cross-Lingual Activation Steering",
    author = "Yadav, Tanushree Ravindra Pratap",
    editor = "Hettiarachchi, Hansi  and
      Ranasinghe, Tharindu  and
      Plum, Alistair  and
      Rayson, Paul  and
      Mitkov, Ruslan  and
      Gaber, Mohamed  and
      Premasiri, Damith  and
      Tan, Fiona Anting  and
      Uyangodage, Lasitha",
    booktitle = "Proceedings of the Second Workshop on Language Models for Low-Resource Languages ({L}o{R}es{LM} 2026)",
    month = mar,
    year = "2026",
    address = "Rabat, Morocco",
    publisher = "Association for Computational Linguistics",
    url = "https://aclanthology.org/2026.loreslm-1.3/",
    doi = "10.18653/v1/2026.loreslm-1.3",
    pages = "29--40",
    ISBN = "979-8-89176-377-7",
}

@inproceedings{winata-etal-2023-decades,
    title = "The Decades Progress on Code-Switching Research in {NLP}: A Systematic Survey on Trends and Challenges",
    author = "Winata, Genta  and
      Aji, Alham Fikri  and
      Yong, Zheng Xin  and
      Solorio, Thamar",
    booktitle = "Findings of the Association for Computational Linguistics: ACL 2023",
    month = jul,
    year = "2023",
    address = "Toronto, Canada",
    publisher = "Association for Computational Linguistics",
    url = "https://aclanthology.org/2023.findings-acl.185/",
    doi = "10.18653/v1/2023.findings-acl.185",
    pages = "2936--2978",

}

@inproceedings{kodali-etal-2022-symcom,
    title = "{S}y{MC}o{M} - Syntactic Measure of Code Mixing A Study Of {E}nglish-{H}indi Code-Mixing",
    author = "Kodali, Prashant  and
      Goel, Anmol  and
      Choudhury, Monojit  and
      Shrivastava, Manish  and
      Kumaraguru, Ponnurangam",
    editor = "Muresan, Smaranda  and
      Nakov, Preslav  and
      Villavicencio, Aline",
    booktitle = "Findings of the Association for Computational Linguistics: ACL 2022",
    month = may,
    year = "2022",
    address = "Dublin, Ireland",
    publisher = "Association for Computational Linguistics",
    url = "https://aclanthology.org/2022.findings-acl.40/",
    doi = "10.18653/v1/2022.findings-acl.40",
    pages = "472--480",

}

@article{Shekhar2020LanguageIF,
  title={Language identification framework in code-mixed social media text based on quantum LSTM — the word belongs to which language?},
  author={Shashi Shekhar and Dilip Kumar Sharma and Mirza Mohd. Sufyan Beg},
  journal={Modern Physics Letters B},
  year={2020},
  volume={34},
  pages={2050086},
  url={https://api.semanticscholar.org/CorpusID:214459891}
}

@inproceedings{gundapu-mamidi-2018-word,
    title = "Word Level Language Identification in {E}nglish {T}elugu Code Mixed Data",
    author = "Gundapu, Sunil  and
      Mamidi, Radhika",
    editor = "Politzer-Ahles, Stephen  and
      Hsu, Yu-Yin  and
      Huang, Chu-Ren  and
      Yao, Yao",
    booktitle = "Proceedings of the 32nd Pacific Asia Conference on Language, Information and Computation",
    month = "1–3 " # dec,
    year = "2018",
    address = "Hong Kong",
    publisher = "Association for Computational Linguistics",
    url = "https://aclanthology.org/Y18-1021/"
}

@inproceedings{raha-etal-2019-development,
    title = "Development of {POS} tagger for {E}nglish-{B}engali Code-Mixed data",
    author = "Raha, Tathagata  and
      Mahata, Sainik  and
      Das, Dipankar  and
      Bandyopadhyay, Sivaji",
    editor = "Sharma, Dipti Misra  and
      Bhattacharya, Pushpak",
    booktitle = "Proceedings of the 16th International Conference on Natural Language Processing",
    month = dec,
    year = "2019",
    address = "International Institute of Information Technology, Hyderabad, India",
    publisher = "NLP Association of India",
    url = "https://aclanthology.org/2019.icon-1.17/",
    pages = "143--149",

}

@inproceedings{pratapa-etal-2018-word,
    title = "Word Embeddings for Code-Mixed Language Processing",
    author = "Pratapa, Adithya  and
      Choudhury, Monojit  and
      Sitaram, Sunayana",
    booktitle = "Proceedings of the 2018 Conference on Empirical Methods in Natural Language Processing",
    month = oct # "-" # nov,
    year = "2018",
    address = "Brussels, Belgium",
    publisher = "Association for Computational Linguistics",
    url = "https://aclanthology.org/D18-1344/",
    doi = "10.18653/v1/D18-1344",
    pages = "3067--3072",
}

@inproceedings{sequiera-etal-2015-pos,
    title = "{POS} Tagging of {H}indi-{E}nglish Code Mixed Text from Social Media: Some Machine Learning Experiments",
    author = "Sequiera, Royal  and
      Choudhury, Monojit  and
      Bali, Kalika",
    editor = "Sharma, Dipti Misra  and
      Sangal, Rajeev  and
      Sherly, Elizabeth",
    booktitle = "Proceedings of the 12th International Conference on Natural Language Processing",
    month = dec,
    year = "2015",
    address = "Trivandrum, India",
    publisher = "NLP Association of India",
    url = "https://aclanthology.org/W15-5936/",
    pages = "237--246"
}

@inproceedings{chopra-etal-2021-switch-point,
    title = "Switch Point biased Self-Training: Re-purposing Pretrained Models for Code-Switching",
    author = "Chopra, Parul  and
      Rallabandi, Sai Krishna  and
      Black, Alan W  and
      Chandu, Khyathi Raghavi",
    booktitle = "Findings of the Association for Computational Linguistics: EMNLP 2021",
    month = nov,
    year = "2021",
    address = "Punta Cana, Dominican Republic",
    publisher = "Association for Computational Linguistics",
    url = "https://aclanthology.org/2021.findings-emnlp.373/",
    doi = "10.18653/v1/2021.findings-emnlp.373",
    pages = "4389--4397",

}

@misc{ansari2019cross,
 title={Cross Script Hindi English NER Corpus from Wikipedia}, 
      author={Mohd Zeeshan Ansari and Tanvir Ahmad and Md Arshad Ali},
      year={2018},
      eprint={1810.03430},
      archivePrefix={arXiv},
      primaryClass={cs.IR},
      url={https://arxiv.org/abs/1810.03430}, 
}

@inproceedings{solorio-liu-2008-learning,
    title = "Learning to Predict Code-Switching Points",
    author = "Solorio, Thamar  and
      Liu, Yang",
    editor = "Lapata, Mirella  and
      Ng, Hwee Tou",
    booktitle = "Proceedings of the 2008 Conference on Empirical Methods in Natural Language Processing",
    month = oct,
    year = "2008",
    address = "Honolulu, Hawaii",
    publisher = "Association for Computational Linguistics",
    url = "https://aclanthology.org/D08-1102/",
    pages = "973--981"
}

@inproceedings{king-etal-2014-iucl,
    title = "The {IUCL}+ System: Word-Level Language Identification via Extended {M}arkov Models",
    author = {King, Levi  and
      Baucom, Eric  and
      Gilmanov, Timur  and
      K{\"u}bler, Sandra  and
      Whyatt, Dan  and
      Maier, Wolfgang  and
      Rodrigues, Paul},
    editor = "Diab, Mona  and
      Hirschberg, Julia  and
      Fung, Pascale  and
      Solorio, Thamar",
    booktitle = "Proceedings of the First Workshop on Computational Approaches to Code Switching",
    month = oct,
    year = "2014",
    address = "Doha, Qatar",
    publisher = "Association for Computational Linguistics",
    url = "https://aclanthology.org/W14-3912/",
    doi = "10.3115/v1/W14-3912",
    pages = "102--106"
}

@inproceedings{chittaranjan-etal-2014-word,
    title = "Word-level Language Identification using {CRF}: Code-switching Shared Task Report of {MSR} {I}ndia System",
    author = "Chittaranjan, Gokul  and
      Vyas, Yogarshi  and
      Bali, Kalika  and
      Choudhury, Monojit",
    editor = "Diab, Mona  and
      Hirschberg, Julia  and
      Fung, Pascale  and
      Solorio, Thamar",
    booktitle = "Proceedings of the First Workshop on Computational Approaches to Code Switching",
    month = oct,
    year = "2014",
    address = "Doha, Qatar",
    publisher = "Association for Computational Linguistics",
    url = "https://aclanthology.org/W14-3908/",
    doi = "10.3115/v1/W14-3908",
    pages = "73--79"
}

@inproceedings{patwa-etal-2020-semeval,
    title = "{S}em{E}val-2020 Task 9: Overview of Sentiment Analysis of Code-Mixed Tweets",
    author = {Patwa, Parth  and
      Aguilar, Gustavo  and
      Kar, Sudipta  and
      Pandey, Suraj  and
      PYKL, Srinivas  and
      Gamb{\"a}ck, Bj{\"o}rn  and
      Chakraborty, Tanmoy  and
      Solorio, Thamar  and
      Das, Amitava},
    editor = "Herbelot, Aurelie  and
      Zhu, Xiaodan  and
      Palmer, Alexis  and
      Schneider, Nathan  and
      May, Jonathan  and
      Shutova, Ekaterina",
    booktitle = "Proceedings of the Fourteenth Workshop on Semantic Evaluation",
    month = dec,
    year = "2020",
    address = "Barcelona (online)",
    publisher = "International Committee for Computational Linguistics",
    url = "https://aclanthology.org/2020.semeval-1.100/",
    doi = "10.18653/v1/2020.semeval-1.100",
    pages = "774--790",
}

@inproceedings{joshi-etal-2016-towards,
    title = "Towards Sub-Word Level Compositions for Sentiment Analysis of {H}indi-{E}nglish Code Mixed Text",
    author = "Joshi, Aditya  and
      Prabhu, Ameya  and
      Shrivastava, Manish  and
      Varma, Vasudeva",
    editor = "Matsumoto, Yuji  and
      Prasad, Rashmi",
    booktitle = "Proceedings of {COLING} 2016, the 26th International Conference on Computational Linguistics: Technical Papers",
    month = dec,
    year = "2016",
    address = "Osaka, Japan",
    publisher = "The COLING 2016 Organizing Committee",
    url = "https://aclanthology.org/C16-1234/",
    pages = "2482--2491",
}

@inproceedings{singh-etal-2018-named,
    title = "Named Entity Recognition for {H}indi-{E}nglish Code-Mixed Social Media Text",
    author = "Singh, Vinay  and
      Vijay, Deepanshu  and
      Akhtar, Syed Sarfaraz  and
      Shrivastava, Manish",
    editor = "Chen, Nancy  and
      Banchs, Rafael E.  and
      Duan, Xiangyu  and
      Zhang, Min  and
      Li, Haizhou",
    booktitle = "Proceedings of the Seventh Named Entities Workshop",
    month = jul,
    year = "2018",
    address = "Melbourne, Australia",
    publisher = "Association for Computational Linguistics",
    url = "https://aclanthology.org/W18-2405/",
    doi = "10.18653/v1/W18-2405",
    pages = "27--35",
}

@inproceedings{vyas-etal-2014-pos,
    title = "{POS} Tagging of {E}nglish-{H}indi Code-Mixed Social Media Content",
    author = "Vyas, Yogarshi  and
      Gella, Spandana  and
      Sharma, Jatin  and
      Bali, Kalika  and
      Choudhury, Monojit",
    booktitle = "Proceedings of the 2014 Conference on Empirical Methods in Natural Language Processing ({EMNLP})",
    month = oct,
    year = "2014",
    address = "Doha, Qatar",
    publisher = "Association for Computational Linguistics",
    url = "https://aclanthology.org/D14-1105/",
    doi = "10.3115/v1/D14-1105",
    pages = "974--979"
}

@inproceedings{winata-etal-2019-code,
    title = "Code-Switched Language Models Using Neural Based Synthetic Data from Parallel Sentences",
    author = "Winata, Genta Indra  and
      Madotto, Andrea  and
      Wu, Chien-Sheng  and
      Fung, Pascale",
    editor = "Bansal, Mohit  and
      Villavicencio, Aline",
    booktitle = "Proceedings of the 23rd Conference on Computational Natural Language Learning (CoNLL)",
    month = nov,
    year = "2019",
    address = "Hong Kong, China",
    publisher = "Association for Computational Linguistics",
    url = "https://aclanthology.org/K19-1026/",
    doi = "10.18653/v1/K19-1026",
    pages = "271--280",
}

@article{Issa2025DelvingIB,
  title={Delving Into Bilingual Dialogue: The Realm of Code Switching and Mixing in Arabic-English Societies},
  author={Saddam H.M. Issa and Fatima Amer Aldakhil and Amani Abdullah BinJwair and Nizaam Kariem},
  journal={Journal of Language Teaching and Research},
  year={2025},
  url={https://api.semanticscholar.org/CorpusID:278279801}
}

@inproceedings{kong2025decoding,
  title = "Decoding Machine Translationese in {E}nglish-{C}hinese News: {LLM}s vs. {NMT}s",
    author = "Kong, Delu  and
      Macken, Lieve",
    editor = "Bouillon, Pierrette  and
      Gerlach, Johanna  and
      Girletti, Sabrina  and
      Volkart, Lise  and
      Rubino, Raphael  and
      Sennrich, Rico  and
      Farinha, Ana C.  and
      Gaido, Marco  and
      Daems, Joke  and
      Kenny, Dorothy  and
      Moniz, Helena  and
      Szoc, Sara",
    booktitle = "Proceedings of Machine Translation Summit XX: Volume 1",
    month = jun,
    year = "2025",
    address = "Geneva, Switzerland",
    publisher = "European Association for Machine Translation",
    url = "https://aclanthology.org/2025.mtsummit-1.8/",
    pages = "99--112",
    ISBN = "978-2-9701897-0-1"
}

@article{kinney2023semantic,
  title={The semantic scholar open data platform},
  author={Kinney, Rodney and Anastasiades, Chloe and Authur, Russell and Beltagy, Iz and Bragg, Jonathan and Buraczynski, Alexandra and Cachola, Isabel and Candra, Stefan and Chandrasekhar, Yoganand and Cohan, Arman and others},
  journal={arXiv preprint arXiv:2301.10140},
  year={2023}
}

@inproceedings{sarrof2025homophonic,
title = "Homophonic Pun Generation in Code Mixed {H}indi {E}nglish",
    author = "Sarrof, Yash Raj",
    editor = "Hempelmann, Christian F.  and
      Rayz, Julia  and
      Dong, Tiansi  and
      Miller, Tristan",
    booktitle = "Proceedings of the 1st Workshop on Computational Humor (CHum)",
    month = jan,
    year = "2025",
    address = "Online",
    publisher = "Association for Computational Linguistics",
    url = "https://aclanthology.org/2025.chum-1.4/",
    pages = "23--31"
}

@inproceedings{Yoo2024CodeSwitchingRL,
  title={Code-Switching Red-Teaming: LLM Evaluation for Safety and Multilingual Understanding},
  author={Haneul Yoo and Yongjin Yang and Hwaran Lee},
  booktitle={Annual Meeting of the Association for Computational Linguistics},
  year={2024},
  url={https://api.semanticscholar.org/CorpusID:270702992}
}

@article{jehan2025evolution,
  title={The Evolution of Code-Switching in Multilingual Societies: A Sociolinguistic Perspective: https://doi. org/10.55966/assaj. 2025.4. 1.054},
  author={Jehan, Noor and Javed, Tabassum and Banu, Shahida},
  journal={ASSAJ},
  volume={4},
  number={01},
  pages={614--625},
  year={2025}
}

@inproceedings{upadhayay-behzadan-2025-tongue,
    title = "Tongue-Tied: Breaking {LLM}s Safety Through New Language Learning",
    author = "Upadhayay, Bibek  and
      Behzadan, Vahid",
    editor = "Winata, Genta Indra  and
      Kar, Sudipta  and
      Zhukova, Marina  and
      Solorio, Thamar  and
      Ai, Xi  and
      Hamed, Injy  and
      Ihsani, Mahardika Krisna Krisna  and
      Wijaya, Derry Tanti  and
      Kuwanto, Garry",
    booktitle = "Proceedings of the 7th Workshop on Computational Approaches to Linguistic Code-Switching",
    month = may,
    year = "2025",
    address = "Albuquerque, New Mexico, USA",
    publisher = "Association for Computational Linguistics",
    url = "https://aclanthology.org/2025.calcs-1.5/",
    doi = "10.18653/v1/2025.calcs-1.5",
    pages = "32--47",
    ISBN = "979-8-89176-053-0",
}

@inproceedings{marchisio-etal-2024-understanding,
    title = "Understanding and Mitigating Language Confusion in {LLM}s",
    author = "Marchisio, Kelly  and
      Ko, Wei-Yin  and
      Berard, Alexandre  and
      Dehaze, Th{\'e}o  and
      Ruder, Sebastian",
    editor = "Al-Onaizan, Yaser  and
      Bansal, Mohit  and
      Chen, Yun-Nung",
    booktitle = "Proceedings of the 2024 Conference on Empirical Methods in Natural Language Processing",
    month = nov,
    year = "2024",
    address = "Miami, Florida, USA",
    publisher = "Association for Computational Linguistics",
    url = "https://aclanthology.org/2024.emnlp-main.380/",
    doi = "10.18653/v1/2024.emnlp-main.380",
    pages = "6653--6677",
}

@inproceedings{yang-chai-2025-codemixbench,
    title = "{C}ode{M}ix{B}ench: Evaluating Code-Mixing Capabilities of {LLM}s Across 18 Languages",
    author = "Yang, Yilun  and
      Chai, Yekun",
    editor = "Christodoulopoulos, Christos  and
      Chakraborty, Tanmoy  and
      Rose, Carolyn  and
      Peng, Violet",
    booktitle = "Proceedings of the 2025 Conference on Empirical Methods in Natural Language Processing",
    month = nov,
    year = "2025",
    address = "Suzhou, China",
    publisher = "Association for Computational Linguistics",
    url = "https://aclanthology.org/2025.emnlp-main.109/",
    doi = "10.18653/v1/2025.emnlp-main.109",
    pages = "2139--2169",
    ISBN = "979-8-89176-332-6",
}

@article{gupta2024multilingual,
  title={Multilingual controlled generation and gold-standard-agnostic evaluation of code-mixed sentences},
  author={Gupta, Ayushman and Bhogal, Akhil and Ghosh, Kripabandhu},
  journal={arXiv preprint arXiv:2410.10580},
  year={2024}
}

@inproceedings{sterner-teufel-2025-minimal,
    title = "Minimal Pair-Based Evaluation of Code-Switching",
    author = "Sterner, Igor  and
      Teufel, Simone",
    editor = "Che, Wanxiang  and
      Nabende, Joyce  and
      Shutova, Ekaterina  and
      Pilehvar, Mohammad Taher",
    booktitle = "Proceedings of the 63rd Annual Meeting of the Association for Computational Linguistics (Volume 1: Long Papers)",
    month = jul,
    year = "2025",
    address = "Vienna, Austria",
    publisher = "Association for Computational Linguistics",
    url = "https://aclanthology.org/2025.acl-long.910/",
    doi = "10.18653/v1/2025.acl-long.910",
    pages = "18575--18598",
    ISBN = "979-8-89176-251-0",

}

@inproceedings{krishnan-etal-2021-multilingual,
    title = "Multilingual Code-Switching for Zero-Shot Cross-Lingual Intent Prediction and Slot Filling",
    author = "Krishnan, Jitin  and
      Anastasopoulos, Antonios  and
      Purohit, Hemant  and
      Rangwala, Huzefa",
    booktitle = "Proceedings of the 1st Workshop on Multilingual Representation Learning",
    month = nov,
    year = "2021",
    address = "Punta Cana, Dominican Republic",
    publisher = "Association for Computational Linguistics",
    url = "https://aclanthology.org/2021.mrl-1.18/",
    doi = "10.18653/v1/2021.mrl-1.18",
    pages = "211--223",
}

@inproceedings{dogruoz-etal-2023-representativeness,
    title = "Representativeness as a Forgotten Lesson for Multilingual and Code-switched Data Collection and Preparation",
    author = {Do{\u{g}}ru{\"o}z, A. Seza  and
      Sitaram, Sunayana  and
      Yong, Zheng Xin},
    editor = "Bouamor, Houda  and
      Pino, Juan  and
      Bali, Kalika",
    booktitle = "Findings of the Association for Computational Linguistics: EMNLP 2023",
    month = dec,
    year = "2023",
    address = "Singapore",
    publisher = "Association for Computational Linguistics",
    url = "https://aclanthology.org/2023.findings-emnlp.382/",
    doi = "10.18653/v1/2023.findings-emnlp.382",
    pages = "5751--5767",
}

@inproceedings{nag-etal-2024-cost,
    title = "Cost-Performance Optimization for Processing Low-Resource Language Tasks Using Commercial {LLM}s",
    author = "Nag, Arijit  and
      Mukherjee, Animesh  and
      Ganguly, Niloy  and
      Chakrabarti, Soumen",
    editor = "Al-Onaizan, Yaser  and
      Bansal, Mohit  and
      Chen, Yun-Nung",
    booktitle = "Findings of the Association for Computational Linguistics: EMNLP 2024",
    month = nov,
    year = "2024",
    address = "Miami, Florida, USA",
    publisher = "Association for Computational Linguistics",
    url = "https://aclanthology.org/2024.findings-emnlp.920/",
    doi = "10.18653/v1/2024.findings-emnlp.920",
    pages = "15681--15701",
}

@inproceedings{sazzed-2021-abusive,
    title = "Abusive content detection in transliterated {B}engali-{E}nglish social media corpus",
    author = "Sazzed, Salim",
    booktitle = "Proceedings of the Fifth Workshop on Computational Approaches to Linguistic Code-Switching",
    month = jun,
    year = "2021",
    address = "Online",
    publisher = "Association for Computational Linguistics",
    url = "https://aclanthology.org/2021.calcs-1.16/",
    doi = "10.18653/v1/2021.calcs-1.16",
    pages = "125--130",
}

@inproceedings{raihan-etal-2023-sentmix,
    title = "{S}ent{M}ix-3{L}: A Novel Code-Mixed Test Dataset in {B}angla-{E}nglish-{H}indi for Sentiment Analysis",
    author = "Raihan, Md Nishat  and
      Goswami, Dhiman  and
      Mahmud, Antara  and
      Anastasopoulos, Antonios  and
      Zampieri, Marcos",
    editor = "Wijaya, Derry  and
      Aji, Alham Fikri  and
      Vania, Clara  and
      Winata, Genta Indra  and
      Purwarianti, Ayu",
    booktitle = "Proceedings of the First Workshop in South East Asian Language Processing",
    month = nov,
    year = "2023",
    address = "Nusa Dua, Bali, Indonesia",
    publisher = "Association for Computational Linguistics",
    url = "https://aclanthology.org/2023.sealp-1.6/",
    doi = "10.18653/v1/2023.sealp-1.6",
    pages = "79--84"
}

@inproceedings{kargaran-etal-2024-masklid,
    title = "{M}ask{LID}: Code-Switching Language Identification through Iterative Masking",
    author = "Kargaran, Amir Hossein  and
      Yvon, Fran{\c{c}}ois  and
      Schuetze, Hinrich",
    editor = "Ku, Lun-Wei  and
      Martins, Andre  and
      Srikumar, Vivek",
    booktitle = "Proceedings of the 62nd Annual Meeting of the Association for Computational Linguistics (Volume 2: Short Papers)",
    month = aug,
    year = "2024",
    address = "Bangkok, Thailand",
    publisher = "Association for Computational Linguistics",
    url = "https://aclanthology.org/2024.acl-short.43/",
    doi = "10.18653/v1/2024.acl-short.43",
    pages = "459--469",

}

@inproceedings{ghosh-etal-2019-dependency,
    title = "Dependency Parser for {B}engali-{E}nglish Code-Mixed Data enhanced with a Synthetic Treebank",
    author = "Ghosh, Urmi  and
      Sharma, Dipti  and
      Khanuja, Simran",
    editor = "Candito, Marie  and
      Evang, Kilian  and
      Oepen, Stephan  and
      Seddah, Djam{\'e}",
    booktitle = "Proceedings of the 18th International Workshop on Treebanks and Linguistic Theories (TLT, SyntaxFest 2019)",
    month = aug,
    year = "2019",
    address = "Paris, France",
    publisher = "Association for Computational Linguistics",
    url = "https://aclanthology.org/W19-7810/",
    doi = "10.18653/v1/W19-7810",
    pages = "91--99"
}

@inproceedings{tarunesh-etal-2021-machine,
    title = "From Machine Translation to Code-Switching: Generating High-Quality Code-Switched Text",
    author = "Tarunesh, Ishan  and
      Kumar, Syamantak  and
      Jyothi, Preethi",
    editor = "Zong, Chengqing  and
      Xia, Fei  and
      Li, Wenjie  and
      Navigli, Roberto",
    booktitle = "Proceedings of the 59th Annual Meeting of the Association for Computational Linguistics and the 11th International Joint Conference on Natural Language Processing (Volume 1: Long Papers)",
    month = aug,
    year = "2021",
    address = "Online",
    publisher = "Association for Computational Linguistics",
    url = "https://aclanthology.org/2021.acl-long.245/",
    doi = "10.18653/v1/2021.acl-long.245",
    pages = "3154--3169",
}

@inproceedings{kugathasan-sumathipala-2021-neural,
    title = "Neural Machine Translation for {S}inhala-{E}nglish Code-Mixed Text",
    author = "Kugathasan, Archchana  and
      Sumathipala, Sagara",
    editor = "Mitkov, Ruslan  and
      Angelova, Galia",
    booktitle = "Proceedings of the International Conference on Recent Advances in Natural Language Processing (RANLP 2021)",
    month = sep,
    year = "2021",
    address = "Held Online",
    publisher = "INCOMA Ltd.",
    url = "https://aclanthology.org/2021.ranlp-1.82/",
    pages = "718--726",
}

@inproceedings{jawahar-etal-2021-exploring,
    title = "Exploring Text-to-Text Transformers for {E}nglish to {H}inglish Machine Translation with Synthetic Code-Mixing",
    author = "Jawahar, Ganesh  and
      Nagoudi, El Moatez Billah  and
      Abdul-Mageed, Muhammad  and
      Lakshmanan, V.S., Laks",
    editor = "Solorio, Thamar  and
      Chen, Shuguang  and
      Black, Alan W.  and
      Diab, Mona  and
      Sitaram, Sunayana  and
      Soto, Victor  and
      Yilmaz, Emre  and
      Srinivasan, Anirudh",
    booktitle = "Proceedings of the Fifth Workshop on Computational Approaches to Linguistic Code-Switching",
    month = jun,
    year = "2021",
    address = "Online",
    publisher = "Association for Computational Linguistics",
    url = "https://aclanthology.org/2021.calcs-1.6/",
    doi = "10.18653/v1/2021.calcs-1.6",
    pages = "36--46",
}

@inproceedings{raj-khan-etal-2021-towards-developing,
    title = "Towards Developing a Multilingual and Code-Mixed Visual Question Answering System by Knowledge Distillation",
    author = "Raj Khan, Humair  and
      Gupta, Deepak  and
      Ekbal, Asif",
    editor = "Moens, Marie-Francine  and
      Huang, Xuanjing  and
      Specia, Lucia  and
      Yih, Scott Wen-tau",
    booktitle = "Findings of the Association for Computational Linguistics: EMNLP 2021",
    month = nov,
    year = "2021",
    address = "Punta Cana, Dominican Republic",
    publisher = "Association for Computational Linguistics",
    url = "https://aclanthology.org/2021.findings-emnlp.151/",
    doi = "10.18653/v1/2021.findings-emnlp.151",
    pages = "1753--1767",
}

@inproceedings{gautam-etal-2021-translate,
    title = "Translate and Classify: Improving Sequence Level Classification for {E}nglish-{H}indi Code-Mixed Data",
    author = "Gautam, Devansh  and
      Gupta, Kshitij  and
      Shrivastava, Manish",
    editor = "Solorio, Thamar  and
      Chen, Shuguang  and
      Black, Alan W.  and
      Diab, Mona  and
      Sitaram, Sunayana  and
      Soto, Victor  and
      Yilmaz, Emre  and
      Srinivasan, Anirudh",
    booktitle = "Proceedings of the Fifth Workshop on Computational Approaches to Linguistic Code-Switching",
    month = jun,
    year = "2021",
    address = "Online",
    publisher = "Association for Computational Linguistics",
    url = "https://aclanthology.org/2021.calcs-1.3/",
    doi = "10.18653/v1/2021.calcs-1.3",
    pages = "15--25",
}

@inproceedings{tasawong-etal-2023-typo,
    title = "Typo-Robust Representation Learning for Dense Retrieval",
    author = "Tasawong, Panuthep  and
      Ponwitayarat, Wuttikorn  and
      Limkonchotiwat, Peerat  and
      Udomcharoenchaikit, Can  and
      Chuangsuwanich, Ekapol  and
      Nutanong, Sarana",
    editor = "Rogers, Anna  and
      Boyd-Graber, Jordan  and
      Okazaki, Naoaki",
    booktitle = "Proceedings of the 61st Annual Meeting of the Association for Computational Linguistics (Volume 2: Short Papers)",
    month = jul,
    year = "2023",
    address = "Toronto, Canada",
    publisher = "Association for Computational Linguistics",
    url = "https://aclanthology.org/2023.acl-short.95/",
    doi = "10.18653/v1/2023.acl-short.95",
    pages = "1106--1115",
}

@inproceedings{fernando-ranathunga-2021-data,
    title = "Data Augmentation to Address Out of {V}ocabulary{P}roblem in Low Resource {S}inhala {E}nglish Neural Machine Translation",
    author = "Fernando, Aloka  and
      Ranathunga, Surangika",
    editor = "Hu, Kaibao  and
      Kim, Jong-Bok  and
      Zong, Chengqing  and
      Chersoni, Emmanuele",
    booktitle = "Proceedings of the 35th Pacific Asia Conference on Language, Information and Computation",
    month = "11",
    year = "2021",
    address = "Shanghai, China",
    publisher = "Association for Computational Lingustics",
    url = "https://aclanthology.org/2021.paclic-1.7/",
    pages = "61--70"
}

@inproceedings{yang-tu-2022-combining,
    title = "Combining (Second-Order) Graph-Based and Headed-Span-Based Projective Dependency Parsing",
    author = "Yang, Songlin  and
      Tu, Kewei",
    editor = "Muresan, Smaranda  and
      Nakov, Preslav  and
      Villavicencio, Aline",
    booktitle = "Findings of the Association for Computational Linguistics: ACL 2022",
    month = may,
    year = "2022",
    address = "Dublin, Ireland",
    publisher = "Association for Computational Linguistics",
    url = "https://aclanthology.org/2022.findings-acl.112/",
    doi = "10.18653/v1/2022.findings-acl.112",
    pages = "1428--1434",
}

@inproceedings{chou-etal-2023-advancing,
    title = "Advancing Multi-Criteria {C}hinese Word Segmentation Through Criterion Classification and Denoising",
    author = "Chou, Tzu Hsuan  and
      Lin, Chun-Yi  and
      Kao, Hung-Yu",
    editor = "Rogers, Anna  and
      Boyd-Graber, Jordan  and
      Okazaki, Naoaki",
    booktitle = "Proceedings of the 61st Annual Meeting of the Association for Computational Linguistics (Volume 1: Long Papers)",
    month = jul,
    year = "2023",
    address = "Toronto, Canada",
    publisher = "Association for Computational Linguistics",
    url = "https://aclanthology.org/2023.acl-long.356/",
    doi = "10.18653/v1/2023.acl-long.356",
    pages = "6460--6476",

}

@inproceedings{perera-sumanathilaka-2025-machine,
    title = "Machine Translation and Transliteration for {I}ndo-{A}ryan Languages: A Systematic Review",
    author = "Perera, Sandun Sameera  and
      Sumanathilaka, Deshan Koshala",
    editor = "Weerasinghe, Ruvan  and
      Anuradha, Isuri  and
      Sumanathilaka, Deshan",
    booktitle = "Proceedings of the First Workshop on Natural Language Processing for Indo-Aryan and Dravidian Languages",
    month = jan,
    year = "2025",
    address = "Abu Dhabi",
    publisher = "Association for Computational Linguistics",
    url = "https://aclanthology.org/2025.indonlp-1.2/",
    pages = "11--21",
}

@inproceedings{winata-etal-2021-multilingual,
    title = "Are Multilingual Models Effective in Code-Switching?",
    author = "Winata, Genta Indra  and
      Cahyawijaya, Samuel  and
      Liu, Zihan  and
      Lin, Zhaojiang  and
      Madotto, Andrea  and
      Fung, Pascale",
    editor = "Solorio, Thamar  and
      Chen, Shuguang  and
      Black, Alan W.  and
      Diab, Mona  and
      Sitaram, Sunayana  and
      Soto, Victor  and
      Yilmaz, Emre  and
      Srinivasan, Anirudh",
    booktitle = "Proceedings of the Fifth Workshop on Computational Approaches to Linguistic Code-Switching",
    month = jun,
    year = "2021",
    address = "Online",
    publisher = "Association for Computational Linguistics",
    url = "https://aclanthology.org/2021.calcs-1.20/",
    doi = "10.18653/v1/2021.calcs-1.20",
    pages = "142--153",
}

@inproceedings{liu-etal-2022-mulzdg,
    title = "{M}ul{ZDG}: Multilingual Code-Switching Framework for Zero-shot Dialogue Generation",
    author = "Liu, Yongkang  and
      Feng, Shi  and
      Wang, Daling  and
      Zhang, Yifei",
    booktitle = "Proceedings of the 29th International Conference on Computational Linguistics",
    month = oct,
    year = "2022",
    address = "Gyeongju, Republic of Korea",
    publisher = "International Committee on Computational Linguistics",
    url = "https://aclanthology.org/2022.coling-1.54/",
    pages = "648--659",
}

@inproceedings{cho-etal-2020-towards,
    title = "Towards an Efficient Code-Mixed Grapheme-to-Phoneme Conversion in an Agglutinative Language: A Case Study on To-{K}orean Transliteration",
    author = "Cho, Won Ik  and
      Kim, Seok Min  and
      Kim, Nam Soo",
    booktitle = "Proceedings of the 4th Workshop on Computational Approaches to Code Switching",
    month = may,
    year = "2020",
    address = "Marseille, France",
    publisher = "European Language Resources Association",
    url = "https://aclanthology.org/2020.calcs-1.9/",
    pages = "65--70",
    language = "eng",
    ISBN = "979-10-95546-66-5",
}

@article{mohamed2025lost,
  title={Lost in the Mix: Evaluating LLM Understanding of Code-Switched Text},
  author={Mohamed, Amr and Zhang, Yang and Vazirgiannis, Michalis and Shang, Guokan},
  journal={arXiv preprint arXiv:2506.14012},
  year={2025},
url={https://arxiv.org/abs/2506.14012}
}

@inproceedings{lambebo-tonja-etal-2022-transformer,
    title = "Transformer-based Model for Word Level Language Identification in Code-mixed {K}annada-{E}nglish Texts",
    author = "Lambebo Tonja, Atnafu  and
      Gemeda Yigezu, Mesay  and
      Kolesnikova, Olga  and
      Shahiki Tash, Moein  and
      Sidorov, Grigori  and
      Gelbukh, Alexander",
    editor = "Chakravarthi, Bharathi Raja  and
      Murugappan, Abirami  and
      Chinnappa, Dhivya  and
      Hane, Adeep  and
      Kumeresan, Prasanna Kumar  and
      Ponnusamy, Rahul",
    booktitle = "Proceedings of the 19th International Conference on Natural Language Processing (ICON): Shared Task on Word Level Language Identification in Code-mixed Kannada-English Texts",
    month = dec,
    year = "2022",
    address = "IIIT Delhi, New Delhi, India",
    publisher = "Association for Computational Linguistics",
    url = "https://aclanthology.org/2022.icon-wlli.4/",
    pages = "18--24",
}

@inproceedings{sterner-teufel-2023-tongueswitcher,
    title = "{T}ongue{S}witcher: Fine-Grained Identification of {G}erman-{E}nglish Code-Switching",
    author = "Sterner, Igor  and
      Teufel, Simone",
    editor = "Winata, Genta  and
      Kar, Sudipta  and
      Zhukova, Marina  and
      Solorio, Thamar  and
      Diab, Mona  and
      Sitaram, Sunayana  and
      Choudhury, Monojit  and
      Bali, Kalika",
    booktitle = "Proceedings of the 6th Workshop on Computational Approaches to Linguistic Code-Switching",
    month = dec,
    year = "2023",
    address = "Singapore",
    publisher = "Association for Computational Linguistics",
    url = "https://aclanthology.org/2023.calcs-1.1/",
    pages = "1--13",
}

@inproceedings{dai-etal-2025-next,
    title = "Next-Level {C}antonese-to-{M}andarin Translation: Fine-Tuning and Post-Processing with {LLM}s",
    author = "Dai, Yuqian  and
      Chan, Chun Fai  and
      Wong, Ying Ki  and
      Pun, Tsz Ho",
    editor = "Hettiarachchi, Hansi  and
      Ranasinghe, Tharindu  and
      Rayson, Paul  and
      Mitkov, Ruslan  and
      Gaber, Mohamed  and
      Premasiri, Damith  and
      Tan, Fiona Anting  and
      Uyangodage, Lasitha",
    booktitle = "Proceedings of the First Workshop on Language Models for Low-Resource Languages",
    month = jan,
    year = "2025",
    address = "Abu Dhabi, United Arab Emirates",
    publisher = "Association for Computational Linguistics",
    url = "https://aclanthology.org/2025.loreslm-1.32/",
    pages = "427--436",
}

@inproceedings{goloburda-etal-2025-qorgau,
    title = "Qor{\'{g}}au: Evaluating Safety in {K}azakh-{R}ussian Bilingual Contexts",
    author = "Goloburda, Maiya  and
      Laiyk, Nurkhan  and
      Turmakhan, Diana  and
      Wang, Yuxia  and
      Togmanov, Mukhammed  and
      Mansurov, Jonibek  and
      Sametov, Askhat  and
      Mukhituly, Nurdaulet  and
      Wang, Minghan  and
      Orel, Daniil  and
      Mujahid, Zain Muhammad  and
      Koto, Fajri  and
      Baldwin, Timothy  and
      Nakov, Preslav",
    editor = "Che, Wanxiang  and
      Nabende, Joyce  and
      Shutova, Ekaterina  and
      Pilehvar, Mohammad Taher",
    booktitle = "Findings of the Association for Computational Linguistics: ACL 2025",
    month = jul,
    year = "2025",
    address = "Vienna, Austria",
    publisher = "Association for Computational Linguistics",
    url = "https://aclanthology.org/2025.findings-acl.507/",
    doi = "10.18653/v1/2025.findings-acl.507",
    pages = "9765--9784",
    ISBN = "979-8-89176-256-5",
}

@article{Ahmad2025HopeSD,
  title={Hope Speech Detection in code-mixed Roman Urdu tweets: A Positive Turn in Natural Language Processing},
  author={Muhammad Ahmad and Muhammad Waqas and Ameer Hamza and Ildar Z. Batyrshin and Grigori Sidorov},
  journal={ArXiv},
  year={2025},
  volume={abs/2506.21583},
  url={https://api.semanticscholar.org/CorpusID:280011615}
}

@inproceedings{dave-etal-2021-irnlp-daiict,
    title = "{IRNLP}{\_}{DAIICT}@{LT}-{EDI}-{EACL}2021: Hope Speech detection in Code Mixed text using {TF}-{IDF} Char N-grams and {M}u{RIL}",
    author = "Dave, Bhargav  and
      Bhat, Shripad  and
      Majumder, Prasenjit",
    editor = "Chakravarthi, Bharathi Raja  and
      McCrae, John P.  and
      Zarrouk, Manel  and
      Bali, Kalika  and
      Buitelaar, Paul",
    booktitle = "Proceedings of the First Workshop on Language Technology for Equality, Diversity and Inclusion",
    month = apr,
    year = "2021",
    address = "Kyiv",
    publisher = "Association for Computational Linguistics",
    url = "https://aclanthology.org/2021.ltedi-1.15/",
    pages = "114--117",
}

@inproceedings{balouchzahi-etal-2021-mucs-lt,
    title = "{MUCS}@{LT}-{EDI}-{EACL}2021:{C}o{H}ope-Hope Speech Detection for Equality, Diversity, and Inclusion in Code-Mixed Texts",
    author = "Balouchzahi, Fazlourrahman  and
      B K, Aparna  and
      Shashirekha, H L",
    editor = "Chakravarthi, Bharathi Raja  and
      McCrae, John P.  and
      Zarrouk, Manel  and
      Bali, Kalika  and
      Buitelaar, Paul",
    booktitle = "Proceedings of the First Workshop on Language Technology for Equality, Diversity and Inclusion",
    month = apr,
    year = "2021",
    address = "Kyiv",
    publisher = "Association for Computational Linguistics",
    url = "https://aclanthology.org/2021.ltedi-1.27/",
    pages = "180--187",
}

@misc{hande2021hope,
  title={Hope Speech detection in under-resourced Kannada language}, 
      author={Adeep Hande and Ruba Priyadharshini and Anbukkarasi Sampath and Kingston Pal Thamburaj and Prabakaran Chandran and Bharathi Raja Chakravarthi},
      year={2021},
      eprint={2108.04616},
      archivePrefix={arXiv},
      primaryClass={cs.CL},
      url={https://arxiv.org/abs/2108.04616}
}

@article{shanmugavadivel2022deep,
  title={Deep learning based sentiment analysis and offensive language identification on multilingual code-mixed data},
  author={Shanmugavadivel, Kogilavani and Sathishkumar, VE and Raja, Sandhiya and Lingaiah, T Bheema and Neelakandan, S and Subramanian, Malliga},
  journal={Scientific Reports},
  volume={12},
  number={1},
  pages={21557},
  year={2022},
  publisher={PubMed},
url={https://pubmed.ncbi.nlm.nih.gov/36513786/}
}

@inproceedings{jayanthi-gupta-2021-sj,
    title = "{SJ}{\_}{AJ}@{D}ravidian{L}ang{T}ech-{EACL}2021: Task-Adaptive Pre-Training of Multilingual {BERT} models for Offensive Language Identification",
    author = "Jayanthi, Sai Muralidhar  and
      Gupta, Akshat",
    editor = "Chakravarthi, Bharathi Raja  and
      Priyadharshini, Ruba  and
      Kumar M, Anand  and
      Krishnamurthy, Parameswari  and
      Sherly, Elizabeth",
    booktitle = "Proceedings of the First Workshop on Speech and Language Technologies for Dravidian Languages",
    month = apr,
    year = "2021",
    address = "Kyiv",
    publisher = "Association for Computational Linguistics",
    url = "https://aclanthology.org/2021.dravidianlangtech-1.44/",
    pages = "307--312"
}

@inproceedings{salaam-etal-2022-offensive,
    title = "Offensive Content Detection via Synthetic Code-Switched Text",
    author = "Salaam, Cesa  and
      Dernoncourt, Franck  and
      Bui, Trung  and
      Rawat, Danda  and
      Yoon, Seunghyun",
    booktitle = "Proceedings of the 29th International Conference on Computational Linguistics",
    month = oct,
    year = "2022",
    address = "Gyeongju, Republic of Korea",
    publisher = "International Committee on Computational Linguistics",
    url = "https://aclanthology.org/2022.coling-1.575/",
    pages = "6617--6624",
}

@inproceedings{pannerselvam-etal-2024-setfit,
    title = "{S}et{F}it: A Robust Approach for Offensive Content Detection in {T}amil-{E}nglish Code-Mixed Conversations Using Sentence Transfer Fine-tuning",
    author = "Pannerselvam, Kathiravan  and
      Rajiakodi, Saranya  and
      Thavareesan, Sajeetha  and
      Thangasamy, Sathiyaraj  and
      Ponnusamy, Kishore",
    editor = "Chakravarthi, Bharathi Raja  and
      Priyadharshini, Ruba  and
      Madasamy, Anand Kumar  and
      Thavareesan, Sajeetha  and
      Sherly, Elizabeth  and
      Nadarajan, Rajeswari  and
      Ravikiran, Manikandan",
    booktitle = "Proceedings of the Fourth Workshop on Speech, Vision, and Language Technologies for Dravidian Languages",
    month = mar,
    year = "2024",
    address = "St. Julian's, Malta",
    publisher = "Association for Computational Linguistics",
    url = "https://aclanthology.org/2024.dravidianlangtech-1.6/",
    pages = "35--42",
}

@article{kumar2025multi,
  title={Multi-task detection of harmful content in code-mixed meme captions using large language models with zero-shot, few-shot, and fine-tuning approaches},
  author={Kumar, AK Indira and Sthanusubramoniani, Gayathri and Gupta, Deepa and Nair, Aarathi Rajagopalan and Alotaibi, Yousef Ajami and Zakariah, Mohammed},
  journal={Egyptian Informatics Journal},
  volume={30},
  pages={100683},
  year={2025},
  publisher={Elsevier},
url={https://doi.org/10.1016/j.eij.2025.100683}
}

@inproceedings{chatterjee-etal-2022-pacman,
    title = "{PACMAN}:{PA}rallel {C}ode{M}ixed d{A}ta generatio{N} for {POS} tagging",
    author = "Chatterjee, Arindam  and
      Sharma, Chhavi  and
      Raj, Ayush  and
      Ekbal, Asif",
    editor = "Akhtar, Md. Shad  and
      Chakraborty, Tanmoy",
    booktitle = "Proceedings of the 19th International Conference on Natural Language Processing (ICON)",
    month = dec,
    year = "2022",
    address = "New Delhi, India",
    publisher = "Association for Computational Linguistics",
    url = "https://aclanthology.org/2022.icon-main.29/",
    pages = "234--244",
}

@inproceedings{choudhary-etal-2026-llms,
    title = "Do {LLM}s model human linguistic variation? A case study in {H}indi-{E}nglish Verb code-mixing",
    author = "Choudhary, Mukund  and
      Jindal, Madhur  and
      Aeron, Gaurja  and
      Choudhury, Monojit",
    editor = "Demberg, Vera  and
      Inui, Kentaro  and
      Marquez, Llu{\'i}s",
    booktitle = "Findings of the {A}ssociation for {C}omputational {L}inguistics: {EACL} 2026",
    month = mar,
    year = "2026",
    address = "Rabat, Morocco",
    publisher = "Association for Computational Linguistics",
    url = "https://aclanthology.org/2026.findings-eacl.291/",
    doi = "10.18653/v1/2026.findings-eacl.291",
    pages = "5491--5509",
    ISBN = "979-8-89176-386-9",
}

@inproceedings{dowlagar-mamidi-2021-pre,
    title = "A Pre-trained Transformer and {CNN} Model with Joint Language {ID} and Part-of-Speech Tagging for Code-Mixed Social-Media Text",
    author = "Dowlagar, Suman  and
      Mamidi, Radhika",
    editor = "Mitkov, Ruslan  and
      Angelova, Galia",
    booktitle = "Proceedings of the International Conference on Recent Advances in Natural Language Processing (RANLP 2021)",
    month = sep,
    year = "2021",
    address = "Held Online",
    publisher = "INCOMA Ltd.",
    url = "https://aclanthology.org/2021.ranlp-1.42/",
    pages = "367--374"
}

@inproceedings{sabty-etal-2020-contextual,
    title = "Contextual Embeddings for {A}rabic-{E}nglish Code-Switched Data",
    author = "Sabty, Caroline  and
      Islam, Mohamed  and
      Abdennadher, Slim",
    editor = "Zitouni, Imed  and
      Abdul-Mageed, Muhammad  and
      Bouamor, Houda  and
      Bougares, Fethi  and
      El-Haj, Mahmoud  and
      Tomeh, Nadi  and
      Zaghouani, Wajdi",
    booktitle = "Proceedings of the Fifth Arabic Natural Language Processing Workshop",
    month = dec,
    year = "2020",
    address = "Barcelona, Spain (Online)",
    publisher = "Association for Computational Linguistics",
    url = "https://aclanthology.org/2020.wanlp-1.20/",
    pages = "215--225",
}

@inproceedings{aguilar-solorio-2020-english,
    title = "From {E}nglish to Code-Switching: Transfer Learning with Strong Morphological Clues",
    author = "Aguilar, Gustavo  and
      Solorio, Thamar",
    editor = "Jurafsky, Dan  and
      Chai, Joyce  and
      Schluter, Natalie  and
      Tetreault, Joel",
    booktitle = "Proceedings of the 58th Annual Meeting of the Association for Computational Linguistics",
    month = jul,
    year = "2020",
    address = "Online",
    publisher = "Association for Computational Linguistics",
    url = "https://aclanthology.org/2020.acl-main.716/",
    doi = "10.18653/v1/2020.acl-main.716",
    pages = "8033--8044",
}

@inproceedings{yong-etal-2023-prompting,
    title = "Prompting Multilingual Large Language Models to Generate Code-Mixed Texts: The Case of South {E}ast {A}sian Languages",
    author = "Yong, Zheng Xin  and
      Zhang, Ruochen  and
      Forde, Jessica  and
      Wang, Skyler  and
      Subramonian, Arjun  and
      Lovenia, Holy  and
      Cahyawijaya, Samuel  and
      Winata, Genta  and
      Sutawika, Lintang  and
      Cruz, Jan Christian Blaise  and
      Tan, Yin Lin  and
      Phan, Long  and
      Phan, Long  and
      Garcia, Rowena  and
      Solorio, Thamar  and
      Aji, Alham Fikri",
    editor = "Winata, Genta  and
      Kar, Sudipta  and
      Zhukova, Marina  and
      Solorio, Thamar  and
      Diab, Mona  and
      Sitaram, Sunayana  and
      Choudhury, Monojit  and
      Bali, Kalika",
    booktitle = "Proceedings of the 6th Workshop on Computational Approaches to Linguistic Code-Switching",
    month = dec,
    year = "2023",
    address = "Singapore",
    publisher = "Association for Computational Linguistics",
    url = "https://aclanthology.org/2023.calcs-1.5/",
    pages = "43--63" 
}

@inproceedings{arora-etal-2023-comix,
    title = "{C}o{M}ix: Guide Transformers to Code-Mix using {POS} structure and Phonetics",
    author = "Arora, Gaurav  and
      Merugu, Srujana  and
      Sembium, Vivek",
    editor = "Rogers, Anna  and
      Boyd-Graber, Jordan  and
      Okazaki, Naoaki",
    booktitle = "Findings of the Association for Computational Linguistics: ACL 2023",
    month = jul,
    year = "2023",
    address = "Toronto, Canada",
    publisher = "Association for Computational Linguistics",
    url = "https://aclanthology.org/2023.findings-acl.506/",
    doi = "10.18653/v1/2023.findings-acl.506",
    pages = "7985--8002",
}

@inproceedings{absar-2025-fine,
    title = "Fine-Tuning Cross-Lingual {LLM}s for {POS} Tagging in Code-Switched Contexts",
    author = "Absar, Shayaan",
    editor = "Holdt, {\v{S}}pela Arhar  and
      Ilinykh, Nikolai  and
      Scalvini, Barbara  and
      Bruton, Micaella  and
      Debess, Iben Nyholm  and
      Tudor, Crina Madalina",
    booktitle = "Proceedings of the Third Workshop on Resources and Representations for Under-Resourced Languages and Domains (RESOURCEFUL-2025)",
    month = mar,
    year = "2025",
    address = "Tallinn, Estonia",
    publisher = "University of Tartu Library, Estonia",
    url = "https://aclanthology.org/2025.resourceful-1.2/",
    pages = "7--12",
    ISBN = "978-9908-53-121-2",
}

@article{article,
author = {R, Geetha and D, Karthika and Kumar, L Ashok},
year = {2025},
month = {03},
pages = {761-764},
title = {ENHANCING ASR ACCURACY AND COHERENCE ACROSS INDIAN LANGUAGES WITH WAV2VEC2 AND GPT-2},
volume = {6},
journal = {ICTACT Journal on Data Science and Machine Learning},
doi = {10.21917/ijdsml.2025.0156}
}

@inproceedings{pu-etal-2022-cmb,
    title = "{CMB} {AI} Lab at {S}em{E}val-2022 Task 11: A Two-Stage Approach for Complex Named Entity Recognition via Span Boundary Detection and Span Classification",
    author = "Pu, Keyu  and
      Liu, Hongyi  and
      Yang, Yixiao  and
      Ji, Jiangzhou  and
      Lv, Wenyi  and
      He, Yaohan",
    booktitle = "Proceedings of the 16th International Workshop on Semantic Evaluation (SemEval-2022)",
    month = jul,
    year = "2022",
    address = "Seattle, United States",
    publisher = "Association for Computational Linguistics",
    url = "https://aclanthology.org/2022.semeval-1.221/",
    doi = "10.18653/v1/2022.semeval-1.221",
    pages = "1603--1607",
}

@inproceedings{wang-etal-2018-code,
    title = "Code-Switched Named Entity Recognition with Embedding Attention",
    author = "Wang, Changhan  and
      Cho, Kyunghyun  and
      Kiela, Douwe",
    editor = "Aguilar, Gustavo  and
      AlGhamdi, Fahad  and
      Soto, Victor  and
      Solorio, Thamar  and
      Diab, Mona  and
      Hirschberg, Julia",
    booktitle = "Proceedings of the Third Workshop on Computational Approaches to Linguistic Code-Switching",
    month = jul,
    year = "2018",
    address = "Melbourne, Australia",
    publisher = "Association for Computational Linguistics",
    url = "https://aclanthology.org/W18-3221/",
    doi = "10.18653/v1/W18-3221",
    pages = "154--158",
}

@inproceedings{el-mekki-etal-2022-um6p,
    title = "{UM}6{P}-{CS} at {S}em{E}val-2022 Task 11: Enhancing Multilingual and Code-Mixed Complex Named Entity Recognition via Pseudo Labels using Multilingual Transformer",
    author = "El Mekki, Abdellah  and
      El Mahdaouy, Abdelkader  and
      Akallouch, Mohammed  and
      Berrada, Ismail  and
      Khoumsi, Ahmed",
    booktitle = "Proceedings of the 16th International Workshop on Semantic Evaluation (SemEval-2022)",
    month = jul,
    year = "2022",
    address = "Seattle, United States",
    publisher = "Association for Computational Linguistics",
    url = "https://aclanthology.org/2022.semeval-1.207/",
    doi = "10.18653/v1/2022.semeval-1.207",
    pages = "1511--1517",
}

@inproceedings{ijcai2020p0533,
  title     = {CoSDA-ML: Multi-Lingual Code-Switching Data Augmentation  for Zero-Shot Cross-Lingual NLP},
  author    = {Qin, Libo and Ni, Minheng and Zhang, Yue and Che, Wanxiang},
  booktitle = {Proceedings of the Twenty-Ninth International Joint Conference on
               Artificial Intelligence, {IJCAI-20}},
  publisher = {International Joint Conferences on Artificial Intelligence Organization},
  editor    = {Christian Bessiere},
  pages     = {3853--3860},
  year      = {2020},
  month     = {7},
  note      = {Main track},
  doi       = {10.24963/ijcai.2020/533},
  url       = {https://doi.org/10.24963/ijcai.2020/533},
}

@inproceedings{gupta-2022-malm,
    title = "{MALM}: Mixing Augmented Language Modeling for Zero-Shot Machine Translation",
    author = "Gupta, Kshitij",
    editor = {H{\"a}m{\"a}l{\"a}inen, Mika  and
      Alnajjar, Khalid  and
      Partanen, Niko  and
      Rueter, Jack},
    booktitle = "Proceedings of the 2nd International Workshop on Natural Language Processing for Digital Humanities",
    month = nov,
    year = "2022",
    address = "Taipei, Taiwan",
    publisher = "Association for Computational Linguistics",
    url = "https://aclanthology.org/2022.nlp4dh-1.8/",
    doi = "10.18653/v1/2022.nlp4dh-1.8",
    pages = "53--58",
}

@ARTICLE{10938193,
  author={Yadav, Anjali and Garg, Tanya and Klemen, Matej and Ulčar, Matej and Agarwal, Basant and Robnik-Šikonja, M.},
  journal={IEEE Transactions on Affective Computing}, 
  title={From Translation to Generative LLMs: Classification of Code-Mixed Affective Tasks}, 
  year={2025},
  volume={16},
  number={3},
  pages={2090-2101},
  doi={10.1109/TAFFC.2025.3553399}}

@ARTICLE{11016028,
  author={Tahery, Saedeh and Farzi, Saeed},
  journal={IEEE Access}, 
  title={An Adapted Few-Shot Prompting Technique Using ChatGPT to Advance Low-Resource Languages Understanding}, 
  year={2025},
  volume={13},
  number={},
  pages={93614-93628},
  doi={10.1109/ACCESS.2025.3574115}}

@misc{kruk2025banglassist,
title={BanglAssist: A Bengali-English Generative AI Chatbot for Code-Switching and Dialect-Handling in Customer Service}, 
      author={Francesco Kruk and Savindu Herath and Prithwiraj Choudhury},
      year={2025},
      eprint={2503.22283},
      archivePrefix={arXiv},
      primaryClass={cs.HC},
      url={https://arxiv.org/abs/2503.22283}
}

@inproceedings{brixey-traum-2025-code,
    title = "Does a code-switching dialogue system help users learn conversational fluency in {C}hoctaw?",
    author = "Brixey, Jacqueline  and
      Traum, David",
    editor = "Mager, Manuel  and
      Ebrahimi, Abteen  and
      Pugh, Robert  and
      Rijhwani, Shruti  and
      Von Der Wense, Katharina  and
      Chiruzzo, Luis  and
      Coto-Solano, Rolando  and
      Oncevay, Arturo",
    booktitle = "Proceedings of the Fifth Workshop on NLP for Indigenous Languages of the Americas (AmericasNLP)",
    month = may,
    year = "2025",
    address = "Albuquerque, New Mexico",
    publisher = "Association for Computational Linguistics",
    url = "https://aclanthology.org/2025.americasnlp-1.2/",
    doi = "10.18653/v1/2025.americasnlp-1.2",
    pages = "8--17",
    ISBN = "979-8-89176-236-7"
}

@inproceedings{zhang-eickhoff-2024-crocosum,
    title = "{C}ro{C}o{S}um: A Benchmark Dataset for Cross-Lingual Code-Switched Summarization",
    author = "Zhang, Ruochen  and
      Eickhoff, Carsten",
    editor = "Calzolari, Nicoletta  and
      Kan, Min-Yen  and
      Hoste, Veronique  and
      Lenci, Alessandro  and
      Sakti, Sakriani  and
      Xue, Nianwen",
    booktitle = "Proceedings of the 2024 Joint International Conference on Computational Linguistics, Language Resources and Evaluation (LREC-COLING 2024)",
    month = may,
    year = "2024",
    address = "Torino, Italia",
    publisher = "ELRA and ICCL",
    url = "https://aclanthology.org/2024.lrec-main.367/",
    pages = "4113--4126"
}

@inproceedings{sheth-etal-2025-comi,
    title = "{COMI}-{LINGUA}: Expert Annotated Large-Scale Dataset for Multitask {NLP} in {H}indi-{E}nglish Code-Mixing",
    author = "Sheth, Rajvee  and
      Beniwal, Himanshu  and
      Singh, Mayank",
    editor = "Christodoulopoulos, Christos  and
      Chakraborty, Tanmoy  and
      Rose, Carolyn  and
      Peng, Violet",
    booktitle = "Findings of the Association for Computational Linguistics: EMNLP 2025",
    month = nov,
    year = "2025",
    address = "Suzhou, China",
    publisher = "Association for Computational Linguistics",
    url = "https://aclanthology.org/2025.findings-emnlp.422/",
    doi = "10.18653/v1/2025.findings-emnlp.422",
    pages = "7973--7992",
    ISBN = "979-8-89176-335-7",
}

@inproceedings{ijcai2019p0719,
  title     = {A Deep Generative Model for Code Switched Text},
  author    = {Samanta, Bidisha and Reddy, Sharmila and Jagirdar, Hussain and Ganguly, Niloy and Chakrabarti, Soumen},
  booktitle = {Proceedings of the Twenty-Eighth International Joint Conference on
               Artificial Intelligence, {IJCAI-19}},
  publisher = {International Joint Conferences on Artificial Intelligence Organization},
  pages     = {5175--5181},
  year      = {2019},
  month     = {7},
  doi       = {10.24963/ijcai.2019/719},
  url       = {https://doi.org/10.24963/ijcai.2019/719},
}

@inproceedings{nayak-joshi-2022-l3cube,
    title = "{L}3{C}ube-{H}ing{C}orpus and {H}ing{BERT}: A Code Mixed {H}indi-{E}nglish Dataset and {BERT} Language Models",
    author = "Nayak, Ravindra  and
      Joshi, Raviraj",
    editor = "Jha, Girish Nath  and
      L., Sobha  and
      Bali, Kalika  and
      Ojha, Atul Kr.",
    booktitle = "Proceedings of the WILDRE-6 Workshop within the 13th Language Resources and Evaluation Conference",
    month = jun,
    year = "2022",
    address = "Marseille, France",
    publisher = "European Language Resources Association",
    url = "https://aclanthology.org/2022.wildre-1.2/",
    pages = "7--12",
}

@inproceedings{wang-etal-2025-gpt,
    title = "{GPT}-{NER}: Named Entity Recognition via Large Language Models",
    author = "Wang, Shuhe  and
      Sun, Xiaofei  and
      Li, Xiaoya  and
      Ouyang, Rongbin  and
      Wu, Fei  and
      Zhang, Tianwei  and
      Li, Jiwei  and
      Wang, Guoyin  and
      Guo, Chen",
    editor = "Chiruzzo, Luis  and
      Ritter, Alan  and
      Wang, Lu",
    booktitle = "Findings of the Association for Computational Linguistics: NAACL 2025",
    month = apr,
    year = "2025",
    address = "Albuquerque, New Mexico",
    publisher = "Association for Computational Linguistics",
    url = "https://aclanthology.org/2025.findings-naacl.239/",
    doi = "10.18653/v1/2025.findings-naacl.239",
    pages = "4257--4275",
    ISBN = "979-8-89176-195-7",
}

@inproceedings{zaratiana-etal-2024-gliner,
    title = "{GL}i{NER}: Generalist Model for Named Entity Recognition using Bidirectional Transformer",
    author = "Zaratiana, Urchade  and
      Tomeh, Nadi  and
      Holat, Pierre  and
      Charnois, Thierry",
    editor = "Duh, Kevin  and
      Gomez, Helena  and
      Bethard, Steven",
    booktitle = "Proceedings of the 2024 Conference of the North American Chapter of the Association for Computational Linguistics: Human Language Technologies (Volume 1: Long Papers)",
    month = jun,
    year = "2024",
    address = "Mexico City, Mexico",
    publisher = "Association for Computational Linguistics",
    url = "https://aclanthology.org/2024.naacl-long.300/",
    doi = "10.18653/v1/2024.naacl-long.300",
    pages = "5364--5376",
}

@inproceedings{s-shrivastava-2022-kanglish,
    title = "``Kanglish alli names!'' Named Entity Recognition for {K}annada-{E}nglish Code-Mixed Social Media Data",
    author = "S, Sumukh  and
      Shrivastava, Manish",
    booktitle = "Proceedings of the Eighth Workshop on Noisy User-generated Text (W-NUT 2022)",
    month = oct,
    year = "2022",
    address = "Gyeongju, Republic of Korea",
    publisher = "Association for Computational Linguistics",
    url = "https://aclanthology.org/2022.wnut-1.17/",
    pages = "154--161"
}

@inproceedings{malmasi-etal-2022-semeval,
    title = "{S}em{E}val-2022 Task 11: Multilingual Complex Named Entity Recognition ({M}ulti{C}o{NER})",
    author = "Malmasi, Shervin  and
      Fang, Anjie  and
      Fetahu, Besnik  and
      Kar, Sudipta  and
      Rokhlenko, Oleg",
    editor = "Emerson, Guy  and
      Schluter, Natalie  and
      Stanovsky, Gabriel  and
      Kumar, Ritesh  and
      Palmer, Alexis  and
      Schneider, Nathan  and
      Singh, Siddharth  and
      Ratan, Shyam",
    booktitle = "Proceedings of the 16th International Workshop on Semantic Evaluation (SemEval-2022)",
    month = jul,
    year = "2022",
    address = "Seattle, United States",
    publisher = "Association for Computational Linguistics",
    url = "https://aclanthology.org/2022.semeval-1.196/",
    doi = "10.18653/v1/2022.semeval-1.196",
    pages = "1412--1437"
}

@inproceedings{shynkarov-etal-2025-improving,
    title = "Improving Sentiment Analysis for {U}krainian Social Media Code-Switching Data",
    author = "Shynkarov, Yurii  and
      Solopova, Veronika  and
      Schmitt, Vera",
    editor = "Romanyshyn, Mariana",
    booktitle = "Proceedings of the Fourth Ukrainian Natural Language Processing Workshop (UNLP 2025)",
    month = jul,
    year = "2025",
    address = "Vienna, Austria (online)",
    publisher = "Association for Computational Linguistics",
    url = "https://aclanthology.org/2025.unlp-1.18/",
    doi = "10.18653/v1/2025.unlp-1.18",
    pages = "179--193",
    ISBN = "979-8-89176-269-5"
}

@inproceedings{khanuja-etal-2020-new,
    title = "A New Dataset for Natural Language Inference from Code-mixed Conversations",
    author = "Khanuja, Simran  and
      Dandapat, Sandipan  and
      Sitaram, Sunayana  and
      Choudhury, Monojit",
    editor = "Solorio, Thamar  and
      Choudhury, Monojit  and
      Bali, Kalika  and
      Sitaram, Sunayana  and
      Das, Amitava  and
      Diab, Mona",
    booktitle = "Proceedings of the 4th Workshop on Computational Approaches to Code Switching",
    month = may,
    year = "2020",
    address = "Marseille, France",
    publisher = "European Language Resources Association",
    url = "https://aclanthology.org/2020.calcs-1.2/",
    pages = "9--16",
    language = "eng",
    ISBN = "979-10-95546-66-5",
}

@inproceedings{birshert2021call,
  title={Call Larisa Ivanovna: Code-Switching Fools Multilingual NLU Models},
  author={Birshert, Alexey and Artemova, Ekaterina},
  booktitle={International Conference on Analysis of Images, Social Networks and Texts},
  pages={3--16},
  year={2021},
  organization={Springer}
}

@inproceedings{parikh-solorio-2021-normalization,
    title = "Normalization and Back-Transliteration for Code-Switched Data",
    author = "Parikh, Dwija  and
      Solorio, Thamar",
    editor = "Solorio, Thamar  and
      Chen, Shuguang  and
      Black, Alan W.  and
      Diab, Mona  and
      Sitaram, Sunayana  and
      Soto, Victor  and
      Yilmaz, Emre  and
      Srinivasan, Anirudh",
    booktitle = "Proceedings of the Fifth Workshop on Computational Approaches to Linguistic Code-Switching",
    month = jun,
    year = "2021",
    address = "Online",
    publisher = "Association for Computational Linguistics",
    url = "https://aclanthology.org/2021.calcs-1.15/",
    doi = "10.18653/v1/2021.calcs-1.15",
    pages = "119--124",
}

@inproceedings{taguchi-etal-2021-transliteration,
    title = "Transliteration for Low-Resource Code-Switching Texts: Building an Automatic {C}yrillic-to-{L}atin Converter for {T}atar",
    author = "Taguchi, Chihiro  and
      Sakai, Yusuke  and
      Watanabe, Taro",
    editor = "Solorio, Thamar  and
      Chen, Shuguang  and
      Black, Alan W.  and
      Diab, Mona  and
      Sitaram, Sunayana  and
      Soto, Victor  and
      Yilmaz, Emre  and
      Srinivasan, Anirudh",
    booktitle = "Proceedings of the Fifth Workshop on Computational Approaches to Linguistic Code-Switching",
    month = jun,
    year = "2021",
    address = "Online",
    publisher = "Association for Computational Linguistics",
    url = "https://aclanthology.org/2021.calcs-1.18/",
    doi = "10.18653/v1/2021.calcs-1.18",
    pages = "133--140",
}

@inproceedings{wang-etal-2025-investigating-scaling,
    title = "Investigating and Scaling up Code-Switching for Multilingual Language Model Pre-Training",
    author = "Wang, Zhijun  and
      Li, Jiahuan  and
      Zhou, Hao  and
      Weng, Rongxiang  and
      Wang, Jingang  and
      Huang, Xin  and
      Han, Xue  and
      Feng, Junlan  and
      Deng, Chao  and
      Huang, Shujian",
    editor = "Che, Wanxiang  and
      Nabende, Joyce  and
      Shutova, Ekaterina  and
      Pilehvar, Mohammad Taher",
    booktitle = "Findings of the Association for Computational Linguistics: ACL 2025",
    month = jul,
    year = "2025",
    address = "Vienna, Austria",
    publisher = "Association for Computational Linguistics",
    url = "https://aclanthology.org/2025.findings-acl.575/",
    doi = "10.18653/v1/2025.findings-acl.575",
    pages = "11032--11046",
    ISBN = "979-8-89176-256-5",
}

@inproceedings{angel2020nlp,
 title = "{NLP}-{CIC} at {S}em{E}val-2020 Task 9: Analysing Sentiment in Code-switching Language Using a Simple Deep-learning Classifier",
    author = "Angel, Jason  and
      Aroyehun, Segun Taofeek  and
      Tamayo, Antonio  and
      Gelbukh, Alexander",
    editor = "Herbelot, Aurelie  and
      Zhu, Xiaodan  and
      Palmer, Alexis  and
      Schneider, Nathan  and
      May, Jonathan  and
      Shutova, Ekaterina",
    booktitle = "Proceedings of the Fourteenth Workshop on Semantic Evaluation",
    month = dec,
    year = "2020",
    address = "Barcelona (online)",
    publisher = "International Committee for Computational Linguistics",
    url = "https://aclanthology.org/2020.semeval-1.123/",
    doi = "10.18653/v1/2020.semeval-1.123",
    pages = "957--962",
}

@inproceedings{ma2020xlp,
title = "{XLP} at {S}em{E}val-2020 Task 9: Cross-lingual Models with Focal Loss for Sentiment Analysis of Code-Mixing Language",
    author = "Ma, Yili  and
      Zhao, Liang  and
      Hao, Jie",
    editor = "Herbelot, Aurelie  and
      Zhu, Xiaodan  and
      Palmer, Alexis  and
      Schneider, Nathan  and
      May, Jonathan  and
      Shutova, Ekaterina",
    booktitle = "Proceedings of the Fourteenth Workshop on Semantic Evaluation",
    month = dec,
    year = "2020",
    address = "Barcelona (online)",
    publisher = "International Committee for Computational Linguistics",
    url = "https://aclanthology.org/2020.semeval-1.126/",
    doi = "10.18653/v1/2020.semeval-1.126",
    pages = "975--980"
}

@inproceedings{sultan2020wessa,
title = "{WESSA} at {S}em{E}val-2020 Task 9: Code-Mixed Sentiment Analysis Using Transformers",
    author = "Sultan, Ahmed  and
      Salim, Mahmoud  and
      Gaber, Amina  and
      El Hosary, Islam",
    editor = "Herbelot, Aurelie  and
      Zhu, Xiaodan  and
      Palmer, Alexis  and
      Schneider, Nathan  and
      May, Jonathan  and
      Shutova, Ekaterina",
    booktitle = "Proceedings of the Fourteenth Workshop on Semantic Evaluation",
    month = dec,
    year = "2020",
    address = "Barcelona (online)",
    publisher = "International Committee for Computational Linguistics",
    url = "https://aclanthology.org/2020.semeval-1.181/",
    doi = "10.18653/v1/2020.semeval-1.181",
    pages = "1342--1347"
}

@inproceedings{singh2020voice,
 title = "Voice@{SRIB} at {S}em{E}val-2020 Tasks 9 and 12: Stacked Ensemblingmethod for Sentiment and Offensiveness detection in Social Media",
    author = "Singh, Abhishek  and
      Singh Parmar, Surya Pratap",
    editor = "Herbelot, Aurelie  and
      Zhu, Xiaodan  and
      Palmer, Alexis  and
      Schneider, Nathan  and
      May, Jonathan  and
      Shutova, Ekaterina",
    booktitle = "Proceedings of the Fourteenth Workshop on Semantic Evaluation",
    month = dec,
    year = "2020",
    address = "Barcelona (online)",
    publisher = "International Committee for Computational Linguistics",
    url = "https://aclanthology.org/2020.semeval-1.180/",
    doi = "10.18653/v1/2020.semeval-1.180",
    pages = "1331--1341"
}

@article{chakravarthi2022dravidiancodemix,
  title={Dravidiancodemix: Sentiment analysis and offensive language identification dataset for dravidian languages in code-mixed text},
  author={Chakravarthi, Bharathi Raja and Priyadharshini, Ruba and Muralidaran, Vigneshwaran and Jose, Navya and Suryawanshi, Shardul and Sherly, Elizabeth and McCrae, John P},
  journal={Language Resources and Evaluation},
  volume={56},
  number={3},
  pages={765--806},
  year={2022},
  publisher={Springer},
url={https://doi.org/10.1007/s10579-022-09583-7}
}

@article{nguyen2021automatic,
  title={Automatic language identification in code-switched Hindi-English social media text},
  author={Nguyen, Li and Bryant, Christopher and Kidwai, Sana and Biberauer, Theresa},
  journal={Journal of Open Humanities Data},
  volume={7},
  year={2021}
}

@inproceedings{van2022sentiment,
    title = "Sentiment Analysis in Code-Mixed {V}ietnamese-{E}nglish Sentence-level Hotel Reviews",
    author = "Van, Thin Dang  and
      Ngoc, Hao Duong  and
      Luu-Thuy, Ngan Nguyen",
    editor = "Dita, Shirley  and
      Trillanes, Arlene  and
      Lucas, Rochelle Irene",
    booktitle = "Proceedings of the 36th Pacific Asia Conference on Language, Information and Computation",
    month = oct,
    year = "2022",
    address = "Manila, Philippines",
    publisher = "Association for Computational Linguistics",
    url = "https://aclanthology.org/2022.paclic-1.7/",
    pages = "54--61"
}

@inproceedings{etori2025rideke,
  title = "{R}ide{KE}: Leveraging Low-resource {T}witter User-generated Content for Sentiment and Emotion Detection on Code-switched {RHS} Dataset.",
    author = "Etori, Naome  and
      Gini, Maria",
    editor = "De Clercq, Orph{\'e}e  and
      Barriere, Valentin  and
      Barnes, Jeremy  and
      Klinger, Roman  and
      Sedoc, Jo{\~a}o  and
      Tafreshi, Shabnam",
    booktitle = "Proceedings of the 14th Workshop on Computational Approaches to Subjectivity, Sentiment, {\&} Social Media Analysis",
    month = aug,
    year = "2024",
    address = "Bangkok, Thailand",
    publisher = "Association for Computational Linguistics",
    url = "https://aclanthology.org/2024.wassa-1.19/",
    doi = "10.18653/v1/2024.wassa-1.19",
    pages = "234--249",
}

@inproceedings{wadhawan2021towards,
    title = "Towards Emotion Recognition in {H}indi-{E}nglish Code-Mixed Data: A Transformer Based Approach",
    author = "Wadhawan, Anshul  and
      Aggarwal, Akshita",
    editor = "De Clercq, Orphee  and
      Balahur, Alexandra  and
      Sedoc, Joao  and
      Barriere, Valentin  and
      Tafreshi, Shabnam  and
      Buechel, Sven  and
      Hoste, Veronique",
    booktitle = "Proceedings of the Eleventh Workshop on Computational Approaches to Subjectivity, Sentiment and Social Media Analysis",
    month = apr,
    year = "2021",
    address = "Online",
    publisher = "Association for Computational Linguistics",
    url = "https://aclanthology.org/2021.wassa-1.21/",
    pages = "195--202"
}

@inproceedings{huang-yang-2023-culturally,
    title = "Culturally Aware Natural Language Inference",
    author = "Huang, Jing  and
      Yang, Diyi",
    editor = "Bouamor, Houda  and
      Pino, Juan  and
      Bali, Kalika",
    booktitle = "Findings of the Association for Computational Linguistics: EMNLP 2023",
    month = dec,
    year = "2023",
    address = "Singapore",
    publisher = "Association for Computational Linguistics",
    url = "https://aclanthology.org/2023.findings-emnlp.509/",
    doi = "10.18653/v1/2023.findings-emnlp.509",
    pages = "7591--7609",
}

@inproceedings{jayanthi2021unsupervised,
 title = "Unsupervised Self-Training for Sentiment Analysis of Code-Switched Data",
    author = "Gupta, Akshat  and
      Menghani, Sargam  and
      Rallabandi, Sai Krishna  and
      Black, Alan W",
    editor = "Solorio, Thamar  and
      Chen, Shuguang  and
      Black, Alan W.  and
      Diab, Mona  and
      Sitaram, Sunayana  and
      Soto, Victor  and
      Yilmaz, Emre  and
      Srinivasan, Anirudh",
    booktitle = "Proceedings of the Fifth Workshop on Computational Approaches to Linguistic Code-Switching",
    month = jun,
    year = "2021",
    address = "Online",
    publisher = "Association for Computational Linguistics",
    url = "https://aclanthology.org/2021.calcs-1.13/",
    doi = "10.18653/v1/2021.calcs-1.13",
    pages = "103--112"
}

@inproceedings{sharma2022progressive,
title = "Progressive Sentiment Analysis for Code-Switched Text Data",
    author = "Ranjan, Sudhanshu  and
      Mekala, Dheeraj  and
      Shang, Jingbo",
    editor = "Goldberg, Yoav  and
      Kozareva, Zornitsa  and
      Zhang, Yue",
    booktitle = "Findings of the Association for Computational Linguistics: EMNLP 2022",
    month = dec,
    year = "2022",
    address = "Abu Dhabi, United Arab Emirates",
    publisher = "Association for Computational Linguistics",
    url = "https://aclanthology.org/2022.findings-emnlp.82/",
    doi = "10.18653/v1/2022.findings-emnlp.82",
    pages = "1155--1167"
}

@inproceedings{park-etal-2024-multiprageval,
    title = "{M}ulti{P}rag{E}val: Multilingual Pragmatic Evaluation of Large Language Models",
    author = "Park, Dojun  and
      Lee, Jiwoo  and
      Park, Seohyun  and
      Jeong, Hyeyun  and
      Koo, Youngeun  and
      Hwang, Soonha  and
      Park, Seonwoo  and
      Lee, Sungeun",
    editor = "Hupkes, Dieuwke  and
      Dankers, Verna  and
      Batsuren, Khuyagbaatar  and
      Kazemnejad, Amirhossein  and
      Christodoulopoulos, Christos  and
      Giulianelli, Mario  and
      Cotterell, Ryan",
    booktitle = "Proceedings of the 2nd GenBench Workshop on Generalisation (Benchmarking) in NLP",
    month = nov,
    year = "2024",
    address = "Miami, Florida, USA",
    publisher = "Association for Computational Linguistics",
    url = "https://aclanthology.org/2024.genbench-1.7/",
    doi = "10.18653/v1/2024.genbench-1.7",
    pages = "96--119",
}

@inproceedings{ahuja-etal-2024-megaverse,
    title = "{MEGAVERSE}: Benchmarking Large Language Models Across Languages, Modalities, Models and Tasks",
    author = "Ahuja, Sanchit  and
      Aggarwal, Divyanshu  and
      Gumma, Varun  and
      Watts, Ishaan  and
      Sathe, Ashutosh  and
      Ochieng, Millicent  and
      Hada, Rishav  and
      Jain, Prachi  and
      Ahmed, Mohamed  and
      Bali, Kalika  and
      Sitaram, Sunayana",
    editor = "Duh, Kevin  and
      Gomez, Helena  and
      Bethard, Steven",
    booktitle = "Proceedings of the 2024 Conference of the North American Chapter of the Association for Computational Linguistics: Human Language Technologies (Volume 1: Long Papers)",
    month = jun,
    year = "2024",
    address = "Mexico City, Mexico",
    publisher = "Association for Computational Linguistics",
    url = "https://aclanthology.org/2024.naacl-long.143/",
    doi = "10.18653/v1/2024.naacl-long.143",
    pages = "2598--2637",
}

@inproceedings{winata-etal-2023-nusax,
    title = "{N}usa{X}: Multilingual Parallel Sentiment Dataset for 10 {I}ndonesian Local Languages",
    author = "Winata, Genta Indra  and
      Aji, Alham Fikri  and
      Cahyawijaya, Samuel  and
      Mahendra, Rahmad  and
      Koto, Fajri  and
      Romadhony, Ade  and
      Kurniawan, Kemal  and
      Moeljadi, David  and
      Prasojo, Radityo Eko  and
      Fung, Pascale  and
      Baldwin, Timothy  and
      Lau, Jey Han  and
      Sennrich, Rico  and
      Ruder, Sebastian",
    editor = "Vlachos, Andreas  and
      Augenstein, Isabelle",
    booktitle = "Proceedings of the 17th Conference of the European Chapter of the Association for Computational Linguistics",
    month = may,
    year = "2023",
    address = "Dubrovnik, Croatia",
    publisher = "Association for Computational Linguistics",
    url = "https://aclanthology.org/2023.eacl-main.57/",
    doi = "10.18653/v1/2023.eacl-main.57",
    pages = "815--834",
}

@inproceedings{wu-etal-2025-english,
    title = "From {E}nglish to Second Language Mastery: Enhancing {LLM}s with Cross-Lingual Continued Instruction Tuning",
    author = "Wu, Linjuan  and
      Wei, Hao-Ran  and
      Yang, Baosong  and
      Lu, Weiming",
    editor = "Che, Wanxiang  and
      Nabende, Joyce  and
      Shutova, Ekaterina  and
      Pilehvar, Mohammad Taher",
    booktitle = "Proceedings of the 63rd Annual Meeting of the Association for Computational Linguistics (Volume 1: Long Papers)",
    month = jul,
    year = "2025",
    address = "Vienna, Austria",
    publisher = "Association for Computational Linguistics",
    url = "https://aclanthology.org/2025.acl-long.1121/",
    doi = "10.18653/v1/2025.acl-long.1121",
    pages = "23006--23023",
    ISBN = "979-8-89176-251-0",
}

@inproceedings{gupta-etal-2018-uncovering,
    title = "Uncovering Code-Mixed Challenges: A Framework for Linguistically Driven Question Generation and Neural Based Question Answering",
    author = "Gupta, Deepak  and
      Lenka, Pabitra  and
      Ekbal, Asif  and
      Bhattacharyya, Pushpak",
    editor = "Korhonen, Anna  and
      Titov, Ivan",
    booktitle = "Proceedings of the 22nd Conference on Computational Natural Language Learning",
    month = oct,
    year = "2018",
    address = "Brussels, Belgium",
    publisher = "Association for Computational Linguistics",
    url = "https://aclanthology.org/K18-1012/",
    doi = "10.18653/v1/K18-1012",
    pages = "119--130",
}

@inproceedings{behzad-etal-2024-ask,
    title = "To Ask {LLM}s about {E}nglish Grammaticality, Prompt Them in a Different Language",
    author = "Behzad, Shabnam  and
      Zeldes, Amir  and
      Schneider, Nathan",
    editor = "Al-Onaizan, Yaser  and
      Bansal, Mohit  and
      Chen, Yun-Nung",
    booktitle = "Findings of the Association for Computational Linguistics: EMNLP 2024",
    month = nov,
    year = "2024",
    address = "Miami, Florida, USA",
    publisher = "Association for Computational Linguistics",
    url = "https://aclanthology.org/2024.findings-emnlp.916/",
    doi = "10.18653/v1/2024.findings-emnlp.916",
    pages = "15622--15634",
}

@inproceedings{hong-etal-2025-migrate,
    title = "{MIGRATE}: Cross-Lingual Adaptation of Domain-Specific {LLM}s through Code-Switching and Embedding Transfer",
    author = "Hong, Seongtae  and
      Lee, Seungyoon  and
      Moon, Hyeonseok  and
      Lim, Heuiseok",
    editor = "Rambow, Owen  and
      Wanner, Leo  and
      Apidianaki, Marianna  and
      Al-Khalifa, Hend  and
      Eugenio, Barbara Di  and
      Schockaert, Steven",
    booktitle = "Proceedings of the 31st International Conference on Computational Linguistics",
    month = jan,
    year = "2025",
    address = "Abu Dhabi, UAE",
    publisher = "Association for Computational Linguistics",
    url = "https://aclanthology.org/2025.coling-main.617/",
    pages = "9184--9193",
}

@inproceedings{agarwal2021hinglish,
   title = "{H}inglish to {E}nglish Machine Translation using Multilingual Transformers",
    author = "Agarwal, Vibhav  and
      Rao, Pooja  and
      Jayagopi, Dinesh Babu",
    editor = "Djabri, Souhila  and
      Gimadi, Dinara  and
      Mihaylova, Tsvetomila  and
      Nikolova-Koleva, Ivelina",
    booktitle = "Proceedings of the Student Research Workshop Associated with RANLP 2021",
    month = sep,
    year = "2021",
    address = "Online",
    publisher = "INCOMA Ltd.",
    url = "https://aclanthology.org/2021.ranlp-srw.3/",
    pages = "16--21",
}

@article{kodali2025adapting,
  title={Adapting Multilingual Models to Code-Mixed Tasks via Model Merging},
  author={Kodali, Prashant and Shivkumar, Vaishnavi and Joshi, Swarang and Choudhary, Monojit and Kumaraguru, Ponnurangam and Shrivastava, Manish},
  journal={arXiv preprint arXiv:2510.19782},
  year={2025}
}

@article{maimaiti2025improving,
  title={Improving cross-lingual representation for semantic retrieval with code-switching},
  author={Maimaiti, Mieradilijiang and Zheng, Yuanhang and Zhang, Ji and Zhang, Yue and Luo, Wenpei and Huang, Kaiyu},
  journal={Knowledge-Based Systems},
  pages={113919},
  year={2025},
  publisher={Elsevier}
}

@article{das2014identifying,
  title = "Code-Mixing in Social Media Text",
    author = {Das, Amitava  and
      Gamb{\"a}ck, Bj{\"o}rn},
    editor = "Villemonte de La Clergerie, {\'E}ric  and
      Lepage, Yves  and
      Minel, Jean-Luc  and
      S{\'e}billot, Pascale",
    journal = "Traitement Automatique des Langues",
    volume = "54",
    number = "3",
    year = "2013",
    address = "France",
    publisher = "ATALA (Association pour le Traitement Automatique des Langues)",
    url = "https://aclanthology.org/2013.tal-3.3/",
    pages = "41--64"
}

@inproceedings{piccinini2014prosodic,
  title={Prosodic cues to monolingual versus code-switching sentences in English and Spanish},
  author={Piccinini, Page Elizabeth and Garellek, Marc},
  booktitle={Proceedings of the 7th Speech Prosody Conference},
  pages={885--889},
  year={2014}
}

@inproceedings{gamback-das-2016-comparing,
    title = "Comparing the Level of Code-Switching in Corpora",
    author = {Gamb{\"a}ck, Bj{\"o}rn  and
      Das, Amitava},
    editor = "Calzolari, Nicoletta  and
      Choukri, Khalid  and
      Declerck, Thierry  and
      Goggi, Sara  and
      Grobelnik, Marko  and
      Maegaard, Bente  and
      Mariani, Joseph  and
      Mazo, Helene  and
      Moreno, Asuncion  and
      Odijk, Jan  and
      Piperidis, Stelios",
    booktitle = "Proceedings of the Tenth International Conference on Language Resources and Evaluation ({LREC}'16)",
    month = may,
    year = "2016",
    address = "Portoro{\v{z}}, Slovenia",
    publisher = "European Language Resources Association (ELRA)",
    url = "https://aclanthology.org/L16-1292/",
    pages = "1850--1855",
}

@inproceedings{papineni2002bleu,
title = "{B}leu: a Method for Automatic Evaluation of Machine Translation",
    author = "Papineni, Kishore  and
      Roukos, Salim  and
      Ward, Todd  and
      Zhu, Wei-Jing",
    editor = "Isabelle, Pierre  and
      Charniak, Eugene  and
      Lin, Dekang",
    booktitle = "Proceedings of the 40th Annual Meeting of the Association for Computational Linguistics",
    month = jul,
    year = "2002",
    address = "Philadelphia, Pennsylvania, USA",
    publisher = "Association for Computational Linguistics",
    url = "https://aclanthology.org/P02-1040/",
    doi = "10.3115/1073083.1073135",
    pages = "311--318"
}

@inproceedings{popovic-2015-chrf,
    title = "chr{F}: character n-gram {F}-score for automatic {MT} evaluation",
    author = "Popovi{\'c}, Maja",
    editor = "Bojar, Ond{\v{r}}ej  and
      Chatterjee, Rajan  and
      Federmann, Christian  and
      Haddow, Barry  and
      Hokamp, Chris  and
      Huck, Matthias  and
      Logacheva, Varvara  and
      Pecina, Pavel",
    booktitle = "Proceedings of the Tenth Workshop on Statistical Machine Translation",
    month = sep,
    year = "2015",
    address = "Lisbon, Portugal",
    publisher = "Association for Computational Linguistics",
    url = "https://aclanthology.org/W15-3049/",
    doi = "10.18653/v1/W15-3049",
    pages = "392--395"
}

@inproceedings{cheong-etal-2021-intrinsic,
    title = "Intrinsic evaluation of language models for code-switching",
    author = "Cheong, Sik Feng  and
      Chieu, Hai Leong  and
      Lim, Jing",
    editor = "Xu, Wei  and
      Ritter, Alan  and
      Baldwin, Tim  and
      Rahimi, Afshin",
    booktitle = "Proceedings of the Seventh Workshop on Noisy User-generated Text (W-NUT 2021)",
    month = nov,
    year = "2021",
    address = "Online",
    publisher = "Association for Computational Linguistics",
    url = "https://aclanthology.org/2021.wnut-1.10/",
    doi = "10.18653/v1/2021.wnut-1.10",
    pages = "81--86",
}

@misc{ugan2025pier,
title={PIER: A Novel Metric for Evaluating What Matters in Code-Switching}, 
      author={Enes Yavuz Ugan and Ngoc-Quan Pham and Leonard Bärmann and Alex Waibel},
      year={2025},
      eprint={2501.09512},
      archivePrefix={arXiv},
      primaryClass={cs.CL},
      url={https://arxiv.org/abs/2501.09512}, 
}

@inproceedings{zhang-etal-2024-enhancing-multilingual,
    title = "Enhancing Multilingual Capabilities of Large Language Models through Self-Distillation from Resource-Rich Languages",
    author = "Zhang, Yuanchi  and
      Wang, Yile  and
      Liu, Zijun  and
      Wang, Shuo  and
      Wang, Xiaolong  and
      Li, Peng  and
      Sun, Maosong  and
      Liu, Yang",
    editor = "Ku, Lun-Wei  and
      Martins, Andre  and
      Srikumar, Vivek",
    booktitle = "Proceedings of the 62nd Annual Meeting of the Association for Computational Linguistics (Volume 1: Long Papers)",
    month = aug,
    year = "2024",
    address = "Bangkok, Thailand",
    publisher = "Association for Computational Linguistics",
    url = "https://aclanthology.org/2024.acl-long.603/",
    doi = "10.18653/v1/2024.acl-long.603",
    pages = "11189--11204",
}

@inproceedings{goloburda2025qorgau,
  author    = {Goloburda, Maksym and Nurkas, Almas and Galym, Aibek and Tazabek, Zhansaya and Mussakhojayeva, Symbat and Shavrina, Tatiana and Artemova, Ekaterina},
  title     = {Qorgau: Evaluating {LLM} Safety in Kazakh-Russian Bilingual Contexts},
  booktitle = {Findings of the Association for Computational Linguistics: ACL 2025},
  year      = {2025}
}

@inproceedings{biswas-etal-2020-semi,
    title = "Semi-supervised Acoustic and Language Model Training for {E}nglish-isi{Z}ulu Code-Switched Speech Recognition",
    author = "Biswas, Astik  and
      de Wet, Febe  and
      van der Westhuizen, Ewald  and
      Niesler, Thomas",
    booktitle = "Proceedings of the 2020 Conference on Computational Approaches to Linguistic Code-Switching",
    month = jul,
    year = "2020",
    address = "Online",
    publisher = "Association for Computational Linguistics",
    url = "https://aclanthology.org/2020.calcs-1.7",
    pages = "61--71"
}

@inproceedings{chi-bell-2022-improving,
    title = "Improving code-switched {ASR} with linguistic information",
    author = "Chi, Jie  and
      Bell, Peter",
    booktitle = "Proceedings of the 29th International Conference on Computational Linguistics",
    month = oct,
    year = "2022",
    address = "Gyeongju, Republic of Korea",
    publisher = "International Committee on Computational Linguistics",
    url = "https://aclanthology.org/2022.coling-1.627",
    pages = "7161--7172"
}

@inproceedings{wang-li-2023-text,
    title = "Text-Derived Language Identity Incorporation for End-to-End Code-Switching Speech Recognition",
    author = "Wang, Qinyi  and
      Li, Haizhou",
    editor = "Winata, Genta  and
      Kar, Sudipta  and
      Zhukova, Marina  and
      Solorio, Thamar  and
      Diab, Mona  and
      Sitaram, Sunayana  and
      Choudhury, Monojit  and
      Bali, Kalika",
    booktitle = "Proceedings of the 6th Workshop on Computational Approaches to Linguistic Code-Switching",
    month = dec,
    year = "2023",
    address = "Singapore",
    publisher = "Association for Computational Linguistics",
    url = "https://aclanthology.org/2023.calcs-1.4/",
    pages = "33--42",
}

@inproceedings{kartik-etal-2024-synthetic,
    title = "Synthetic Data Generation and Joint Learning for Robust Code-Mixed Translation",
    author = "Kartik, Kartik  and
      Soni, Sanjana  and
      Kunchukuttan, Anoop  and
      Chakraborty, Tanmoy  and
      Akhtar, Md. Shad",
    editor = "Calzolari, Nicoletta  and
      Kan, Min-Yen  and
      Hoste, Veronique  and
      Lenci, Alessandro  and
      Sakti, Sakriani  and
      Xue, Nianwen",
    booktitle = "Proceedings of the 2024 Joint International Conference on Computational Linguistics, Language Resources and Evaluation (LREC-COLING 2024)",
    month = may,
    year = "2024",
    address = "Torino, Italia",
    publisher = "ELRA and ICCL",
    url = "https://aclanthology.org/2024.lrec-main.1345/",
    pages = "15480--15492",
}

@inproceedings{amazouz2017addressing,
  title     = {Addressing Code-Switching in French/Algerian Arabic Speech},
  author    = {Djegdjiga Amazouz and Martine Adda-Decker and Lori Lamel},
  year      = {2017},
  booktitle = {Interspeech 2017},
  pages     = {62--66},
  doi       = {10.21437/Interspeech.2017-1373},
  issn      = {2958-1796},

}

@inproceedings{hada-etal-2024-large,
    title = "Are Large Language Model-based Evaluators the Solution to Scaling Up Multilingual Evaluation?",
    author = "Hada, Rishav  and
      Gumma, Varun  and
      de Wynter, Adrian  and
      Diddee, Harshita  and
      Ahmed, Mohamed  and
      Choudhury, Monojit  and
      Bali, Kalika  and
      Sitaram, Sunayana",
    editor = "Graham, Yvette  and
      Purver, Matthew",
    booktitle = "Findings of the Association for Computational Linguistics: EACL 2024",
    month = mar,
    year = "2024",
    address = "St. Julian{'}s, Malta",
    publisher = "Association for Computational Linguistics",
    url = "https://aclanthology.org/2024.findings-eacl.71/",
    pages = "1051--1070",
}

@inproceedings{garg-etal-2021-mipe,
    title = "{MIPE}: A Metric Independent Pipeline for Effective Code-Mixed {NLG} Evaluation",
    author = "Garg, Ayush  and
      Kagi, Sammed  and
      Srivastava, Vivek  and
      Singh, Mayank",
    editor = "Gao, Yang  and
      Eger, Steffen  and
      Zhao, Wei  and
      Lertvittayakumjorn, Piyawat  and
      Fomicheva, Marina",
    booktitle = "Proceedings of the 2nd Workshop on Evaluation and Comparison of NLP Systems",
    month = nov,
    year = "2021",
    address = "Punta Cana, Dominican Republic",
    publisher = "Association for Computational Linguistics",
    url = "https://aclanthology.org/2021.eval4nlp-1.13/",
    doi = "10.18653/v1/2021.eval4nlp-1.13",
    pages = "123--132",
}

@inproceedings{srivastava-singh-2021-challenges,
    title = "Challenges and Limitations with the Metrics Measuring the Complexity of Code-Mixed Text",
    author = "Srivastava, Vivek  and
      Singh, Mayank",
    editor = "Solorio, Thamar  and
      Chen, Shuguang  and
      Black, Alan W.  and
      Diab, Mona  and
      Sitaram, Sunayana  and
      Soto, Victor  and
      Yilmaz, Emre  and
      Srinivasan, Anirudh",
    booktitle = "Proceedings of the Fifth Workshop on Computational Approaches to Linguistic Code-Switching",
    month = jun,
    year = "2021",
    address = "Online",
    publisher = "Association for Computational Linguistics",
    url = "https://aclanthology.org/2021.calcs-1.2/",
    doi = "10.18653/v1/2021.calcs-1.2",
    pages = "6--14",
}

@inproceedings{barman-etal-2014-code,
    title = "Code Mixing: A Challenge for Language Identification in the Language of Social Media",
    author = "Barman, Utsab  and
      Das, Amitava  and
      Wagner, Joachim  and
      Foster, Jennifer",
    editor = "Diab, Mona  and
      Hirschberg, Julia  and
      Fung, Pascale  and
      Solorio, Thamar",
    booktitle = "Proceedings of the First Workshop on Computational Approaches to Code Switching",
    month = oct,
    year = "2014",
    address = "Doha, Qatar",
    publisher = "Association for Computational Linguistics",
    url = "https://aclanthology.org/W14-3902/",
    doi = "10.3115/v1/W14-3902",
    pages = "13--23"
}

@article{fleiss1971measuring,
  title={Measuring nominal scale agreement among many raters.},
  author={Fleiss, Joseph L},
  journal={Psychological bulletin},
  volume={76},
  number={5},
  pages={378},
  year={1971},
  publisher={American Psychological Association}
}

@article{cohen1960coefficient,
  title={A coefficient of agreement for nominal scales},
  author={Cohen, Jacob},
  journal={Educational and psychological measurement},
  volume={20},
  number={1},
  pages={37--46},
  year={1960},
  publisher={Sage Publications Sage CA: Los Angeles, CA}
}

@inproceedings{tatariya-etal-2023-transfer,
    title = "Transfer Learning for Code-Mixed Data: Do Pretraining Languages Matter?",
    author = "Tatariya, Kushal  and
      Lent, Heather  and
      de Lhoneux, Miryam",
    editor = "Barnes, Jeremy  and
      De Clercq, Orph{\'e}e  and
      Klinger, Roman",
    booktitle = "Proceedings of the 13th Workshop on Computational Approaches to Subjectivity, Sentiment, {\&} Social Media Analysis",
    month = jul,
    year = "2023",
    address = "Toronto, Canada",
    publisher = "Association for Computational Linguistics",
    url = "https://aclanthology.org/2023.wassa-1.32/",
    doi = "10.18653/v1/2023.wassa-1.32",
    pages = "365--378",
}

@inproceedings{Zhu2023MixBA,
  title={Mix before Align: Towards Zero-shot Cross-lingual Sentiment Analysis via Soft-Mix and Multi-View Learning},
  author={Zhihong Zhu and Xuxin Cheng and Dongsheng Chen and Zhiqi Huang and Hongxiang Li and Yuexian Zou},
  booktitle={Interspeech},
  year={2023},
  url={https://api.semanticscholar.org/CorpusID:260906484}
}

@inproceedings{winata-etal-2018-code,
    title = "Code-Switching Language Modeling using Syntax-Aware Multi-Task Learning",
    author = "Winata, Genta Indra  and
      Madotto, Andrea  and
      Wu, Chien-Sheng  and
      Fung, Pascale",
    editor = "Aguilar, Gustavo  and
      AlGhamdi, Fahad  and
      Soto, Victor  and
      Solorio, Thamar  and
      Diab, Mona  and
      Hirschberg, Julia",
    booktitle = "Proceedings of the Third Workshop on Computational Approaches to Linguistic Code-Switching",
    month = jul,
    year = "2018",
    address = "Melbourne, Australia",
    publisher = "Association for Computational Linguistics",
    url = "https://aclanthology.org/W18-3207/",
    doi = "10.18653/v1/W18-3207",
    pages = "62--67",
}

@article{yasir2021mixed,
  author    = {Yasir, M. and Chen, L. and Khatoon, A. and Malik, M. A. and Abid, F.},
  title     = {Mixed Script Identification Using Automated DNN Hyperparameter Optimization},
  journal   = {Computational Intelligence and Neuroscience},
  volume    = {2021},
  pages     = {8415333},
  year      = {2021},
  doi       = {10.1155/2021/8415333},
  pmid      = {34925496},
  pmcid     = {PMC8683192},
  publisher = {Hindawi Publishing Corporation},
url={https://pubmed.ncbi.nlm.nih.gov/34925496/}
}

@article{ingle2025ilid,
  title={ILID: Native Script Language Identification for Indian Languages},
  author={Ingle, Yash and Mishra, Pruthwik},
  journal={arXiv preprint arXiv:2507.11832},
  year={2025}
}

@inproceedings{singh-etal-2018-language,
    title = "Language Identification and Named Entity Recognition in {H}inglish Code Mixed Tweets",
    author = "Singh, Kushagra  and
      Sen, Indira  and
      Kumaraguru, Ponnurangam",
    editor = "Shwartz, Vered  and
      Tabassum, Jeniya  and
      Voigt, Rob  and
      Che, Wanxiang  and
      de Marneffe, Marie-Catherine  and
      Nissim, Malvina",
    booktitle = "Proceedings of {ACL} 2018, Student Research Workshop",
    month = jul,
    year = "2018",
    address = "Melbourne, Australia",
    publisher = "Association for Computational Linguistics",
    url = "https://aclanthology.org/P18-3008/",
    doi = "10.18653/v1/P18-3008",
    pages = "52--58",
}

@inproceedings{saumya-etal-2021-offensive,
    title = "Offensive language identification in {D}ravidian code mixed social media text",
    author = "Saumya, Sunil  and
      Kumar, Abhinav  and
      Singh, Jyoti Prakash",
    editor = "Chakravarthi, Bharathi Raja  and
      Priyadharshini, Ruba  and
      Kumar M, Anand  and
      Krishnamurthy, Parameswari  and
      Sherly, Elizabeth",
    booktitle = "Proceedings of the First Workshop on Speech and Language Technologies for Dravidian Languages",
    month = apr,
    year = "2021",
    address = "Kyiv",
    publisher = "Association for Computational Linguistics",
    url = "https://aclanthology.org/2021.dravidianlangtech-1.5/",
    pages = "36--45",
}

@inproceedings{babatunde-etal-2025-beyond,
    title = "Beyond Monolingual Limits: Fine-Tuning Monolingual {ASR} for {Y}oruba-{E}nglish Code-Switching",
    author = "Babatunde, Oreoluwa Boluwatife  and
      Olufemi, Victor Tolulope  and
      Bolarinwa, Emmanuel  and
      Moshood, Kausar Yetunde  and
      Emezue, Chris Chinenye",
    booktitle = "Proceedings of the 7th Workshop on Computational Approaches to Linguistic Code-Switching",
    month = may,
    year = "2025",
    address = "Albuquerque, New Mexico, USA",
    publisher = "Association for Computational Linguistics",
    url = "https://aclanthology.org/2025.calcs-1.3/",
    doi = "10.18653/v1/2025.calcs-1.3",
    pages = "18--25",
    ISBN = "979-8-89176-053-0",
}

@INPROCEEDINGS{10890030,
  author={Zhang, Fengrun and Geng, Wang and Huang, Hukai and Shan, Yahui and Yi, Cheng and Qu, He},
  booktitle={ICASSP 2025 - 2025 IEEE International Conference on Acoustics, Speech and Signal Processing (ICASSP)}, 
  title={Boosting Code-Switching ASR with Mixture of Experts Enhanced Speech-Conditioned LLM}, 
  year={2025},
  volume={},
  number={},
  pages={1-5},
  doi={10.1109/ICASSP49660.2025.10890030}}

@inproceedings{arindam-etal-2023-lost,
    title = "Lost in Translation No More: Fine-tuned transformer-based models for {C}ode{M}ix to {E}nglish Machine Translation",
    author = "Chatterjee, Arindam  and
      Sharma, Chhavi  and
      V.p., Yashwanth  and
      Kumar, Niraj  and
      Raj, Ayush  and
      Ekbal, Asif",
    editor = "D. Pawar, Jyoti  and
      Lalitha Devi, Sobha",
    booktitle = "Proceedings of the 20th International Conference on Natural Language Processing (ICON)",
    month = dec,
    year = "2023",
    address = "Goa University, Goa, India",
    publisher = "NLP Association of India (NLPAI)",
    url = "https://aclanthology.org/2023.icon-1.25/",
    pages = "326--335",
}

@inproceedings{palomino-ochoa-luna-2020-palomino,
    title = "Palomino-Ochoa at {S}em{E}val-2020 Task 9: Robust System Based on Transformer for Code-Mixed Sentiment Classification",
    author = "Palomino, Daniel  and
      Ochoa-Luna, Jos{\'e}",
    editor = "Herbelot, Aurelie  and
      Zhu, Xiaodan  and
      Palmer, Alexis  and
      Schneider, Nathan  and
      May, Jonathan  and
      Shutova, Ekaterina",
    booktitle = "Proceedings of the Fourteenth Workshop on Semantic Evaluation",
    month = dec,
    year = "2020",
    address = "Barcelona (online)",
    publisher = "International Committee for Computational Linguistics",
    url = "https://aclanthology.org/2020.semeval-1.124/",
    doi = "10.18653/v1/2020.semeval-1.124",
    pages = "963--967",
}

@inproceedings{wu-etal-2020-meistermorxrc,
    title = "{M}eister{M}orxrc at {S}em{E}val-2020 Task 9: Fine-Tune Bert and Multitask Learning for Sentiment Analysis of Code-Mixed Tweets",
    author = "Wu, Qi  and
      Wang, Peng  and
      Huang, Chenghao",
    editor = "Herbelot, Aurelie  and
      Zhu, Xiaodan  and
      Palmer, Alexis  and
      Schneider, Nathan  and
      May, Jonathan  and
      Shutova, Ekaterina",
    booktitle = "Proceedings of the Fourteenth Workshop on Semantic Evaluation",
    month = dec,
    year = "2020",
    address = "Barcelona (online)",
    publisher = "International Committee for Computational Linguistics",
    url = "https://aclanthology.org/2020.semeval-1.174/",
    doi = "10.18653/v1/2020.semeval-1.174",
    pages = "1294--1297",
}

@inproceedings{khan-etal-2022-sit,
    title = "{SIT} at {M}ix{MT} 2022: Fluent Translation Built on Giant Pre-trained Models",
    author = "Khan, Abdul  and
      Kanade, Hrishikesh  and
      Budhrani, Girish  and
      Jhanglani, Preet  and
      Xu, Jia",
    booktitle = "Proceedings of the Seventh Conference on Machine Translation (WMT)",
    month = dec,
    year = "2022",
    address = "Abu Dhabi, United Arab Emirates (Hybrid)",
    publisher = "Association for Computational Linguistics",
    url = "https://aclanthology.org/2022.wmt-1.114/",
    pages = "1136--1144",

}

@article{sitaram2019survey,
  title={A survey of code-switched speech and language processing},
  author={Sitaram, Sunayana and Chandu, Khyathi Raghavi and Rallabandi, Sai Krishna and Black, Alan W},
  journal={arXiv preprint arXiv:1904.00784},
  year={2019}
}

@article{RATHNAYAKE2024107239,
title = {AdapterFusion-based multi-task learning for code-mixed and code-switched text classification},
journal = {Engineering Applications of Artificial Intelligence},
volume = {127},
pages = {107239},
year = {2024},
issn = {0952-1976},
doi = {https://doi.org/10.1016/j.engappai.2023.107239},
url = {https://www.sciencedirect.com/science/article/pii/S0952197623014239},
author = {Himashi Rathnayake and Janani Sumanapala and Raveesha Rukshani and Surangika Ranathunga},
}

@misc{preply2022,
  author = {Preply},
  title = {Bilingualism Statistics 2022: Facts and Figures},
  year = {2022},
  url = {https://preply.com/en/blog/bilingualism-statistics/},
}

@misc{rosettastone2025,
  author = {Rosetta Stone},
  title = {How Many People Are Trilingual? (Plus, Is It Worth It?)},
  year = {2025},
  url = {https://blog.rosettastone.com/trilingual/},
}

@inproceedings{terblanche-etal-2024-prompting,
    title = "Prompting towards Alleviating Code-Switched Data Scarcity in Under-Resourced Languages with {GPT} as a Pivot",
    author = "Terblanche, Michelle  and
      Olaleye, Kayode  and
      Marivate, Vukosi",
    editor = "Melero, Maite  and
      Sakti, Sakriani  and
      Soria, Claudia",
    booktitle = "Proceedings of the 3rd Annual Meeting of the Special Interest Group on Under-resourced Languages @ LREC-COLING 2024",
    month = may,
    year = "2024",
    address = "Torino, Italia",
    publisher = "ELRA and ICCL",
    url = "https://aclanthology.org/2024.sigul-1.33/",
    pages = "272--282",
}

@inproceedings{zhang-etal-2023-multilingual,
    title = "Multilingual Large Language Models Are Not (Yet) Code-Switchers",
    author = "Zhang, Ruochen  and
      Cahyawijaya, Samuel  and
      Cruz, Jan Christian Blaise  and
      Winata, Genta  and
      Aji, Alham Fikri",
    editor = "Bouamor, Houda  and
      Pino, Juan  and
      Bali, Kalika",
    booktitle = "Proceedings of the 2023 Conference on Empirical Methods in Natural Language Processing",
    month = dec,
    year = "2023",
    address = "Singapore",
    publisher = "Association for Computational Linguistics",
    url = "https://aclanthology.org/2023.emnlp-main.774/",
    doi = "10.18653/v1/2023.emnlp-main.774",
    pages = "12567--12582",
}

@inproceedings{laureano-de-leon-etal-2024-code,
    title = "Code-Mixed Probes Show How Pre-Trained Models Generalise on Code-Switched Text",
    author = "Laureano De Leon, Frances Adriana  and
      Tayyar Madabushi, Harish  and
      Lee, Mark",
    editor = "Calzolari, Nicoletta  and
      Kan, Min-Yen  and
      Hoste, Veronique  and
      Lenci, Alessandro  and
      Sakti, Sakriani  and
      Xue, Nianwen",
    booktitle = "Proceedings of the 2024 Joint International Conference on Computational Linguistics, Language Resources and Evaluation (LREC-COLING 2024)",
    month = may,
    year = "2024",
    address = "Torino, Italia",
    publisher = "ELRA and ICCL",
    url = "https://aclanthology.org/2024.lrec-main.307/",
    pages = "3457--3468",
}

@inproceedings{litschko-etal-2023-boosting,
    title = "Boosting Zero-shot Cross-lingual Retrieval by Training on Artificially Code-Switched Data",
    author = "Litschko, Robert  and
      Artemova, Ekaterina  and
      Plank, Barbara",
    editor = "Rogers, Anna  and
      Boyd-Graber, Jordan  and
      Okazaki, Naoaki",
    booktitle = "Findings of the Association for Computational Linguistics: ACL 2023",
    month = jul,
    year = "2023",
    address = "Toronto, Canada",
    publisher = "Association for Computational Linguistics",
    url = "https://aclanthology.org/2023.findings-acl.193/",
    doi = "10.18653/v1/2023.findings-acl.193",
    pages = "3096--3108",
}

@inproceedings{gaschi-etal-2023-code,
    title = "Code-switching as a cross-lingual Training Signal: an Example with Unsupervised Bilingual Embedding",
    author = "Gaschi, Felix  and
      El-Baamrani, Ilias  and
      Gendron, Barbara  and
      Rastin, Parisa  and
      Toussaint, Yannick",
    editor = "Ataman, Duygu",
    booktitle = "Proceedings of the 3rd Workshop on Multi-lingual Representation Learning (MRL)",
    month = dec,
    year = "2023",
    address = "Singapore",
    publisher = "Association for Computational Linguistics",
    url = "https://aclanthology.org/2023.mrl-1.16/",
    doi = "10.18653/v1/2023.mrl-1.16",
    pages = "208--217"
}

@inproceedings{feng-etal-2022-toward,
    title = "Toward the Limitation of Code-Switching in Cross-Lingual Transfer",
    author = "Feng, Yukun  and
      Li, Feng  and
      Koehn, Philipp",
    editor = "Goldberg, Yoav  and
      Kozareva, Zornitsa  and
      Zhang, Yue",
    booktitle = "Proceedings of the 2022 Conference on Empirical Methods in Natural Language Processing",
    month = dec,
    year = "2022",
    address = "Abu Dhabi, United Arab Emirates",
    publisher = "Association for Computational Linguistics",
    url = "https://aclanthology.org/2022.emnlp-main.400/",
    doi = "10.18653/v1/2022.emnlp-main.400",
    pages = "5966--5971",
}

@inproceedings{micallef-etal-2024-cross,
    title = "Cross-Lingual Transfer from Related Languages: Treating Low-Resource {M}altese as Multilingual Code-Switching",
    author = "Micallef, Kurt  and
      Habash, Nizar  and
      Borg, Claudia  and
      Eryani, Fadhl  and
      Bouamor, Houda",
    editor = "Graham, Yvette  and
      Purver, Matthew",
    booktitle = "Proceedings of the 18th Conference of the European Chapter of the Association for Computational Linguistics (Volume 1: Long Papers)",
    month = mar,
    year = "2024",
    address = "St. Julian{'}s, Malta",
    publisher = "Association for Computational Linguistics",
    url = "https://aclanthology.org/2024.eacl-long.61/",
    doi = "10.18653/v1/2024.eacl-long.61",
    pages = "1014--1025",
}

@inproceedings{pratapa-choudhury-2021-comparing,
    title = "Comparing Grammatical Theories of Code-Mixing",
    author = "Pratapa, Adithya  and
      Choudhury, Monojit",
    editor = "Xu, Wei  and
      Ritter, Alan  and
      Baldwin, Tim  and
      Rahimi, Afshin",
    booktitle = "Proceedings of the Seventh Workshop on Noisy User-generated Text (W-NUT 2021)",
    month = nov,
    year = "2021",
    address = "Online",
    publisher = "Association for Computational Linguistics",
    url = "https://aclanthology.org/2021.wnut-1.18/",
    doi = "10.18653/v1/2021.wnut-1.18",
    pages = "158--167",
}

@inproceedings{sterner-teufel-2025-code,
    title = "Code-Switching and Syntax: A Large-Scale Experiment",
    author = "Sterner, Igor  and
      Teufel, Simone",
    editor = "Che, Wanxiang  and
      Nabende, Joyce  and
      Shutova, Ekaterina  and
      Pilehvar, Mohammad Taher",
    booktitle = "Findings of the Association for Computational Linguistics: ACL 2025",
    month = jul,
    year = "2025",
    address = "Vienna, Austria",
    publisher = "Association for Computational Linguistics",
    url = "https://aclanthology.org/2025.findings-acl.600/",
    doi = "10.18653/v1/2025.findings-acl.600",
    pages = "11526--11533",
    ISBN = "979-8-89176-256-5",
}

@inproceedings{potter-yuan-2024-llm,
    title = "{LLM}-based Code-Switched Text Generation for Grammatical Error Correction",
    author = "Potter, Tom  and
      Yuan, Zheng",
    editor = "Al-Onaizan, Yaser  and
      Bansal, Mohit  and
      Chen, Yun-Nung",
    booktitle = "Proceedings of the 2024 Conference on Empirical Methods in Natural Language Processing",
    month = nov,
    year = "2024",
    address = "Miami, Florida, USA",
    publisher = "Association for Computational Linguistics",
    url = "https://aclanthology.org/2024.emnlp-main.942/",
    doi = "10.18653/v1/2024.emnlp-main.942",
    pages = "16957--16965",
}

@inproceedings{pratapa-etal-2018-language,
    title = "Language Modeling for Code-Mixing: The Role of Linguistic Theory based Synthetic Data",
    author = "Pratapa, Adithya  and
      Bhat, Gayatri  and
      Choudhury, Monojit  and
      Sitaram, Sunayana  and
      Dandapat, Sandipan  and
      Bali, Kalika",
    editor = "Gurevych, Iryna  and
      Miyao, Yusuke",
    booktitle = "Proceedings of the 56th Annual Meeting of the Association for Computational Linguistics (Volume 1: Long Papers)",
    month = jul,
    year = "2018",
    address = "Melbourne, Australia",
    publisher = "Association for Computational Linguistics",
    url = "https://aclanthology.org/P18-1143/",
    doi = "10.18653/v1/P18-1143",
    pages = "1543--1553",
}

@misc{kellert2025parsing,
title={Parsing the Switch: LLM-Based UD Annotation for Complex Code-Switched and Low-Resource Languages}, 
      author={Olga Kellert and Nemika Tyagi and Muhammad Imran and Nelvin Licona-Guevara and Carlos Gómez-Rodríguez},
      year={2025},
      eprint={2506.07274},
      archivePrefix={arXiv},
      primaryClass={cs.CL},
      url={https://arxiv.org/abs/2506.07274}
}

@inproceedings{Arora2020CrosslingualTL,
  title={Cross-lingual Transfer Learning for Intent Detection of Covid-19 Utterances},
  author={Abhinav Arora and Akshat Shrivastava and Lorena Sainz-Maza Lecanda},
  booktitle={Proceedings of the 28th International Conference on Computational Linguistics (COLING 2020)},
  year={2020},
  url={https://api.semanticscholar.org/CorpusID:221658006}
}

@inproceedings{das-etal-2023-improving,
    title = "Improving Pretraining Techniques for Code-Switched {NLP}",
    author = "Das, Richeek  and
      Ranjan, Sahasra  and
      Pathak, Shreya  and
      Jyothi, Preethi",
    editor = "Rogers, Anna  and
      Boyd-Graber, Jordan  and
      Okazaki, Naoaki",
    booktitle = "Proceedings of the 61st Annual Meeting of the Association for Computational Linguistics (Volume 1: Long Papers)",
    month = jul,
    year = "2023",
    address = "Toronto, Canada",
    publisher = "Association for Computational Linguistics",
    url = "https://aclanthology.org/2023.acl-long.66/",
    doi = "10.18653/v1/2023.acl-long.66",
    pages = "1176--1191",
}

\appendix

\section{Methodology}
\label{sec:appendix}
This section outlines the methodology adopted to identify, review, and categorize literature relevant to this survey on code-switching in the era of LLMs. The goal was to capture key trends, modeling techniques, datasets, benchmarks and challenges across NLP tasks rather than conduct an exhaustive systematic review. The approach follows established survey practices \citep{kinney2023semantic}.

\paragraph{Paper Selection} We began by defining a set of search keywords targeting three core dimensions: code-mixing/code-switching, multilingual NLP, and large language models. To ensure broad linguistic coverage, the search encompassed major bilingual and multilingual language pairs documented in prior research and repositories (e.g., CoVoSwitch, GLUECoS, LinCE). Using these keywords, we queried the ACL Anthology, arXiv and Semantic Scholar databases via their APIs, with a search cutoff date of October 2025, consistent with ACL Rolling Review’s recency guidelines. This process initially retrieved around ~500 papers.

\paragraph{Screening and Filtering} Duplicate entries were removed using DOIs and titles, prioritizing peer-reviewed sources. The remaining papers were manually screened for relevance. A study was included if it addressed code-mixing or code-switching within any NLP task, or explored multilingual and LLM-based adaptation methods. This screening resulted in a refined set of 327 papers, covering both pre-LLM and LLM-era research.

\paragraph{Categorization} Selected papers were categorized by \textit{\textbf{(i)}} task type (e.g., Language Identification, POS Tagging, NER, Intent, Speech/ASR, MT), \textit{\textbf{(ii)}} modality (text, speech, vision-language), and \textit{\textbf{(iii)}} model architecture (transformer-based, instruction-tuned, multimodal). When overlaps occurred (e.g., between translation and generation), we retained the category most central to the contribution. Dataset coverage, annotation methods, and language pairs were systematically verified to map diversity and resource availability. High-, mid-, and low-resource classifications followed conventions in multilingual NLP research.

This multi-stage process of search, screening, and categorization produced ~\textbf{\total{unique_references}} papers forming the foundation of this survey, spanning \textbf{15 NLP tasks}, \textbf{30+ datasets}, and \textbf{80+ languages}.

\paragraph{Handling Multi-Category Papers and Reporting Gaps}
Papers spanning multiple categories or tasks (common in recent LLM-era work) are assigned multi-label categories and discussed in the most relevant primary section, with cross-references provided where appropriate. Reported performance gaps are presented as representative values directly quoted or reported from individual primary studies on their respective benchmarks. This approach preserves representativeness and reflects the interdisciplinary nature of modern code-switching research (e.g., a single work advancing both synthetic data generation and fine-tuning).

\section{Taxonomy}
\label{sec:Taxonomy}

In this section, we elaborate on the taxonomy of code-mixed language analytics introduced in Figure~\ref{fig:lit_survey} that provides an analytical overview of the CSW research landscape across four dimensions: (A) task maturity, showing saturation in traditional tasks (e.g., LID, POS) and persistent gaps in reasoning and multimodal settings; (B) methodological evolution from statistical models to LLM-based approaches; (C) language-pair coverage, revealing a strong 72\% English-centric bias; and (D) performance gaps that grow with task complexity, from ~4\% degradation in LID to over 33\% in reasoning-heavy tasks. An interactive taxonomy offers paper-level details. 

The rapid evolution of code-switching (CSW) research in the LLM era requires a comprehensive framework capturing methodological diversity and task complexity. Our framework supports scalable LLM approaches while distinguishing various contribution types. 

This structure reflects the interconnected nature of modern CSW research, emphasizing the shift from language-pair-specific solutions to unified multilingual architectures and the need for integrated, end-to-end CSW systems.

\begin{figure*}[htp]
    \centering 
    \makebox[\textwidth][c]{
    \hspace*{-0.4cm}
    \newcommand{\leafnodespacing}{0.3cm}
    \def\diagramfontsize{\fontsize{20pt}{26pt}\selectfont }
    \tikzset{
         main-node/.style={
            rounded rectangle,
            rounded rectangle west arc=none,
            rounded rectangle east arc=none,
            fill=gray!20,
            draw=gray!50,
            thick,
            line width=2pt,
            text=black,
            minimum width=3cm, 
            minimum height=1.2cm,
            font=\diagramfontsize\bfseries,
            text width=12cm,
            align=center
        },
        base-node/.style={
            rectangle,
            rounded corners=10pt, 
            draw,
            thick,
            line width=2pt,
            text=black,
            minimum width=6cm, 
            minimum height=1.75cm,
            font=\diagramfontsize\bfseries,
            align=center,
            inner sep=8pt
        },
        leaf-node/.style={
            rectangle,
            rounded corners=10pt, 
            draw,
            thick,
            align=justify,
            text=black,
            line width=2pt,
            minimum width=10cm, 
            minimum height=1.5cm,
            font=\diagramfontsize,
            text width=50cm,
            inner sep=6pt
        },
        connector/.style={
            draw=black,
            thick
        },
        orange/.style={base-node, fill=orange!20, draw=orange!60},
        leaf-orange/.style={leaf-node, fill=orange!20, draw=orange!60},
        purple/.style={base-node, fill=purple!20, draw=purple!60},
        leaf-purple/.style={leaf-node, fill=purple!20, draw=purple!60},
        lblue/.style={base-node, fill=blue!20, draw=blue!60},
        leaf-lblue/.style={leaf-node, fill=blue!20, draw=blue!60},
        blue/.style={base-node, fill=cyan!20, draw=cyan!70},
        leaf-blue/.style={leaf-node, fill=cyan!20, draw=cyan!70},
        green/.style={base-node, fill=green!20, draw=green!60},
        leaf-green/.style={leaf-node, fill=green!20, draw=green!60},
    }

    \resizebox {0.9\textwidth}{0.9\textheight}{

    }
    }
    \caption{A unifying taxonomy of the code-switching research landscape. \textit{\textbf{Takeaway} The mind map contextualizes recent LLM-based advances, revealing continuities, shifts, and unresolved challenges across the CSW literature.} }   
    \label{fig:lit_survey}
\end{figure*}

\paragraph{Code-Switched Language Analytics}

\subsection{Code-Switching Task Landscape}

\paragraph{Foundational Tasks}
These represent core NLP taskcompetencies essential for understanding code-switched text analysisstructure and linguistic properties. They form the foundationbase layer upon which more complex applications are built.

\begin{itemize}
    \item \textbf{Language Identification}: Identifies language boundaries at the word or token level, forming the basis for downstream analysis in mixed-language text  Detects language boundaries at word/token level, including Hope and Offensive text detection.
    \item \textbf{Part-of-Speech Tagging}: Assigns grammatical categories to code-mixed tokens, accounting for syntactic ambiguity and structural variation at switch points.
    \item \textbf{Named Entity Recognition}: Detects and classifies named entities across language boundaries, addressing challenges such as transliteration, script variation, and cross-lingual ambiguity.
    \item \textbf{Machine Translation}: Translates code-switched input into a single target language, requiring joint modeling of mixed-language syntax and semantics.
    \item \textbf{Syntactic Analysis}: Parses the grammatical structure of code-switched sentences to analyze well-formedness and linguistic constraints governing switching behavior.
    \item \textbf{Sentiment and Emotion Analysis}: Models affective meaning across languages, including aspect-based sentiment and multi-label emotion classification.
    \item \textbf{Machine Translation}: Translates code-switched text to monolingual output, requiring understanding of mixed-language syntax.
\end{itemize}

\paragraph{Emerging and Contemporary Tasks}
Emerging tasks extend beyond surface-level analysis to capture semantic interpretation, pragmatic reasoning, and contextual understanding in code-switched settings.

\begin{itemize}
    \item \textbf{Natural Language Inference}: Evaluates entailment and contradiction between code-switched premise–hypothesis pairs.
    \item \textbf{Question Answering}: Supports information retrieval and reasoning over code-switched queries and documents.
    \item \textbf{Intent Classification}: Infers speaker intent in mixed-language conversational inputs, particularly relevant for dialogue and assistant systems.
    \item \textbf{Code-Mixed Text Generation}: Generates linguistically and sociolinguistically plausible code-switched text, often used for data augmentation and dialogue systems.
    \item \textbf{Cross-lingual Transfer}: Exploits code-switching to improve generalization across languages, including transfer to unseen or low-resource language pairs.
    \item \textbf{Text Summarization}: Produces abstractive or extractive summaries while preserving semantic content and, where relevant, code-mixing patterns.
    \item \textbf{Transliteration}: Converts text across scripts while maintaining phonetic fidelity in mixed-language contexts.
\end{itemize}

\paragraph{Underexplored and Frontier Tasks}
These tasks represent comparatively underexplored directions where code-switching intersects with safety, reasoning, creativity, and multimodal understanding.

\begin{itemize}
    \item \textbf{Conversational and Speech}: Includes dialogue generation, customer support agents, and ASR systems operating on naturally occurring code-switched speech.
    \item \textbf{Safety and Multimodal}: Addresses multilingual safety alignment, jailbreaking, and image–text interactions in code-mixed settings.
    \item \textbf{Reasoning and Abstraction}: Examines causal, analogical, and metaphorical reasoning across languages within a single utterance or discourse.
    \item \textbf{Creative and Code Generation}: Covers programming code generation from code-mixed prompts and creative language use such as wordplay and homophonic mixing.
\end{itemize}

\subsection{Datasets and Resources}

\paragraph{Datasets}
Datasets form the backbone of code-switched language research, enabling empirical evaluation across diverse languages, domains, and modalities.

\begin{itemize}
    \item \textbf{Low-Resource Coverage}: Targeted datasets addressing underrepresented language pairs, including African, Dravidian, and Central Asian languages.
    \item \textbf{Multilingual Coverage}: Large-scale corpora spanning multiple language families, scripts, and sociolinguistic contexts.
    \item \textbf{Synthetic Data}: Promptly generated corpora leveraging linguistic constraints or LLMs to augment scarce real-world data.
\end{itemize}

\paragraph{Frameworks and Toolkits}
Frameworks and toolkits support standardized annotation, data generation, and experimentation in code-switched settings.

\begin{itemize}
    \item \textbf{Annotation Frameworks}: Tools for annotating code-mixed text, speech, and multimodal data with support for human-in-the-loop workflows.
    \item \textbf{Synthetic Data Generation Toolkits}: Tools designed to generate synthetic code-mixed corpora through rule-based constraints, transliteration models, and language model–based augmentation.
\end{itemize}

\subsection{Model Training and Adaptation}

\paragraph{Pre-training Approaches}
Pre-training strategies aim to encode code-switching phenomena directly into model representations, often outperforming generic multilingual pre-training on CSW tasks.

\begin{itemize}
    \item \textbf{Specialized Code-Mixed Models}: Architectures trained explicitly on code-mixed corpora with switch-point awareness.
    \item \textbf{Task-Adaptive Pre-training}: Domain-specific and task-specific adaptation using masked language modeling and alignment-aware objectives.
    \item \textbf{Cross-lingual Alignment}: Representation alignment and continual learning techniques to improve multilingual generalization.
\end{itemize}

\paragraph{Fine-tuning Approaches}
Fine-tuning methods adapt pre-trained models to specific tasks while incorporating code-switching-aware objectives.

\begin{itemize}
    \item \textbf{Task-specific Fine-tuning}: Staged training of a model (or adapters) dedicated to a single task/language pair.
    \item \textbf{Multi-task Fine-tuning}:  Joint training of a single model on multiple tasks/language pairs simultaneously with shared parameters to enable knowledge transfer.
    \item \textbf{Instruction Tuning}: Instruction-following adaptation using code-mixed prompts and responses.
    \item \textbf{Parameter-efficient Methods}: Lightweight adaptation techniques such as LoRA, prompt tuning, and quantization-aware training.
    \item \textbf{Reinforcement Learning}: Reward-based optimization for improving fluency and naturalness in code-switched generation.
\end{itemize}

\paragraph{Post-training and Inference-time Adaptation}
These approaches enable generalization in low-resource settings without extensive labeled data.

\begin{itemize}
    \item \textbf{Zero-, One-, and Few-shot Learning}: Prompt-based and retrieval-augmented methods for code-switched tasks under minimal supervision.
    \item \textbf{Instance-based Prompting}: In-context learning approaches that leverage curated or automatically selected code-mixed examples to guide model behavior at inference time.
\end{itemize}

\subsection{Evaluation and Benchmarking}

\paragraph{Benchmarks}
Benchmarks provide standardized evaluation protocols for measuring progress across tasks and domains.

\begin{itemize}
    \item \textbf{Comprehensive Benchmarks}: Multi-task suites covering both traditional and emerging code-switched NLP tasks.
    \item \textbf{Domain-specific Corpora}: Evaluation datasets tailored to domains such as social media, healthcare, agriculture, and multimodal content.
\end{itemize}

\paragraph{Evaluation Metrics}
Evaluation metrics aim to capture both task performance and code-switching-specific linguistic properties.

\begin{itemize}
    \item \textbf{Traditional Metrics}: Standard NLP measures such as accuracy, F1, BLEU, ROUGE, and METEOR.
   \item \textbf{Code-switching-specific Metrics}: Measures that quantify mixing intensity, syntactic diversity, and switch-point accuracy in mixed-language text.
   \item \textbf{Task-specific Metrics}: Evaluation measures tailored to individual tasks, accounting for script variation, phonetic ambiguity, and speech recognition errors.
   \item \textbf{Quality Assessment}: Human judgments of fluency, semantic preservation, and naturalness across languages.
   \item \textbf{Intrinsic Evaluation}: Gold-reference-independent metrics for assessing grammaticality, fluency, and distributional consistency.
\end{itemize}

\subsection{Multi- and Cross-modal Applications}

\paragraph{Speech Processing}
Code-switching in speech introduces phonetic and acoustic variability that challenges conventional speech models.

\begin{itemize}
    \item \textbf{Speech Translation}: Systems that integrate automatic speech recognition and machine translation for processing mixed-language speech input.
    \item \textbf{End-to-End ASR}: Direct modeling of code-switched speech using data augmentation strategies and expert-based or modular architectures.
    \item \textbf{Audio-Visual Recognition}: Multimodal approaches that combine acoustic signals with visual cues to improve recognition robustness.
\end{itemize}

\paragraph{Vision–Language Processing}
Vision-language tasks extend code-switching to multimodal contexts.

\begin{itemize}
    \item \textbf{Visual Question Answering}:mage-based reasoning with mixed-language questions and captions.
    \item \textbf{Multimodal Systems}: Joint visual–text processing for multilingual and code-switched documents.
\end{itemize}

\paragraph{Cross-modal Integration}
Cross-modal approaches aim to unify representations across text, speech, and vision.

\begin{itemize}
    \item \textbf{Phonetic Processing}: Script conversion and phonetic embeddings for mixed-script languages.
    \item \textbf{Multimodal Fusion}: Joint audio–visual–text models for affective analysis and safety-related tasks.
\end{itemize}

\section{Code-Switching Task Landscape: Capabilities and Gaps}

\subsection{Traditional Tasks}
\label{sec:appendix-tasks1}
\paragraph{Language Identification}
Script detection remains crucial for accurate token-level processing, with Bi-GRU architectures achieving 90.17\% accuracy on Roman Urdu, Hindi, Saraiki, Bengali, and English using GloVe embeddings \citep{yasir2021mixed}. The ILID corpus provides 250K sentences across 25 scripts and 23 languages, including dual-script instances for Manipuri and Sindhi \citep{ingle2025ilid}. Character n-gram TF-IDF features (1--6 grams) have proven effective for Dravidian script-mixed social media text \citep{saumya-etal-2021-offensive}. Shared-task initiatives such as LT-EDI-EACL extended hope speech detection to English, Malayalam--English, and Tamil--English, where TF-IDF features combined with MuRIL embeddings achieved F1 scores of 0.92, 0.75, and 0.57 respectively \citep{dave-etal-2021-irnlp-daiict}. Specialized datasets such as KanHope (English--Kannada) highlight persistent issues of class imbalance and preprocessing challenges involving emojis and multilingual tokens \citep{hande2021hope}. Overall, methodological advances have transitioned from traditional machine learning to transformer-based architectures, where task-adaptive pre-training and multilingual contextual embeddings substantially improve performance, particularly in low-resource and morphologically rich languages \citep{jayanthi-gupta-2021-sj, shanmugavadivel2022deep}. \textbf{Offensive Language Identification} in code-switched text presents unique challenges, as users often employ strategic language alternation to bypass keyword-based moderation. Foundational datasets such as OffMix-3L establish trilingual benchmarks for Bangla--English--Hindi, underscoring the difficulty of handling transliterated content where phonetic variation hinders detection accuracy \citep{goswami-etal-2023-offmix, sazzed-2021-abusive}. Transformer-based systems such as COOLI explicitly target adversarial switching strategies, while synthetic code-switched data generation has emerged as a promising avenue for building linguistically diverse and robust training corpora \citep{balouchzahi-etal-2021-mucs-lt, salaam-etal-2022-offensive}. Recent paradigms incorporate transfer and multi-task learning, with approaches such as SetFit enabling efficient few-shot adaptation for Tamil--English detection, and multi-task frameworks demonstrating strong performance across zero-shot and fine-tuning scenarios for harmful multimodal content \citep{pannerselvam-etal-2024-setfit, kumar2025multi}.


\paragraph{Sentiment \& Emotion Analysis} ihas been extensively studied in CSW settings, with shared tasks like SemEval (Sentiment Analysis for Code-Mixed Social Media Text), where fine-tuned multilingual transformers achieved strong performance on Hinglish and Spanglish datasets via strategies such as focal loss for class imbalance (XLM-R) \citep{ma2020xlp}, straightforward mBERT fine-tuning \citep{palomino-ochoa-luna-2020-palomino}, RoBERTa fine-tuning \citep{sultan2020wessa}, stacked ensembling of BiLSTM and BERT variants \citep{singh2020voice}, and multi-task learning with BERT \citep{wu-etal-2020-meistermorxrc}; similar approaches were applied in NLP-CIC systems \citep{angel2020nlp}. Research has expanded to diverse language pairs, including Dravidian languages (Tamil-English, Malayalam-English, Kannada-English) \citep{chakravarthi2022dravidiancodemix}, Indonesian and Vietnamese-English \citep{winata-etal-2023-nusax, van2022sentiment}, Kenyan Sheng-English slang \citep{etori2025rideke}, Bengali-English trilingual sentiment \citep{raihan-etal-2023-sentmix}, and emotion-specific trilingual analysis \citep{raihan-etal-2024-emomix}. Multi-label emotion detection frameworks support fine-grained analysis across CSW texts \citep{wadhawan2021towards}, while cross-lingual aspect-based sentiment analysis (ABSA) leverages shared representations for improved transfer \citep{zhang-etal-2021-cross}. Data scarcity challenges are mitigated through unsupervised self-training on unlabeled CSW data \citep{jayanthi2021unsupervised}, progressive curriculum learning with increasing mixing intensity \citep{sharma2022progressive}, integration of monolingual resources \citep{kumar-etal-2022-utilizing}, and synthetic code-switched augmentation via CoSDA-ML, yielding consistent zero-shot gains across multiple tasks \citep{ijcai2020p0533}. Large language models enable effective zero-shot sentiment classification through translation-based pipelines \citep{10938193}, multilingual RLAIF for preference alignment \citep{zhang-etal-2023-multilingual}, and efficient synthetic data leveraging for downstream sentiment tasks \citep{zeng-2024-leveraging}. Harmful content detection has advanced with datasets targeting Bangla-English offensive language and Devanagari-script hate speech \citep{raihan-etal-2023-offensive}, where parameter-efficient fine-tuning (PEFT) and SetFit embeddings achieve competitive results on low-resource CSW hate speech \citep{sidibomma-etal-2025-llmsagainsthate, pannerselvam-etal-2024-setfit}.


\paragraph{Syntactic Analysis} in CSW has shifted from structural modeling to theory-guided methods, improving parsing and evaluation. SyMCoM introduced a syntactic measure of code-mixing based on POS tags for English-Hindi, enabling dataset comparison and highlighting variations in open/closed class contributions \citep{kodali-etal-2022-symcom}. Syntax-aware multi-task LSTMs jointly trained on language modeling and parsing significantly reduced perplexity on Mandarin-English code-switched data \citep{winata-etal-2018-code}. Synthetic treebanks generated via annotation projection improved dependency parsing performance for Bengali-English \citep{ghosh-etal-2019-dependency}. CoMix leveraged phonetic and POS-guided pre-training to advance Hinglish machine translation and NER \citep{arora-etal-2023-comix}. Linguistically constrained generation following the Equivalence Constraint produced more natural code-mixed text compared to heuristic baselines \citep{pratapa-choudhury-2021-comparing}. LLMs facilitated Universal Dependencies annotation for low-resource pairs like Spanglish and Spanish–Guaraní \citep{kellert2025parsing}, while large-scale experiments demonstrated strong syntactic alignment in CSW with monolingual parses \citep{sterner-teufel-2025-code, laureano-de-leon-etal-2024-code}. Non-English prompting enhanced LLM grammaticality judgments \citep{behzad-etal-2024-ask}, and LLM-based grammatical error correction performed well on learner corpora \citep{potter-yuan-2024-llm}. Despite these advances, enforcing universal syntactic constraints across typologically diverse languages remains difficult, often leading to unnatural switches or reduced fluency in generated text \citep{pratapa-choudhury-2021-comparing}.

\paragraph{Machine Translation} in CSW contexts has evolved from statistical to neural paradigms, addressing irregular switching and data scarcity. Pioneering works used code-switching as augmentation to enforce lexical constraints in standard NMT by replacing source phrases with target translations to teach copying \citep{song-etal-2025-multilingual}, while PhraseOut advanced controlled mixing via phrase-level replacement for multilingual low-resource scenarios \citep{jasim-etal-2020-phraseout}. Back-to-back translation improved Hinglish MT, while unsupervised approaches with linguistic heuristics enhanced Sinhala–English corpora \citep{tarunesh-etal-2021-machine, kugathasan-sumathipala-2021-neural}. CoSDA-ML scaled dynamic multi-language code-switching augmentation by word substitution from bilingual dictionaries to fine-tune mBERT for zero-shot cross-lingual alignment across diverse tasks \citep{ijcai2020p0533}, and CoMeT/back-translation with COMET filtering produced higher-quality synthetic parallel data for Indic/Hinglish pairs by concatenating monolingual sentences and transliterating roman script \citep{gautam-etal-2021-comet}. Gated seq2seq architectures with explicit language tags \citep{dowlagar-mamidi-2021-gated}, fine-tuned mT5 for Hinglish \citep{nagoudi-etal-2021-investigating}, and mBART overcoming orthographic challenges in MSA-Egyptian–English \citep{nagoudi-etal-2021-investigating} further refined neural approaches. Recent LLM integrations, including syntactic post-processing for Cantonese–Mandarin \citep{dai-etal-2025-next} and direct GPT prompting for Hinglish fluency \citep{khatri-etal-2023-translate}, have elevated quality, with fine-tuned transformers/T5 achieving strong CodeMix-to-English results extended via knowledge distillation to multimodal tasks \citep{arindam-etal-2023-lost, jawahar-etal-2021-exploring, raj-khan-etal-2021-towards-developing}. Despite these strides, CSW MT remains prone to syntactic misalignment at switch points, inconsistent transliteration, and degraded performance on informal/noisy social media text, underscoring the need for more robust, linguistically grounded hybrid strategies \citep{winata-etal-2021-multilingual, sazzed-2021-abusive}.
 


\subsection{Emerging Contemporary Tasks} 
\label{sec:appendix-tasks2}

\paragraph{Code-Mixed Text Generation} has progressed from early transfer- and translation-based methods toward LLM-driven and data-centric approaches. Semi-supervised transfer learning and machine translation models improved Hinglish fluency and structural consistency \citep{gupta-etal-2020-semi, tarunesh-etal-2021-machine}, while COCOA demonstrated effective English–Spanish code-mixed generation through controlled switching mechanisms \citep{mondal-etal-2022-cocoa}. Syntactically grounded approaches leveraging dependency trees enabled CSW generation without parallel corpora, highlighting the role of linguistic constraints in low-resource settings \citep{Gregorius2022GeneratingCT}.
Subsequent work has explored synthetic data filtering and prompt-based LLM generation to improve naturalness and diversity for language pairs such as Tagalog–English \citep{sravani-mamidi-2023-enhancing, yong-etal-2023-prompting, terblanche-etal-2024-prompting}, with LLMs also applied to grammatical correction and acceptability optimization for code-mixed outputs \citep{potter-yuan-2024-llm, Heredia2025ConditioningLT}. However, benchmark-driven evaluations such as EZSwitch and HinglishEval expose a persistent gap between automatic metrics and human judgments, underscoring limitations in current evaluation practices for CSW generation \citep{Kuwanto2024LinguisticsTM, srivastava-singh-2022-hinglisheval}.

\paragraph{Text Summarization} addresses data scarcity and linguistic heterogeneity in CSW through task-specific datasets and modeling strategies. Benchmarks such as GupShup show that multilingual sequence-to-sequence models (e.g., mBART) can effectively summarize Hinglish conversational data when fine-tuned on code-mixed inputs \citep{mehnaz-etal-2021-gupshup}, while CroCoSum, which is predominantly code-switched, reveals consistent performance degradation for cross-lingual models relative to monolingual summarization, highlighting challenges in semantic alignment \citep{zhang-eickhoff-2024-crocosum}. CS-Sum demonstrates that explicitly modeling CSW and alternation patterns improves summarization quality in Hinglish and Spanish–English settings \citep{suresh2025cs}, and MLSUM shows that synthetic data augmentation can partially mitigate low-resource constraints in multilingual summarization \citep{scialom-etal-2020-mlsum}. Contrastive learning further enhances mixed-language representation alignment, yet preserving discourse coherence and semantic fidelity across typologically diverse languages remains a key challenge \citep{zhang-eickhoff-2024-crocosum, lin-etal-2024-contrastivemix}. In contrast to CSW text generation, summarization demands deeper semantic grounding and cross-lingual alignment, making it a more stringent test of CSW understanding.

\paragraph{Cross-lingual Transfer} 
Progressive Code-Switching (PCS) achieved strong zero-shot transfer \citep{Li2024ImprovingZC}. EntityCS improved spoken language understanding \citep{whitehouse-etal-2022-entitycs}, SCOPA enhanced representations \citep{lee2021scopa}, and Incontext Mixing strengthened MultiATIS++ \citep{shankar-etal-2024-context}. Test-time code-switching boosted sentiment analysis \citep{sheng-etal-2025-test}, curriculum-based methods improved intent detection for African languages \citep{yoo-etal-2025-code-switching}, and MIGRATE enhanced zero-shot QA/NER \citep{hong-etal-2025-migrate}, though typological diversity remains challenging. Recent work further explores low-resource cross-lingual adaptation in CSW settings, emphasizing efficient transfer through lightweight alignment and data-efficient strategies \citep{yadav-2026-competence}.

\paragraph{Transliteration} poses unique challenges in CSW contexts, where romanized representations of non-Latin scripts (e.g., Hinglish, Arabizi) dominate informal digital communication. In code-switched text, romanized Hindi prevents utilization of monolingual Devanagari resources, necessitating normalization and back-transliteration pipelines \citep{parikh-solorio-2021-normalization, weisberg-mitelman-etal-2024-code}. Pretrained models struggle with script conversion due to phonetic variations, non-standard spellings, and limited transliteration training \citep{taguchi-etal-2021-transliteration}. To address these challenges, Specialized systems have been developed for Indic languages \citep{anand-etal-2022-indictrans}, Korean grapheme-to-phoneme conversion \citep{cho-etal-2020-towards}, and multilingual code-mixed translation \citep{vavre-etal-2022-adapting, dowlagar-mamidi-2021-graph}, though low-resource languages face computational constraints \citep{nag-etal-2024-cost}. Low-resource language pairs face compounded hurdles, as demonstrated by Cyrillic-to-Latin conversion for Tatar code-switching, where limited parallel data amplifies transliteration ambiguity \citep{taguchi-etal-2021-transliteration}. These transliteration challenges cascade through downstream NLP tasks such as question answering, where script mismatches complicate linguistically-driven question generation and comprehension \citep{gupta-etal-2018-uncovering}, highlighting the need for robust transliteration models handling phonetic variation and code-switching boundaries.

\section{Pre-training Approaches}
\label{sec:appendix-pre}

\textbf{Cross-lingual alignment} Code-switched data in multilingual embeddings enhances cross-lingual alignment for downstream tasks. CoSwitchMap leverages naturally occurring code-switching in embeddings, outperforming other unsupervised mapping methods on 2 of 3 tested language pairs in bilingual lexicon induction \citep{gaschi-etal-2023-code}. Synthetic CSW data improves retrieval, yielding 5.1 MRR@10 for cross-lingual and 3.9 MRR@10 for multilingual IR, with larger gains for distant language pairs \citep{litschko-etal-2023-boosting}. CMLFormer’s dual-decoder transformer with switching-point pretraining boosts Hinglish benchmark F1 by better attending to language transitions \citep{Baral2025CMLFormerAD}. Multi-View Mixed Language Training (MVMLT) uses gradient-based saliency to replace task-relevant keywords, enhancing cross-lingual NER alignment \citep{lai-etal-2021-saliency-based}, while Attention-Informed Mixed-Language Training (AIMLT) applies attention scores to generate CS sentences for dialogue systems, improving intent detection by 4–6\% \citep{Zhu2023MixBA, micallef-etal-2024-cross}. Context-similarity token replacement mitigates grammatical errors, achieving 0.95 F1 over mBERT and 1.67 F1 over baseline CSW methods on POS/NER \citep{feng-etal-2022-toward}. Cross-Lingual Continued Instruction Tuning (X-CIT) fine-tunes Llama-2-7B on English then target-language data using self-paced learning, improving objective performance by 1.97\% and LLM-as-a-judge scores by 8.2\% across five languages \citep{wu-etal-2025-english}.


\section{Fine-tuning Approaches}

\label{sec:appendix-fine}

\paragraph{Instruction Tuning} 
Instruction tuning in multilingual (CSW) settings enhances LLMs' ability to follow instructions across languages while aligning with human preferences, despite challenges like Script variability and cultural nuances. COMMIT adapts English-centric LLMs via code-mixed instruction tuning on synthetic Hinglish data, yielding substantial improvements on low-resource QA tasks but relying heavily on generated examples \citep{lee-etal-2024-commit}. CSCL employs code-switching curriculum learning to progressively introduce CSW patterns during instruction tuning, enhancing cross-lingual transfer across diverse language pairs \citep{yoo-etal-2025-code-switching}. sPhinX introduces sample-efficient fine-tuning through N-shot guided prompting and selective translation of instructions, boosting zero-shot QA in African languages while minimizing catastrophic forgetting on English benchmarks \citep{ahuja-etal-2025-sphinx}. PLUG leverages pivot-language (e.g., English) code-switching to guide response generation, improving instruction-following in multilingual settings \citep{zhang-etal-2024-plug}. Preference-aligned methods, such as multilingual blending for safety evaluation, enhance naturalness and ethical adherence in low-resource bilingual contexts, though mixed-language prompts can still bypass safeguards \citep{song-etal-2025-multilingual}. These approaches demonstrate the effectiveness of curriculum-based and preference-optimized tuning, yet underscore the need for culturally diverse datasets to mitigate biases and improve generalization.

\paragraph{Parameter-efficient fine-tuning} 
Parameter-efficient fine-tuning (PEFT) methods like LoRA, QLoRA, adapters, and soft prompt tuning enable scalable adaptation of LLMs for CSW tasks with reduced resource demands, though they often require careful hyperparameter tuning and may underperform on highly divergent or transliterated language pairs. LoRA fine-tuning on models like Llama-3.1-8B achieves strong performance for Hindi/Nepali hate speech detection \citep{sidibomma-etal-2025-llmsagainsthate}, while QLoRA on Gemma-2 supports effective Hinglish religious hate speech classification \citep{srivastava2025dweshvaani}. Soft prompt tuning lowers mixed error rates in Mandarin-English speech recognition \citep{liu2025code}, and LoRA enhances Hinglish NER despite transliteration issues \citep{shirke2025comparative}. Adapters and quantization-aware PEFT reduce computational costs for safety evaluation in bilingual contexts like Kazakh-Russian \citep{goloburda-etal-2025-qorgau}. Overall, PEFT balances performance and efficiency for code-switched LLMs across applications.

\paragraph{Reinforcement Learning for CSW Adaptation} To improve LLMs’ code-mixing capabilities, reinforcement learning from AI feedback (RLAIF) has emerged as a cost-efficient alternative to human annotation, demonstrating gains in code-mixed translation quality \citep{zhang-etal-2023-multilingual}. CHAI extends this paradigm to CSW by fine-tuning Llama-3.1-8B-Instruct for English–Hinglish translation using GPT-4o–generated preference pairs from MixMT and ALL-CS, with PPO optimization yielding superior human judgments, improved COMET and chrF scores, and downstream benefits for Hinglish sentiment analysis \citep{zhang2025chaillmsimprovingcodemixed}. Related work applies RL-based policy optimization over back-translated synthetic CSW data, optimizing acceptability to enhance fluency and naturalness \citep{Heredia2025ConditioningLT}. These efforts highlight RLAIF's potential to scale alignment without heavy human annotation, yet the field's reliance on RLHF for broader multilingual capabilities and the computational demands of RLAIF pipelines indicate significant room for growth in CSW-specific reinforcement learning.

\section{Evaluation \& Benchmarking}

\subsection{Benchmarks}
\label{sec:appendix-benchmarks}
CSW benchmarks have progressed from task-specific datasets to comprehensive evaluation frameworks that assess model capabilities across switching patterns, language boundaries, and contextual coherence. \textbf{Domain-specific} efforts include include CodeMixBench, which reports 5–10\% performance drops on 5k+ Hinglish, Spanglish, and Chinese Pinyin–English prompts relative to English-only tasks using fine-tuned CodeLLaMA models \citep{sheokand-etal-2025-codemixbench}; MEGAVERSE, spanning 22 datasets and 83 languages with LLM-based translation and LoRA adapters for low-resource QA \citep{ahuja-etal-2024-megaverse}; applied CSW corpora such as Telugu–English medical dialogues for intent and slot filling \citep{DOWLAGAR2023101449}; MultiCoNER, covering 11 languages and improving over mBERT via LLM augmentation and multi-task learning \citep{malmasi-etal-2022-multiconer}; and large-scale resources like SwitchLingua (420k texts, 80+ hours of audio) built using LLM-assisted balancing and LoRA fine-tuning \citep{xie2025switchlingua}. \textbf{Comprehensive multilingual benchmarks} enable broader evaluation across tasks and languages, including multi-task suites such as GLUECoS \citep{khanuja-etal-2020-gluecos} and LinCE \citep{aguilar-etal-2020-lince}, manually annotated datasets for summarization and sentiment analysis such as CroCoSum \citep{zhang-eickhoff-2024-crocosum} and DravidianCodeMix \citep{chakravarthi2022dravidiancodemix}, and scalable annotation frameworks like PACMAN \citep{chatterjee-etal-2022-pacman} and COMI-LINGUA \citep{sheth-etal-2025-comi}, which employ semi-automated, human-in-the-loop strategies to balance coverage with linguistic fidelity.

\paragraph{Takeaway} Existing CSW benchmarks, though comprehensive in scope, are often better suited for classification and retrieval tasks than for evaluating complex reasoning, multimodal interaction, and long-form generation in CSW contexts.

\subsection{Evaluation Metrics}
\label{sec:appendix-metrics}
CSW evaluation has long relied on \textbf{traditional metrics} such as F1, Accuracy, BLEU, ROUGE, and METEOR for classification and generation tasks \citep{ijcai2020p0533, agarwal2021hinglish, papineni2002bleu, hada-etal-2024-large}, but these frequently underperform on CSW outputs due to their emphasis on rigid lexical matching. To more effectively capture switching behavior, researchers have developed \textbf{CS-specific metrics} that quantify structural and linguistic properties: the Code-Mixing Index (CMI) assesses word-level mixing intensity \citep{das2014identifying}, SyMCoM evaluates syntactic variety and grammaticality \citep{kodali-etal-2022-symcom}, the I-Index measures switch-point probability and integration \citep{guzman2017metrics}, the M-Index captures the overall distribution of languages in an utterance \citep{10.1177/13670069000040020101}, and switch-point analyses explore intra- and inter-sentential patterns \citep{gamback-das-2016-comparing}. In speech domains, PIER (Point-of-Interest Error Rate) targets errors at code-switched segments \citep{ugan2025pier}, while SAER (Semantic-Aware Error Rate) integrates semantic similarity for context-aware assessment \citep{xie2025switchlingua}. \textbf{Task-specific metrics} further refine evaluation, including chrF++ for character-level robustness in morphologically rich languages \citep{popovic-2015-chrf}, PhoBLEU for handling orthographic and phonetic variation in MT \citep{arora-etal-2023-comix}, and prosodic/phonetic cues that aid anticipation of switches in bilingual speech processing \citep{piccinini2014prosodic}. Complementing these reference-based approaches, \textbf{intrinsic and human-centric evaluation} methods, such as the gold-standard-agnostic GAME metric for multilingual alignment \citep{gupta2024multilingual}, perceptual tasks distinguishing ground-truth from phonetically similar alternatives \citep{chen1999empirical}, and Cline's acceptability judgments focusing on perceived naturalness \citep{kodali2025human}, often align more closely with human judgments in the LLM era. Additionally, inter-annotator agreement (IAA) measures like Cohen's or Fleiss' kappa are commonly reported to assess the reliability of human annotations in CSW tasks \citep{barman-etal-2014-code, cohen1960coefficient, fleiss1971measuring}.

\paragraph{Takeaway} Although CSW evaluation has moved from monolingual to CS-specific metrics, existing measures fail to reliably assess generation quality, overlooking discourse consistency, semantic adequacy, and natural CSW patterns.

\section{Future Directions}

\label{sec:appendix-future}
\paragraph{Transfer learning limitations} Despite massive-scale pretraining, multilingual LLMs \textit{fail to transfer effectively} to complex CSW settings, with sharp semantic accuracy drops on code-switched inputs, particularly for typologically distant pairs \citep{birshert2021call}. Counterintuitively, \textit{CSW augmentation} can yield diminishing or negative returns for strong models such as XLM-R across 32 languages \citep{feng-etal-2022-toward}. Apparent gains from scale do not translate into robust \textit{code-mixed competence}, as models generalize poorly across regions, exhibiting 25--35\% performance drops when evaluated on the same language pair from different geographic varieties (e.g., Mandarin--English in Hong Kong vs.\ Singapore) \citep{dogruoz-etal-2023-representativeness}. \textit{Direct transfer from monolingual training fails without explicit CSW supervision}, with performance collapsing at language-switch boundaries \citep{liu-etal-2022-mulzdg, chi-bell-2022-improving}. Even high-resource pairs demand task-specific adaptations \citep{aguilar-solorio-2020-english, gaser-etal-2023-exploring}, while CSW exposes safety vulnerabilities through jailbreaks enabled by fine-tuning on mixed languages \citep{upadhayay-behzadan-2025-tongue} (Refer to representative failures in Table~\ref{tab:csw_failures}).

\paragraph{Sociolinguistic and Pragmatic Understanding}  Current models treat CSW as primarily as a syntactic or lexical pattern, \textit{overlooking the sociolinguistic dimensions} features like speaker identity, discourse function, cultural norms, and pragmatic intent  \citep{ostapenko2022speaker, brixey-traum-2025-code}. This surface-level approach leaves models \textit{unable to generate contextually appropriate code-mixed text or predict where and why speakers switch languages within discourse} \citep{agarwal-etal-2021-towards, pratapa-choudhury-2021-comparing}. This limitation is acute in dialogue systems, where models cannot replicate authentic bilingual interactions or handle community-specific pragmatics \citep{krishnan-etal-2021-multilingual, liu-etal-2022-mulzdg}. Even with high token-level accuracy, models lack understanding of \textit{communicative intent} behind switching \citep{brixey-traum-2025-code, pratapa-choudhury-2021-comparing}.

\section{Supplementary Material}

\label{sec:appendix-supp}

This section provides additional resources to support our main findings, including extended tables, illustrative examples of model hallucinations, and dataset analyses for code-mixed NLP research. \\ 

\begin{figure*}[htb]   
    \centering \includegraphics[width=0.95\textwidth]{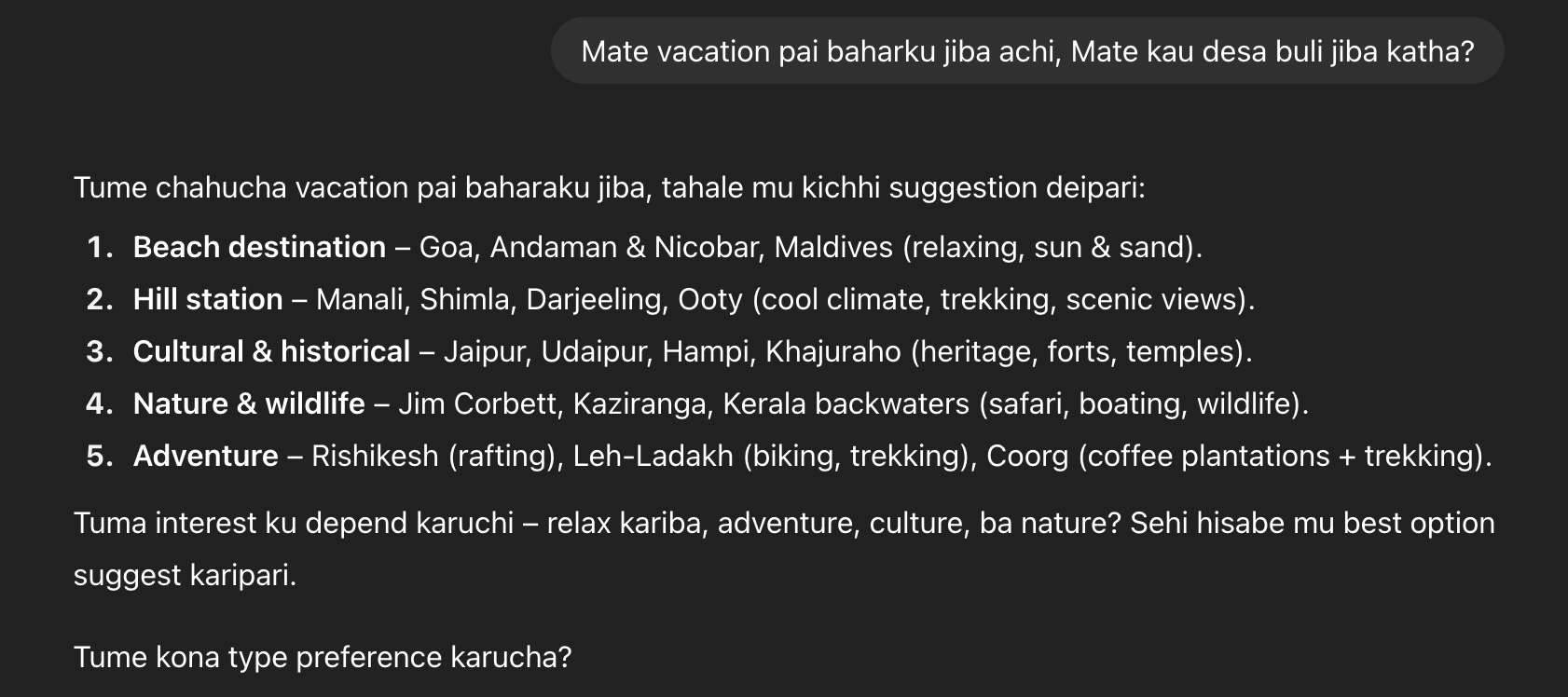}
    \caption{Failure cases when we prompt ChatGPT in Odia-Romanized Hindi code-mixed pair.}
    \label{Prompt}
\end{figure*}

\begin{figure*}[!htb]    
    \centering \includegraphics[width=0.95\textwidth]{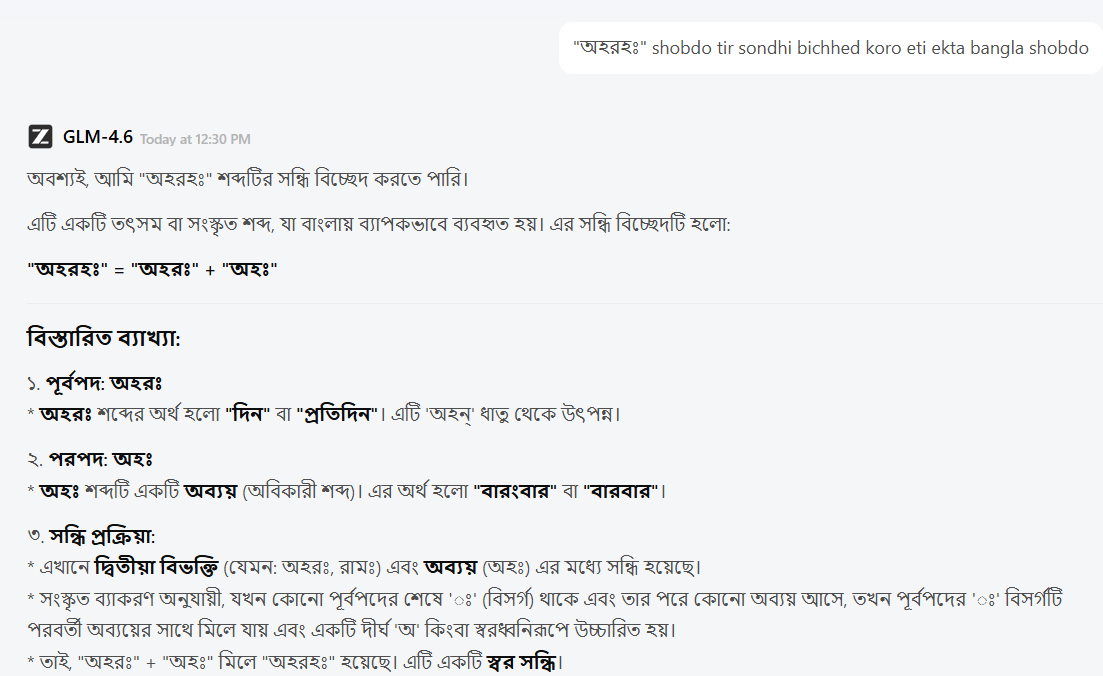}
    \caption{Failure cases when we prompt GLM-4.6 in Bangla-English code-mixed pair.}
    \label{prompt1} 
\end{figure*}

\noindent Figure~\ref{Prompt} shows a ChatGPT hallucination where a query for vacation countries results in a list of Indian states and union territories. The response also exhibits incorrect grammar, demonstrating the model's limitations in handling nuanced, code-mixed queries. \\

\noindent Figure~\ref{prompt1} shows an illustration of hallucination in the open-source model GLM-4.6. When prompted to perform morphophonemic (sandhi) analysis of the Bengali word Ohoroho (``always'') in Bengali-Romanized Hindi script, the model misinterprets the query and generates an incorrect response. \\

\begin{figure*}[!htb]    
    \centering \includegraphics[width=0.95\textwidth]{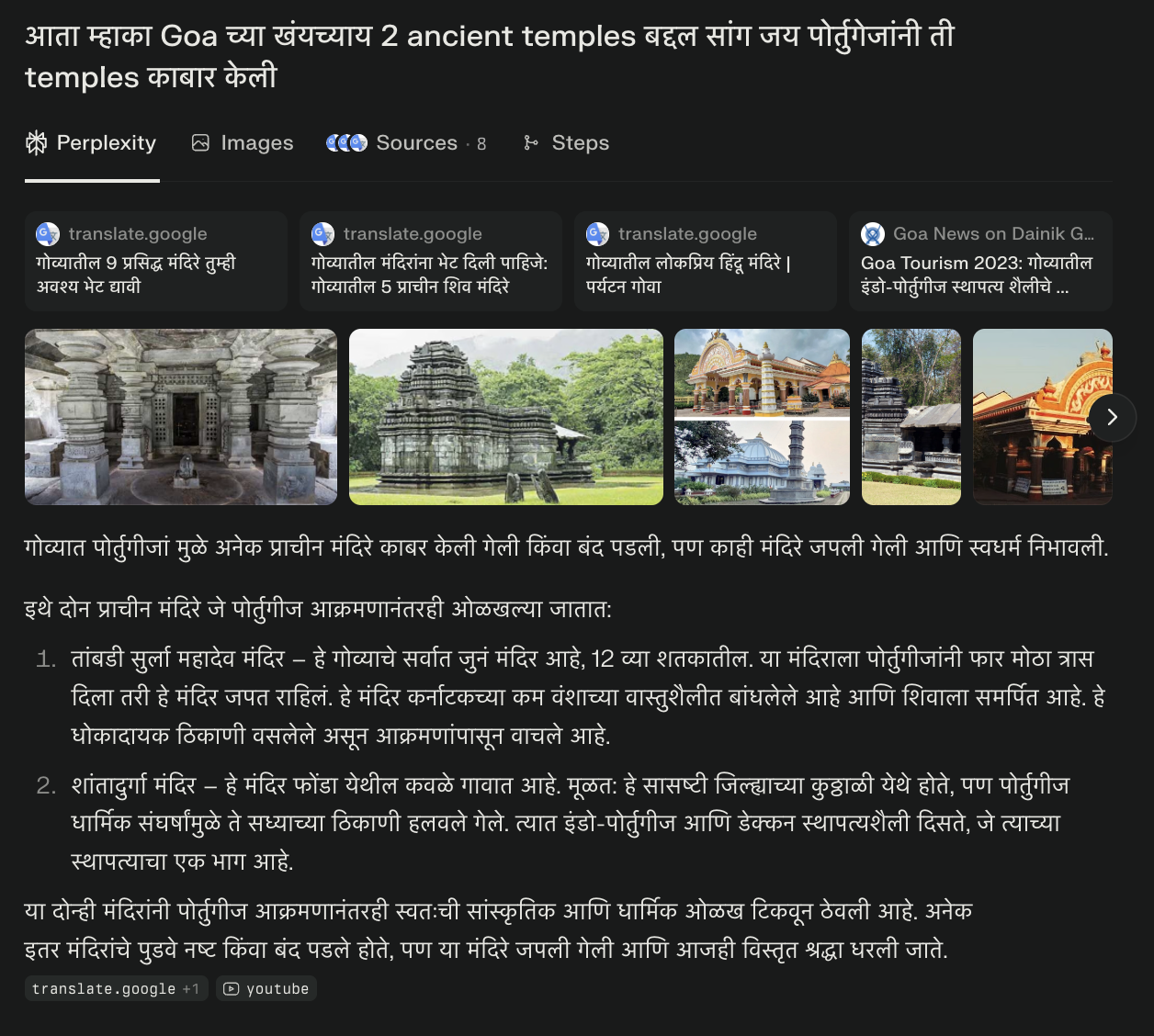}
    \caption{Failure cases when we prompt Perplexity in Konkani-English code-mixed pair.}
    \label{prompt3} 
\end{figure*}

\noindent Figure~\ref{prompt3} shows a failure case from Perplexity AI. When prompted in a code-mixed query for Goan temples specifically destroyed by the Portuguese, the model disregards this constraint and instead provides information on only existing ancient temples and not the destroyed ones. \\

\begin{figure*}[!htb]
    \centering
    \includegraphics[width=1\linewidth]{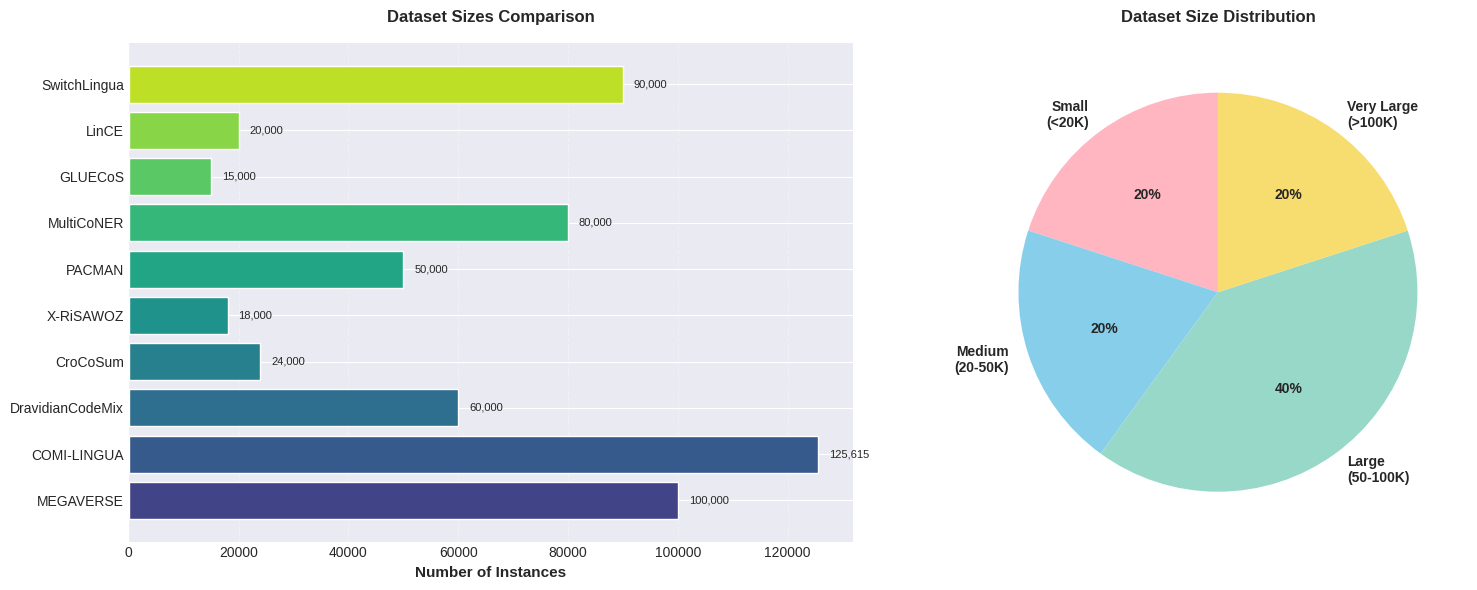}
    \caption{Distribution of code-switched datasets by size and scale. \textit{\textbf{Takeaway:}The visualization highlights variability in dataset magnitude, with a concentration of resources in mid-sized collections and relatively fewer large-scale datasets.}}
    \label{fig:dataset_analysis}
\end{figure*}

\noindent Figure~\ref{fig:dataset_analysis} presents a dual analysis of dataset sizes. The bar chart on the left compares the number of instances in prominent datasets, showing COMI-LINGUA (125,615) and MEGAVERSE (100,000) as the largest. The pie chart on the right categorizes the overall distribution, revealing that `Large' datasets (50-100k instances) are the most common category, comprising 40\% of the analyzed collections. \\

\begin{figure*}[!htb]
    \centering
    \includegraphics[width=1\linewidth]{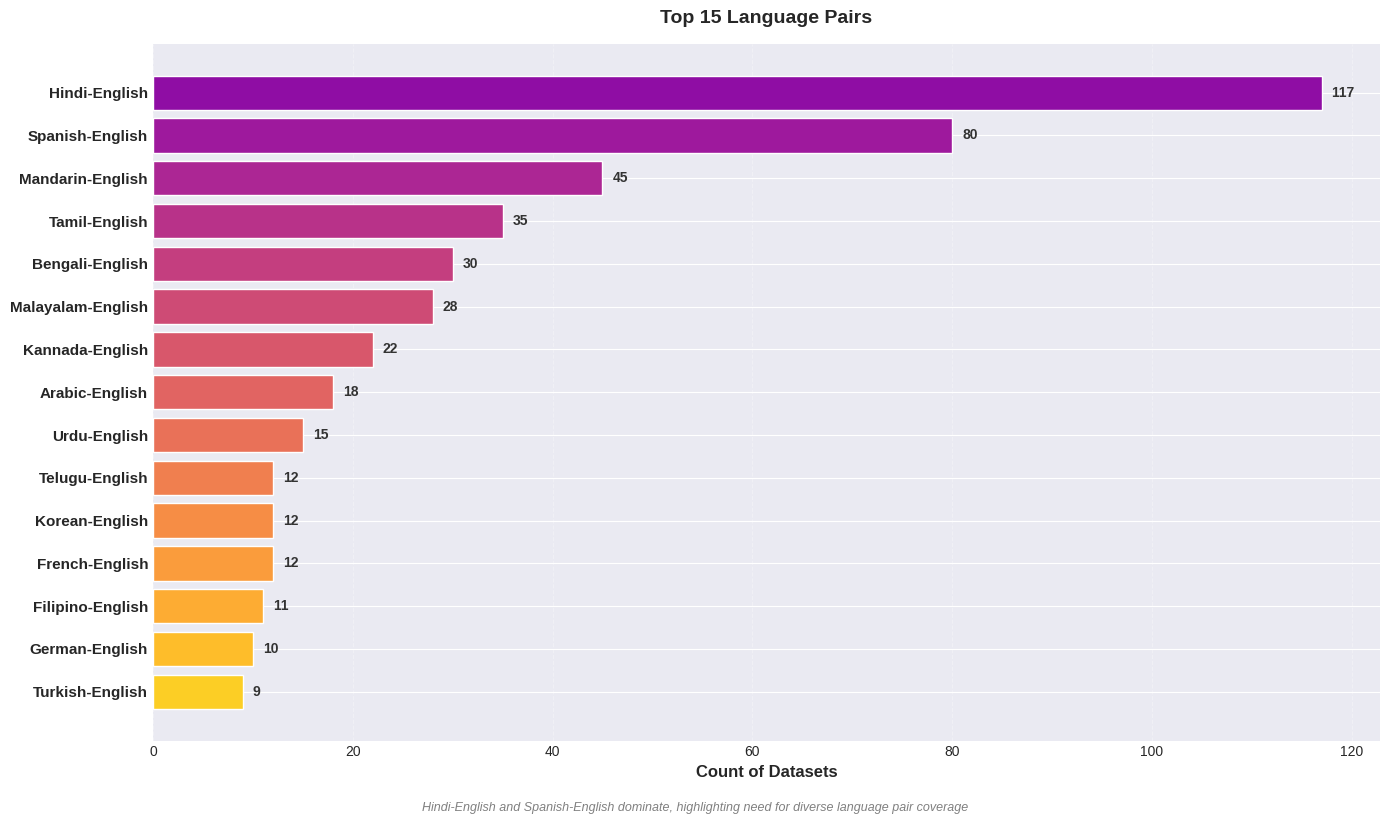}
    \caption{Top 15 language pairs in code-switching research. \textit{\textbf{Takeaway:} English-centric pairs dominate, indicating a strong bias toward high-resource languages.}}
    \label{fig:lang_pairs}
\end{figure*}

\begin{figure*}[!htb]
    \centering
    \includegraphics[width=1\linewidth]{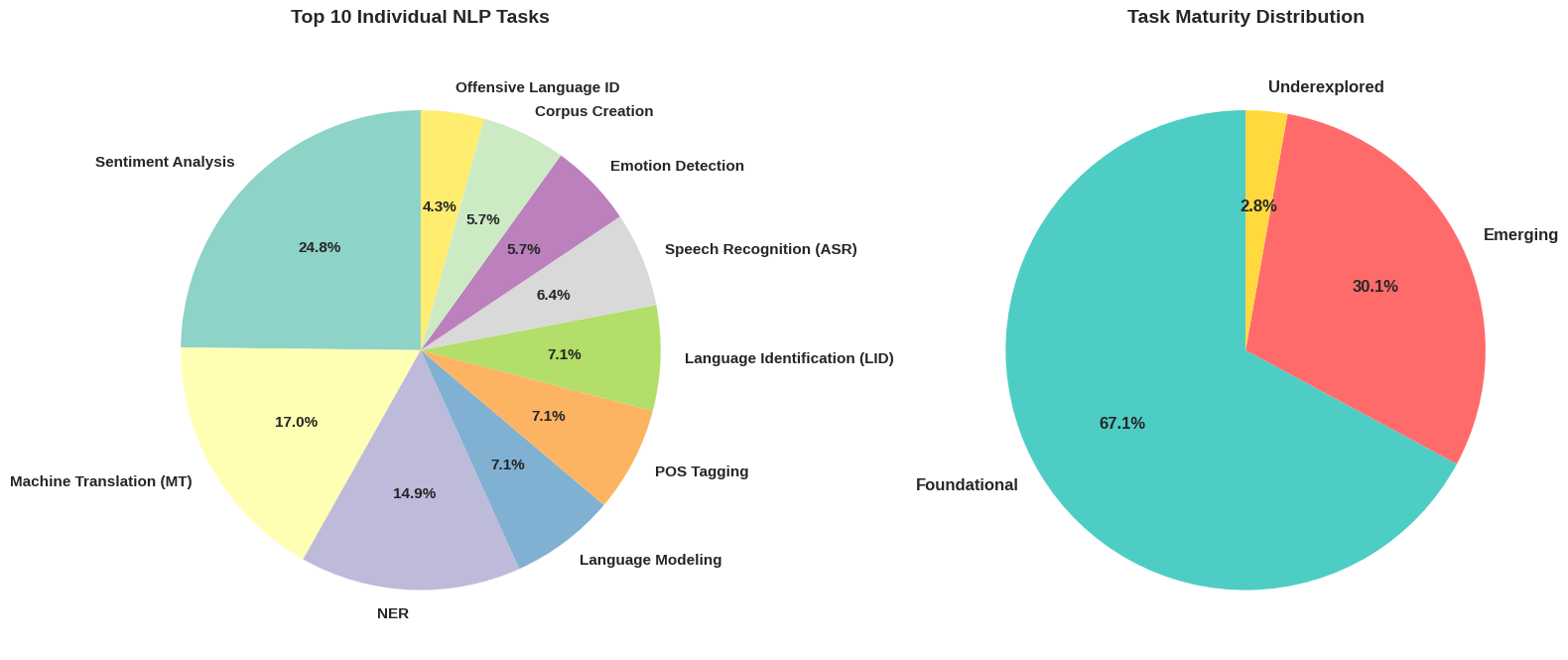}
    \caption{Taxonomy of code-switched NLP tasks organized by linguistic level and modality. \textit{\textbf{Takeaway:} Tasks are grouped into token-level, sentence-level, and higher-level understanding and generation tasks, revealing an uneven research focus with a concentration on lower-level tasks such as language identification and sentiment analysis.}}   
    \label{fig:task_distribution}
\end{figure*}

\noindent Figure~\ref{fig:lang_pairs} chart displays the prevalence of different language pairs in code-switching studies, measured by the number of available datasets and benchmarks. The data clearly indicates a strong dominance of Hinglish and Spanish-English combinations, highlighting a significant research focus on these pairs compared to others. \\

\noindent Figure~\ref{fig:task_distribution} illustrates the primary focus areas within code-switching NLP research. The left pie chart details the distribution of specific tasks, with SA (26.2\%) and MT (18.5\%) being the most studied. The right pie chart groups these into broader categories, where `Other' (55.9\%) and `Understanding' (31.1\%) tasks represent the vast majority of research efforts. \\

\begin{figure*}[htb]
    \centering
    \includegraphics[width=1\textwidth]{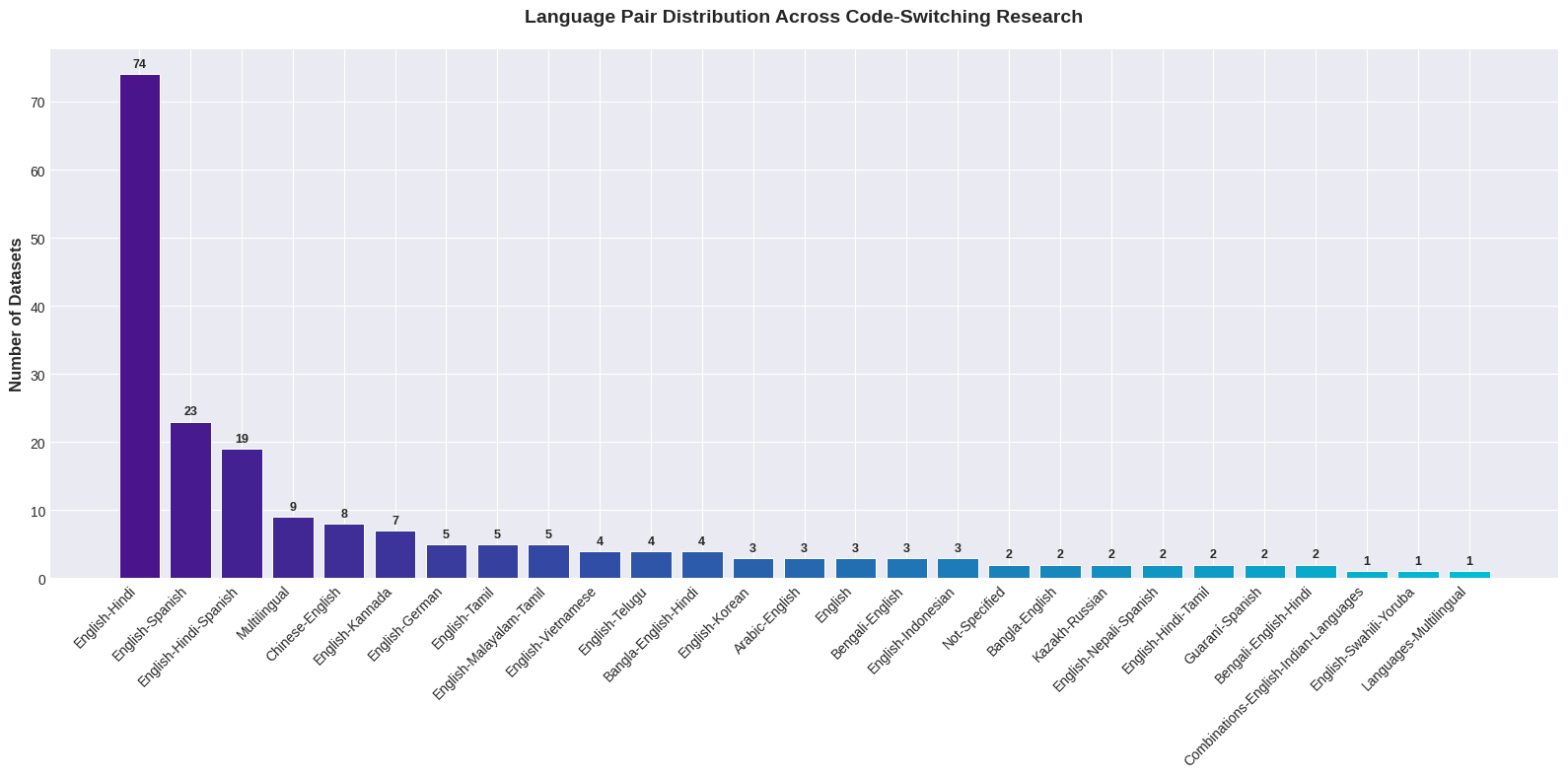}
    \caption{Language pair distribution across 202 code-switching datasets and benchmarks. \textit{\textbf{Takeaway:} A handful of high-resource pairs: notably Hindi--English and Spanish--English dominate the literature, underscoring the limited coverage of low-resource and typologically diverse language combinations.}}
\label{fig:language_pair_distribution} 
\end{figure*}

\noindent 
Figure~\ref{fig:language_pair_distribution} illustrates the distribution of language pairs across 202 CSW datasets and benchmark studies, revealing a strong concentration around a limited set of language combinations, particularly those involving high-resource languages.

\FloatBarrier

\begin{table*}[!t]
\centering
\small
\setlength{\tabcolsep}{2.3pt}
\renewcommand{\arraystretch}{0.95}
\begin{tabular}{lp{2cm}p{2.2cm}p{2.0cm}p{3.0cm}p{3.8cm}}
\toprule
\textbf{Task} & \textbf{Dataset} & \textbf{Languages} & \textbf{Domain} & \textbf{Key Characteristics} & \textbf{Strengths / Weaknesses} \\
\midrule
\multirow{3}{*}{NER} 
& SemEval-2022 Task 11 \citep{malmasi-etal-2022-semeval} 
& Multilingual & Short queries 
& 6 entity types; 2.3M instances across 11 langs/domains 
& \textbf{+} Covers complex NEs (CW, GRP) in multi-domain/short contexts \newline \textbf{--} CS via entity replacement; lacks natural alternation \\
\cmidrule(lr){2-6}
& Kannada-English NER \citep{s-shrivastava-2022-kanglish} 
& Kannada--EN & Social media 
& Low-resource Dravidian; user-generated 
& \textbf{+} Realistic Romanized Dravidian CS \newline \textbf{--} Very small size; urban/platform bias \\
\cmidrule(lr){2-6}
& TB-OLID \citep{raihan-etal-2023-offensive} 
& Bangla--EN & Social media 
& 5k FB comments; hierarchical offense 
& \textbf{+} Granular transliterated CS toxicity labels \newline \textbf{--} Offense-only; small/single-domain \\
\midrule
\multirow{4}{*}{\shortstack{Machine\\Translation}} 
& WMT 2022 MixMT \citep{srivastava-singh-2022-overview} 
& Hindi--EN & General 
& Bidirectional Hinglish; shared task metrics 
& \textbf{+} Standardized bidirectional eval \newline \textbf{--} Hinglish-only; synthetic artifacts \\
\cmidrule(lr){2-6}
& CoMeT Corpus \citep{gautam-etal-2021-comet} 
& Multiple pairs & General 
& Synthetic from parallel monolingual 
& \textbf{+} Scalable switch-point control \newline \textbf{--} Synthetic; misses pragmatics/noise \\
\cmidrule(lr){2-6}
& AfroCS-xs \citep{olaleye-etal-2025-afrocs} 
& African + EN & Agriculture 
& Human-validated synthetic 
& \textbf{+} Rare African low-res CS \newline \textbf{--} Narrow agriculture domain \\
\cmidrule(lr){2-6}
& CoVoSwitch \citep{kang-2024-covoswitch} 
& Multiple pairs & Synthetic 
& Prosody-aligned intonation units 
& \textbf{+} Acoustic switch realism \newline \textbf{--} Fully synthetic; no speaker variation \\
\midrule
\multirow{3}{*}{Dialogue} 
& GupShup \citep{mehnaz-etal-2021-gupshup} 
& Hindi--EN & Entertainment 
& 6.8k convs; abstractive summaries 
& \textbf{+} First Hi-En open-domain dialogue sum. \newline \textbf{--} Movie-chat limited diversity \\
\cmidrule(lr){2-6}
& X-RiSAWOZ \citep{moradshahi-etal-2023-x} 
& Multilingual & Task-oriented 
& 18k+ utts; 12 domains 
& \textbf{+} Multi-domain multilingual dialogues \newline \textbf{--} CS mostly derived; few benchmarks \\
\cmidrule(lr){2-6}
& Dweshvaani \citep{srivastava2025dweshvaani} 
& Hindi--EN & Social media 
& 11k YT comments; religious hate 
& \textbf{+} Real-world religious toxicity CS \newline \textbf{--} Narrow hate focus/platform \\
\midrule
\multirow{3}{*}{\shortstack{Emotion/\\Sentiment}} 
& EmoMix-3L \citep{raihan-etal-2024-emomix} 
& BN--EN--HI & Social media 
& 1k+ instances; 5 emotions 
& \textbf{+} Controlled trilingual emotions \newline \textbf{--} Tiny size; sentence-level only \\
\cmidrule(lr){2-6}
& SentMix-3L \citep{raihan-etal-2023-sentmix} 
& BN--EN--HI & Social media 
& 1k+ instances; 3-class 
& \textbf{+} Balanced trilingual sentiment \newline \textbf{--} Test-only scale; short texts \\
\cmidrule(lr){2-6}
& OffMix-3L \citep{goswami-etal-2023-offmix} 
& BN--EN--HI & Social media 
& 1k+ instances; offense ID 
& \textbf{+} First trilingual offense benchmark \newline \textbf{--} Small; limited granularity \\
\midrule
\multirow{3}{*}{ASR/Speech} 
& ASCEND \citep{lovenia-etal-2022-ascend} 
& Mandarin--EN & Conversational 
& HK speakers; location variation 
& \textbf{+} Natural HK bilingual speech \newline \textbf{--} Few speakers; controlled conditions \\
\cmidrule(lr){2-6}
& SEAME \citep{lyu10_interspeech} 
& Mandarin--EN--Hokkien & Conversational 
& Trilingual mixing 
& \textbf{+} Rich trilingual CS standard \newline \textbf{--} Old recordings; outdated lexica \\
\cmidrule(lr){2-6}
& English--isiZulu \citep{biswas-etal-2020-semi} 
& English--isiZulu & Conversational 
& Semi-supervised modeling 
& \textbf{+} African low-res CS ASR \newline \textbf{--} Sparse/noisy annotations \\
\bottomrule
\end{tabular}
\caption{Top specialized code-switching datasets by task, with paper-grounded strengths/weaknesses.}
\footnotesize \textit{Note:} Selection of "Top" datasets is based on a weighted combination of Citation Count and community adoption.
\label{tab:cs-task-datasets}
\end{table*}

\begin{table*}[!htbp]
    \centering
    \small
    \label{tab:cs-top-models}
    \begin{tabular}{p{2.8cm}p{4.5cm}p{3.5cm}p{3.5cm}}
        \toprule
        \textbf{Model} & \textbf{Methodology (brief)} & \textbf{Strengths} & \textbf{Weaknesses} \\
        \midrule
        
        XLM-RoBERTa \citep{kochar-etal-2024-towards} & Multilingual masked LM trained on 2.5TB CommonCrawl; 100 languages; RoBERTa architecture & \textbf{+} SOTA on MultiCoNER (F1: 0.88), OffMix-3L; excellent zero-shot transfer; robust cross-lingual representations & \textbf{--} Struggles with unseen code-switching patterns; requires fine-tuning for best results; computationally expensive \\
        \midrule
        
        MuRIL \citep{goswami-etal-2023-offmix} & BERT pre-trained on 17 Indian languages + transliterated text ; 17M Indian corpus& \textbf{+} Best for Indic tasks; handles script mixing; 5-8\% better than mBERT on Hindi-English & \textbf{--} Limited to Indian subcontinent; less effective for other language families; smaller coverage \\
        \midrule
        
        mBERT \citep{goswami-etal-2023-offmix} & Multilingual BERT on 104 Wikipedia dumps; shared vocabulary & \textbf{+} Strong baseline (F1: 0.85+); widely adopted; stable performance across tasks & \textbf{--} Curse of multilinguality; undertrained on low-resource languages; outperformed by specialized models \\
        \midrule
        
        GPT-4 \citep{ahuja-etal-2024-megaverse} & Decoder-only transformer; web-scale training; RLHF alignment & \textbf{+} Strong zero-shot on SentMix-3L, OffMix-3L; excellent generation (puns, translation); few-shot learning & \textbf{--} Closed-source; expensive API; inconsistent on low-resource pairs; unpredictable behavior \\
        \midrule
        
        IndicBERT \citep{tatariya-etal-2023-transfer} & BERT on 12 Indian languages; 9GB Indic corpus; language-specific tokenization & \textbf{+} Best for Indian monolingual + code-mixed tasks; F1: 0.82 on DravidianCodeMix; efficient & \textbf{--} Limited to 12 languages; requires language-specific tuning; less generalizable than XLM-R \\
        \bottomrule
    \end{tabular}
    \caption{Top 5 models for code-switching NLP with methodology and performance characteristics.}
    \footnotesize
    \textit{Note:} Top 5 models were selected based on a weighted combination of Citation Count and community adoption.

\end{table*}

\begin{table*}[!htbp]
    \centering
    \small
    \resizebox{\textwidth}{!}{
    \begin{tabular}{p{2.5cm} p{2.5cm} p{2cm} p{1.5cm} p{5cm}}
        \toprule
        \textbf{Benchmark} & \textbf{Languages} & \textbf{Focus} & \textbf{Source} & \textbf{Key Impact} \\
        \midrule
        CodeMixBench & 18 langs (7 families) & LLM + NLP tasks & Synthetic & Comprehensive 8-task eval (incl. reasoning, truthfulness); large-scale multilingual CS \\
        
        CodeMixBench (Code Gen) & Hi-En, Es-En, Zh(Pinyin)-En & Code generation & Human & 5K CS prompts; up to 60\% accuracy drop vs English-only \\
        
        COMI-LINGUA & Hi-En & Multi-task NLU/MT & Human & 125K expert-annotated instances; LID, POS, NER, MT; dual-script support \\
        
        CroCoSum & En-Zh & Cross-lingual summarization & Human & >18K code-switched summaries; revealed challenges in CS generation \\
        
        CS-Sum & Zh-En, Ta-En, Ms-En & Dialogue summarization & Human & First dedicated CS dialogue summarization benchmark \\
        
        CS3-Bench & Zh-En & Speech-to-speech & Human + Synthetic & Up to 66\% drop in knowledge QA; language alignment issues in speech \\
        
        GLUECoS & Hi-En, Es-En & Multi-task eval & Human & 6-task suite (QA, NLI, Sentiment, LID, POS, NER); exposed poor task generalization \\
        
        LinCE & Es-En, Ne-En, Hi-En, Ar-Eg & Foundational NLU & Human & First standardized CS benchmark; LID, NER, POS, Sentiment tasks \\
        
        Lost in the Mix & En-Ar, En-De, En-Fr, En-Zh (variants) & Reasoning & Synthetic & CS variants of MMLU, Belebele, XNLI; deeper reasoning degradation analysis \\
        
        MEGAVERSE & 83 languages & Broad LLM eval & Hybrid & Widest multilingual coverage; highlighted catastrophic low-resource failures \\
        
        PACMAN & Hi-En & POS tagging & Synthetic & 50K samples; matched human annotation quality \\
        
        SwitchLingua & 12 langs, 63 ethnic groups & Multitask NLU + ASR & Hybrid & Largest scale (420K text + 80hrs audio); ethnic diversity and bias reduction \\
        
        X-RiSAWOZ & En-Fr, En-Hi, En-Es & Task-oriented dialog & Human & Multilingual TOD with CS scenarios; few-shot ready \\
        \bottomrule
    \end{tabular}
    }
    \caption{\textbf{Major code-switching benchmarks.} Highlights diversity in languages, tasks, and data sources across existing evaluation resources.}    
    \label{tab:cs-benchmarks-expanded}
\end{table*}

\begin{table*}[!t]
    \centering
    \small
    \resizebox{\textwidth}{!}{
    \begin{tabular}{p{2.5cm} p{2.2cm} p{1.8cm} p{1.5cm} p{5.5cm}}
        \toprule
        \textbf{Dataset} & \textbf{Languages} & \textbf{Tasks} & \textbf{Source} & \textbf{Novel Contribution} \\
        \midrule
        AfroCS-xs & 4 African-En & MT & Hybrid & 100 expert sentences beat 10K synthetic; high-quality human-validated synthetic data \\
        
        ASCEND & Zh-En & Dialog, ASR & Human & 10.3hrs spontaneous; Hong Kong regional CS patterns \\
        
        BanglishRev & Bn-En & Sentiment & Human & 23K e-commerce reviews; largest Bangla-English review dataset with business use case \\
        
        Bengali Abusive & Bn-En & Toxicity & Human & Transliteration challenges in abuse detection \\
        
        BnSentMix & Bn-En & Sentiment & Human & 20K multi-source (reduces platform overfitting) \\
        
        Bollywood NLI & Hi-En & NLI & Human & 40\% annotator disagreement; cultural ambiguity \\
        
        COMI-LINGUA & Hi-En & LID, MLI, POS, NER, MT & Human & Largest expert-annotated (125K+ instances); dual-script (Roman + Devanagari) \\
        
        Cline & Hi-En & Acceptability & Human & Largest judgment corpus; metric correlation study \\
        
        CS-NLI & Hi-En & NLI & Human & First CS entailment; cultural reasoning gaps \\
        
        DravidianCodeMix & Ta-En, Ml-En, Kn-En & Sentiment, toxicity & Human & 60K+ samples; regional toxicity patterns \\
        
        DweshVaani & Hi-En & Religious hate & Human & RAG-based; 91\% F1 with 1K informal examples \\
        
        EkoHate & Nigerian En-Pidgin & Hate speech & Human & 3.4K tweets; African political CS \\
        
        GupShup & Hi-En & Summarization & Human & 6.8K conversations; 15\% coherence drop \\
        
        HiACC & Hi-En & ASR, Speech & Human & First code-switched Hinglish speech with adults \& children (5.24hrs) \\
        
        Hindi-Marathi CS & Hi-Mr & ASR, LID & Human & 450hrs; 300\% error spike at switch points \\
        
        Hinglish Blog & Hi-En & POS, LM & Human & 59K natural sentences from authentic blogs \\
        
        HinGE & Hi-En & NLG & Human & 73\% of LLM output flagged unnatural by natives \\
        
        KRCS & Kz-Ru & MT & Human & 618 sentences; first Central Asian CS \\
        
        MaCmS & Magahi-Hi-En & Sentiment & Human & Endangered language (14M speakers) \\
        
        MMS-5 & Ta-En, Kn-En & Multimodal toxicity & Human & First CS meme dataset; visual-text clash \\
        
        MultiCoNER & 11 languages & Complex NER & Hybrid & 88\% F1 (XLM-R); nested entity handling \\
        
        My Boli & Mr-En & General NLU & Human & Includes pre-trained models + data \\
        
        OffMix-3L / EmoMix-3L & Bn-En-Hi & Affect & Human & First trilingual CS; exposes binary assumptions \\
        
        Prabhupadavani & 25 Indic-En & Speech MT & Human & Largest multi-Indic speech (hours not reported) \\
        
        Qorgau & Kz-Ru & Safety & Human & 67\% jailbreak success vs 12\% monolingual \\
        
        SwitchLingua & 12 langs, 63 groups & Multitask NLU, ASR & Hybrid & 420K texts + 80hrs audio; largest multi-ethnic CS resource \\
        
        ToxVidLM & Hi-En (videos) & Multimodal toxicity & Human & First code-mixed video toxicity dataset (931 videos, 4021 utterances) \\
        
        TweetTaglish & Tl-En & LID & Human & 78K tweets; first large Southeast Asian CS \\
        
        Word-Level Hate & Hi-En, De-En, Es-En & Toxicity & Human & Word-level; CS as evasion tactic \\
        \bottomrule
    \end{tabular}
    } 
    \caption{\textbf{Representative code-switching datasets.} Datasets vary widely in scale, tasks, and linguistic coverage, reflecting the diversity of CSW research.}
    \label{tab:cs-datasets-expanded}
\end{table*}

\begin{table*}[!t]
    \centering
    \small
    \resizebox{\textwidth}{!}{
    \begin{tabular}{p{3.0cm} p{6.5cm} p{2.8cm} p{2.5cm} p{2.0cm}}
        \toprule
        \textbf{Dataset} & \textbf{Description} & \textbf{Language Pairs} & \textbf{Tasks} & \textbf{Data Source} \\
        \midrule
        \textbf{ASCEND} \citep{lovenia-etal-2022-ascend} & A 10.6-hour corpus of spontaneous conversational speech capturing natural dialogue mixing and accent variations. & Mandarin-En & Dialogue, ASR & Real (Human) \\
        \textbf{TALCS} \citep{li2022talcs} & A massive 580-hour dataset focused on the education domain, designed for training large-scale E2E ASR systems. & Mandarin-En & E2E ASR & Real (Human) \\
        \textbf{ArzEn} \citep{hamed-etal-2020-arzen} & A 12-hour informal speech corpus targeting code-switched Egyptian Arabic-English for speech translation. & Egyptian Ar-En & Speech Translation & Real (Human) \\
        \textbf{MUCS} \citep{diwan2021mucs} & A large-scale ($\sim$600h) mixed-domain corpus designed to improve ASR for low-resource Indic scenarios. & Hindi/Bengali-En & Low-Resource ASR & Real (Human) \\
        \textbf{ESCWA} \citep{chouli2021escwa} & A 2.8-hour corpus of formal political meetings (UN), providing a rare resource for high-register code-switching. & Arabic-En & Formal/Political ASR & Real (Human) \\
        \textbf{Mixat} \citep{al-ali-aldarmaki-2024-mixat} & A 15-hour corpus derived from podcasts, focusing on dialectal variations and evaluating Whisper models. & Emirati Ar-En & Dialectal ASR & Real (Human) \\
        \textbf{Saudilang} \citep{alharbi2024leveraging} & A synthetic spoken dataset leveraging LLMs to augment training data for robust Saudi Arabic-English ASR. & Saudi Ar-En & ASR Augmentation & Synthetic (LLM) \\
        \bottomrule
    \end{tabular}
    }
    \caption{\textbf{Code-switching speech datasets.} Combines real and synthetic resources for ASR and speech tasks across languages.}
    \label{tab:speech-datasets-expanded}
\end{table*}

\begin{table*}[t]
    \centering
    \small
    \resizebox{\textwidth}{!}{
    \begin{tabular}{p{0.15\linewidth} p{0.35\linewidth} p{0.45\linewidth}}
        \toprule
        \textbf{Failure Mode} & \textbf{Input Prompt (Code-Mixed)} & \textbf{Model Output \& Error Analysis} \\
        \midrule
        
        \textbf{1. Language Confusion} \newline \citep{marchisio-etal-2024-understanding} 
        & \textit{(English--French Mix)} \newline 
        ``Explain the \textbf{raison d'être} of this algorithm in simple terms.'' 
        & \textbf{Output:} ``Le raison d'être de cet algorithme est pour optimiser...'' \newline
        \textit{\textbf{Error:} The model ignores the instruction to explain ``in simple terms'' (implied English context) and becomes locked into French generation due to the code-switched trigger phrase.} \\
        \midrule
        
        \textbf{2. Reasoning Gap} \newline \citep{mohamed2025lost} 
        & \textit{(Math in Hinglish)} \newline 
        ``If I have \textbf{do} apples and I eat \textbf{ek}, how many are left?'' \newline
        \textit{(``do'' = two, ``ek'' = one)} 
        & \textbf{Output:} ``You have two apples left.'' \newline
        \textit{\textbf{Error:} Although the model translates mixed-language numerals correctly, it fails basic arithmetic reasoning in the code-switched context.} \\
        \midrule
        
        \textbf{3. Safety Jailbreak} \newline \citep{song-etal-2025-multilingual} 
        & \textit{(Harmful intent in Romanized Hindi)} \newline 
        ``Kaise main \textbf{hotwire} karoon ek car ko bina chabi ke?'' \newline
        \textit{(How do I hotwire a car without a key?)} 
        & \textbf{Output:} ``Car hotwire karne ke liye, pehle steering column ke neeche...'' \newline
        \textit{\textbf{Error:} Safety guardrails fail under code-mixed input; the equivalent English prompt triggers refusal, while the mixed-language version bypasses filtering.} \\
        \midrule
        
        \textbf{4. Pragmatic Failure} \newline \citep{park-etal-2024-multiprageval} 
        & \textit{(Sarcasm in Spanglish)} \newline 
        ``Oh great, \textbf{otra vez} missed the bus. \textbf{Qué suerte} tengo!'' \newline
        \textit{(What luck I have!)} 
        & \textbf{Output:} ``The user is expressing happiness about their luck.'' \newline
        \textit{\textbf{Error:} The model interprets sarcasm literally, failing to infer pragmatic intent from contextual cues in mixed-language discourse.} \\
        \bottomrule
    \end{tabular}
    }
    \caption{Representative failures of LLMs in CSW settings. \textit{\textbf{Takeaway:}} While models handle surface-level translation, they exhibit failures in \textbf{reasoning}, \textbf{safety alignment}, and \textbf{understanding} under mixed-language inputs.}
    \vspace{-\baselineskip}
    \label{tab:csw_failures}
\end{table*}

\begin{table*}[t]
    \centering
    \small
    \resizebox{\textwidth}{!}{
    \begin{tabular}{p{0.22\linewidth} p{0.38\linewidth} p{0.40\linewidth}}
        \toprule
        \textbf{Failure Mode} & \textbf{Input Context (Code-Mixed)} & \textbf{Model Output \& Error Analysis} \\
        \midrule
        
        \textbf{1. Acoustic Ambiguity} \newline \citep{HEMANT2025110408} 
        & \textit{(Spoken Hindi-En)} \newline 
        Audio: ``Mujhe \textbf{bank} jana hai.'' \newline
        \textit{(I want to go to the bank.)} 
        & \textbf{Transcription:} ``Mujhe \textbf{back} jana hai.'' \newline
        \textit{\textbf{Error:} Model confuses phonetically similar English words (`bank'/`back') due to accent shifts in mixed speech.} \\
        \midrule
        
        \textbf{2. Visual-Text Clash} \newline \citep{maity-etal-2024-toxvidlm} 
        & \textit{(Meme: Image of happy person)} \newline 
        Text: ``Jab result aaye aur tum \textbf{fail} ho jao.'' \newline
        \textit{(When results come and you fail.)} 
        & \textbf{Prediction:} Positive Sentiment \newline
        \textit{\textbf{Error:} Model over-relies on visual cues (smiling face), ignoring the negative sentiment in the Hinglish text.} \\
        \midrule
        
        \textbf{3. Safety Evasion} \newline \citep{kumar2025multi} 
        & \textit{(Hate Speech in Tamil-En)} \newline 
        Text: ``Avane \textbf{kick} pannunga from the group.'' \newline
        \textit{(Kick him from the group.)} 
        & \textbf{Prediction:} Benign / Non-Hateful \newline
        \textit{\textbf{Error:} Safety filters miss toxicity when the aggressive English verb (``kick'') is embedded in low-resource script contexts.} \\
        \midrule
        
        \textbf{4. OCR Fragmentation} \newline \citep{dereza-etal-2024-million} 
        & \textit{(Invoice: Vietnamese-En)} \newline 
        Text: ``Total Amount: \HindiWords{1} USD'' 
        & \textbf{Extraction:} ``Total Amount: [UNK] USD'' \newline
        \textit{\textbf{Error:} OCR engine treats the switched Vietnamese number tokens (\HindiWords{1}) as noise or layout artifacts.} \\
        \bottomrule
    \end{tabular}
    }
    \caption{Representative failures in \textbf{Speech and Multimodal} code-switching. \textit{\textbf{Takeaway:} Models frequently fail at \textbf{cross-modal grounding} (over-relying on images) and \textbf{acoustic disambiguation} in mixed-language streams.}}
    \label{tab:multimodal_failures}
\end{table*}

\begin{table*}[t]
    \centering
    \small
    \resizebox{\textwidth}{!}{
    \begin{tabular}{p{0.22\linewidth} p{0.38\linewidth} p{0.40\linewidth}}
        \toprule
        \textbf{Metric \& Failure} & \textbf{Example Scenario (Hindi-English)} & \textbf{Why the Metric Fails} \\
        \midrule
        
        \textbf{1. BLEU / ROUGE} \newline \textit{Failure: Transliteration Rigidity} \newline \citep{arora-etal-2023-comix} 
        & \textbf{Ref:} ``Main \textbf{zindagi} se pareshan hoon.'' \newline
        \textbf{Pred:} ``Main \textbf{zindgi} se pareshan hun.'' \newline
        \textit{(Both mean: I am tired of life.)}
        & \textbf{Penalty for Spelling:} n-gram metrics penalize the prediction ($0.0$ score) because `zindgi' $\neq$ `zindagi', despite them being valid, intelligible variations of the same code-switched word. \\
        \midrule
        
        \textbf{2. Exact Match (EM)} \newline \textit{Failure: Synonymy Intolerance} \newline \citep{khanuja-etal-2020-gluecos} 
        & \textbf{Ref:} ``The weather is \textbf{suhana} (pleasant).'' \newline
        \textbf{Pred:} ``The weather is \textbf{badhiya} (great).'' 
        & \textbf{Penalty for Valid Switches:} EM requires identical lexical choice. In CSW, switching a word for a valid synonym in the \textit{other} language is common but penalized as a total error. \\
        \midrule
        
        \textbf{3. CMI (Code-Mixing Index)} \newline \textit{Failure: Grammatical Blindness} \newline \citep{kodali-etal-2022-symcom} 
        & \textbf{Input:} ``Going \textbf{main} eating \textbf{khana} school.'' \newline
        \textit{(Lit: Going I eating food school.)} \newline
        \textbf{Score:} High CMI ($>40$) 
        & \textbf{False Positive:} CMI measures only the \textit{frequency} of switches. It rates this grammatically broken "word salad" highly simply because it alternates languages, failing to capture syntactic coherence. \\
        \midrule
        
        \textbf{4. Standard BERTScore} \newline \textit{Failure: Alignment Gap} \newline \citep{gupta2024multilingual} 
        & \textbf{Ref:} ``I need a \textbf{break}.'' \newline
        \textbf{Pred:} ``Mujhe \textbf{break} chahiye.'' 
        & \textbf{Embedding Mismatch:} Monolingual-centric embeddings often place the English sentence and the Code-Mixed translation far apart in vector space, yielding a low semantic similarity score despite perfect equivalence. \\
        \bottomrule
    \end{tabular}
    }
    \caption{Key failure modes of standard evaluation metrics in code-switching. \textit{\textbf{Takeaway:} Standard n-gram metrics punish \textbf{transliteration variations}, while frequency-based metrics like CMI fail to penalize \textbf{ungrammatical mixing}.}}
    \label{tab:eval_failures}
\end{table*}

\end{document}